\documentclass[11pt]{article}

\usepackage[utf8]{inputenc} 
\usepackage[T1]{fontenc}    
\usepackage{hyperref}       
\usepackage{url}            
\usepackage{booktabs}       
\usepackage{amsfonts}       
\usepackage{nicefrac}       
\usepackage{microtype}      
\usepackage{xcolor}         
\usepackage[numbers, compress]{natbib}

\usepackage[
    top=2.0cm,
    bottom=2.0cm,
    left=2.2cm,
    right=2.2cm
]{geometry}

\title{How Does Parameter Pruning Reshape DNN Representations? An Interaction-Driven Exploration}

\author{%
    Fangbo Li\thanks{Email: \texttt{lifangbo\_2003@sjtu.edu.cn}}
    \and
    Junpeng Zhang\thanks{Email: \texttt{zhangjp63@sjtu.edu.cn}}
    \and
    Qihan Ren\thanks{Email: \texttt{renqihan@sjtu.edu.cn}}
    \and
    Quanshi Zhang\thanks{Corresponding author. Email: \texttt{zqs1022@sjtu.edu.cn}}
}

\date{}

\usepackage{amsmath,amsfonts,bm}

\def\eqref#1{equation~\ref{#1}}

\def\1{\bm{1}}

\def\vx{{\bm{x}}}

\DeclareMathAlphabet{\mathsfit}{\encodingdefault}{\sfdefault}{m}{sl}
\SetMathAlphabet{\mathsfit}{bold}{\encodingdefault}{\sfdefault}{bx}{n}

\DeclareMathOperator{\sign}{sign}

\usepackage[utf8]{inputenc}
\usepackage[T1]{fontenc}
\usepackage{hyperref}
\usepackage{url}
\usepackage{booktabs}
\usepackage{amsfonts}
\usepackage{nicefrac}
\usepackage{microtype}
\usepackage{xcolor}
\usepackage{graphicx} 
 
\newtheorem{definition}{Definition}[section] 
\newtheorem{proof}{Proof}[section] 
\usepackage{placeins}
\usepackage{bm}
\usepackage{bbm}
\usepackage{algorithm,algorithmic}
\usepackage[rightcaption]{sidecap}
\usepackage{footmisc} 
\usepackage{cleveref}
\usepackage{enumitem}
\usepackage[most]{tcolorbox}
\usepackage{amsmath}
\usepackage{makecell}

\newcommand{\x}{\vx}

\newcommand{\y}{y^*}

\newcommand{\typeand}{\operatorname{and}}
\newcommand{\typeor}{\operatorname{or}}

\newcommand{\omegaand}{\Omega^{\typeand}}
\newcommand{\omegaor}{\Omega^{\typeor}}

\newcommand{\high}{\textnormal{high}}

\newcommand{\low}{\textnormal{low}}

\newcommand{\p}{\tilde{v}}

\newcommand{\offset}{\kappa}

\newcommand{\omegaandp}{\widetilde{\Omega}^{\typeand}}
\newcommand{\omegaorp}{\widetilde{\Omega}^{\typeor}}

\newcommand{\Rand}{R^{\typeand}_{S}}
\newcommand{\Ror}{R^{\typeor}_{S}}

\newcommand{\Pand}{P^{\typeand}_{S}}
\newcommand{\Por}{P^{\typeor}_{S}}

\newcommand{\Eand}{E^{\typeand}_{S}}
\newcommand{\Eor}{E^{\typeor}_{S}}

\newcommand{\Gand}{G^{\typeand}_{S}}
\newcommand{\Gor}{G^{\typeor}_{S}}

\newcommand{\omegaandR}{\Omega^{\typeand}_{\textnormal{removed}}}
\newcommand{\omegaorR}{\Omega^{\typeor}_{\textnormal{removed}}}

\newcommand{\omegaandP}{\Omega^{\typeand}_{\textnormal{preserved}}}
\newcommand{\omegaorP}{\Omega^{\typeor}_{\textnormal{preserved}}}

\newcommand{\order}{\operatorname{order}}
\newcommand{\m}{\operatorname{m}}

\newcommand{\dpos}{{\bf I}^{(\m),+}}
\newcommand{\dneg}{{\bf I}^{(\m),-}}

\newcommand{\dposR}{{\bf I}^{(\m),+}_{\textnormal{removed}}}
\newcommand{\dnegR}{{\bf I}^{(\m),-}_{\textnormal{removed}}}

\newcommand{\dposP}{{\bf I}^{(\m),+}_{\textnormal{preserved}}}
\newcommand{\dnegP}{{\bf I}^{(\m),-}_{\textnormal{preserved}}}

\newcommand{\triggerand}{\mathbbm{1}_{\operatorname{AND}}(S\!\mid\! \x')}
\newcommand{\triggeror}{\mathbbm{1}_{\operatorname{OR}}(S\!\mid\! \x')}

\newcommand{\triggerS}{\mathbbm{1}(\scalebox{0.85}{$|S|\!=\!\m$})\!\cdot\!}

\newcommand{\rhol}{\rho_\low}
\newcommand{\rhoh}{\rho_\high}

\newcommand{\conditionl}{1 \!\leq\! |S| \!\leq\! 3}
\newcommand{\conditionh}{4 \!\leq\! |S| \!\leq\! n}

\newcommand{\M}{\mathcal{M}}

\newcommand{\loss}{L}

\newcommand{\iand}{I_S^{\typeand}}
\newcommand{\ior}{I_S^{\typeor}}

\newcommand{\iandp}{\tilde{I}_S^{\typeand}}

\newcommand{\reference}{\textnormal{*}}
\newcommand{\vref}{v^\reference}

\newcommand{\iandref}{I_S^{\typeand,\reference}}
\newcommand{\iorref}{I_S^{\typeor,\reference}}

\newcommand{\omegaandref}{\Omega^{\typeand,\reference}}
\newcommand{\omegaorref}{\Omega^{\typeor,\reference}}

\newcommand{\omegaandG}{\Omega^{\typeand}_\textnormal{general}}
\newcommand{\omegaorG}{\Omega^{\typeor}_\textnormal{general}}
\newcommand{\dposG}{{\bf I}^{(\m),+}_{\textnormal{general}}}
\newcommand{\dnegG}{{\bf I}^{(\m),-}_{\textnormal{general}}}

\newcommand{\rhoR}{\rho_\textnormal{removed}}

\newcommand{\fnmark}[1]{\textsuperscript{\textnormal{\ref{#1}}}}

\newtcolorbox{myquote}{
    colback=white,
    enhanced,
    breakable,
    sharp corners,
    boxrule=0pt,
    leftrule=3pt,
    colframe=gray!30,
    left=8pt,
    right=0pt,
    top=2pt,
    bottom=2pt
}

\begin{document}

\maketitle

\begin{abstract}
This study focuses on the scientific problem of understanding internal factors that govern the diverse performance degradation of deep neural networks (DNNs) when different parameters are pruned.
In order to explain why pruning certain parameters leads to significant performance degradation but pruning other parameters does not, we examine how the pruning operation affects the interaction patterns encoded by the DNN.
We find that when we progressively increase the pruning ratio, the interaction patterns encoded by DNNs exhibit a distinct three-phase dynamics, \emph{i.e.}, model performance is not largely affected until the pruning operation begins to remove low-order interactions, and low-order interactions exhibit strong generalizability.
Moreover, we find that the high sensitivity of DNN performance to the pruning of certain modules is attributed to whether the pruning operation removes generalizable low-order interaction patterns.
\end{abstract}

\section{Introduction}

Many studies on neural network compression~\citep{han2016deep,fang2023depgraph} have long been centered on the trade off between model performance and parameter efficiency.
However, in this study, we do not develop a new compression algorithm.
Instead, we hope to explain the effectiveness of model compression, which moves beyond superficial explanations, such as the importance of the pruned parameters. We focus on a new scientific problem, \emph{i.e.}, \textit{what internal factors lead to performance degradation in a compressed deep neural network (DNN).}

In this study, we no longer regard the performance degradation of a DNN as an indivisible property. We conduct an exploratory attempt towards the new question, \emph{i.e.}, is it plausible to quantitatively examine and interpret the explicit changes in inference patterns encoded by compressed DNNs?

To this end, we are inspired by the recent proof~\citep{ren2024proving, chen2024defining} that the complex prediction logic of DNNs can be explained by a small number of interaction patterns.
We adopt parameter pruning as a representative lens, and \textbf{find that the changes in the generalizability of interaction patterns modeled in pruned DNNs can well explain why pruning different parameters leads to diverse performance degradation.}


\begin{figure}[!t]
  	\centering
	\includegraphics[width=\textwidth]{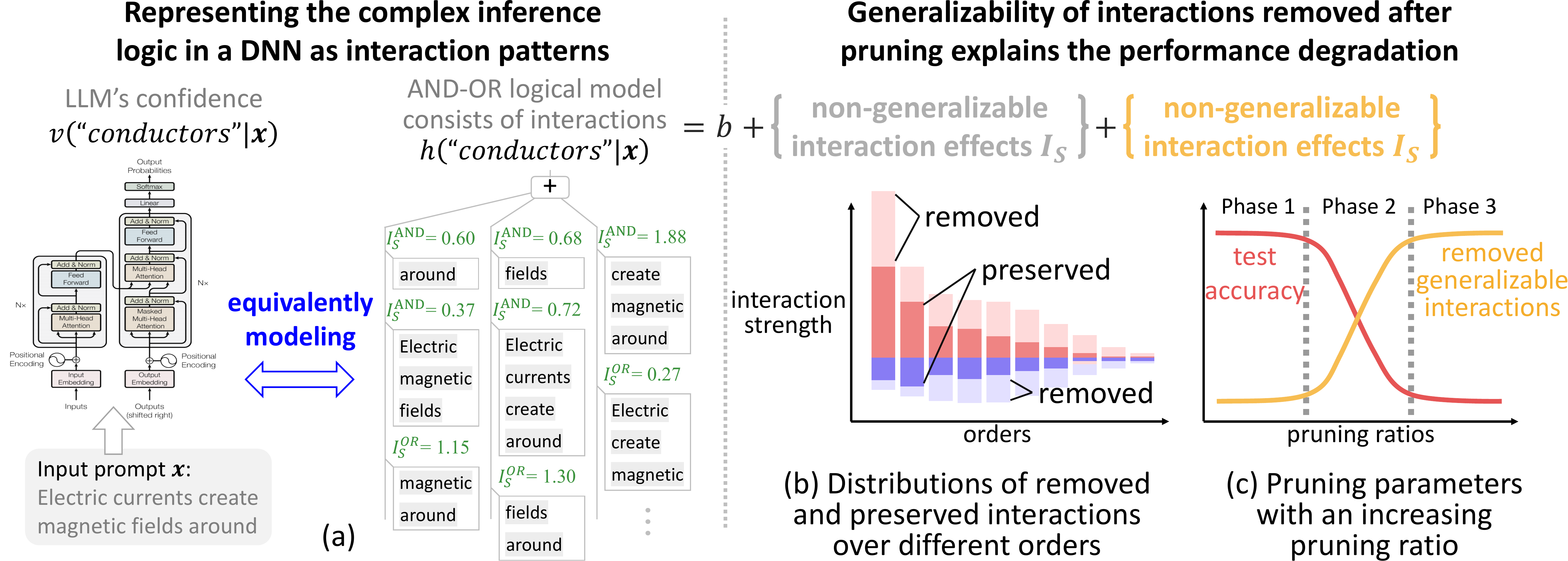}
  	\caption{(a) It is proven \citep{ren2024proving} that the complex prediction logic can be mathematically decomposed into a small set of AND/OR interactions. (b) Some interactions are removed (shown in light color), while others are preserved (shown in dark color) after pruning. (c) When we progressively increase the pruning ratio, we find that the test accuracy does not decrease noticeably until the pruning operation begins to remove generalizable interactions.
	}
  	\label{fig:main}
\end{figure}

\begin{myquote}
As the background, many theoretical studies~\citep{ren2024proving, chen2024defining} and empirical evidence~\citep{li2023does, zhou2024explaining} show that the output score of a DNN can be faithfully decomposed into the numerical effects of a small set of interactions.
As Figure \ref{fig:main} shows, an interaction represents a nonlinear relationship among input variables encoded by a DNN. For example, given a prompt describing electromagnetic induction, a large language model (LLM) encodes the phrase \{``create'', ``magnetic'', ``around''\} as an interaction, and its numerical effect can be explicitly quantified. Specifically, the interaction contributes $1.88$ to increasing the confidence of predicting the next token ``conductors''. Similarly, in vision models, interactions represent nonlinear visual patterns among multiple image patches encoded by the model.
\end{myquote}

Motivated by these findings, in this paper, we follow~\citep{he2025generalizability} to distinguish generalizable interactions and non-generalizable interactions, such that we can investigate how parameter pruning gradually affects the generalizability of interactions encoded by a DNN. Our findings can be summarized as follows.

(1) We progressively increase the pruning ratio to prune more parameters from a DNN and measure the changes in interactions during the pruning process.
We discover a distinct three-phase dynamics of interactions as the pruning ratio progressively increases.
Under low pruning ratios, the pruning operation mainly removes high-order interactions (\emph{i.e.}, complex interactions), and DNN performance remains largely unaffected. Under high pruning ratios, the pruning operation starts to remove low-order (simple) interactions, and DNN performance degrades significantly. These results indicate that the removal of low-order interactions is highly related to DNN performance degradation.

(2) The above findings can be extended to more general pruning scenarios. When we prune insensitive modules (i.e., those do not significantly degrade DNN performance), only high-order interactions are removed. In contrast, when we prune sensitive modules that degrade DNN performance, low-order interactions start to be significantly removed.

(3) In addition, we find that low-order interactions exhibit stronger generalizability than high-order interactions. Moreover, about a half of the high-order interactions exhibit positive effects, while the other half exhibit negative effects. The strong offsetting effects indicate that high-order interactions act like noise patterns, compared to low-order interactions.

\section{Method}

\subsection{Preliminaries: interactions}
\label{subsec:preliminaries}

Let $v(\cdot)$ denote a scalar function of a DNN, and let $\x=[x_1,\cdots,x_n]^T$ denote an input sample, where $x_i$ represents the input variable at index $i$\footnote{
\label{fn:variable}Following the experimental settings in~\citep{ren2024proving}, for the image classification task, a variable is defined as an image patch, while for the language modeling task, a variable is defined as one or more tokens that correspond to a word.
}, with the variable index set $N=\{1,2,\cdots,n\}$. The scalar function $v(\cdot)$ can be defined in different ways. For classification tasks, it is typically defined as the prediction confidence score \citep{ren2024proving,deng2022discovering}:

\begin{equation}
	\label{eq:v}
	v(\x)\stackrel{\text{def}}{=}\log\frac{p(y=\y \mid \x)}{1-p(y=\y \mid \x)}
\end{equation}

where $p(y = \y|\x)$ denotes the predicted classification probability of the ground-truth category $\y$.

We can construct the following logical function $g(\cdot)$. It is proven that $g(\cdot)$ can well match the shape of the DNN function $v(\cdot)$, and is therefore considered as a faithful surrogate function of the DNN~\citep{ren2024proving, chen2024defining}.

\begin{equation}
\label{eq:logic}
\begin{aligned}
&\forall \x' \in \Psi,\quad |v(\x') - g(\x')| < \epsilon, \\
\text{s.t.}\quad &g(\x') \overset{\text{def}}{=} 
\sum_{S \in \omegaand} \underbrace{\iand \cdot \triggerand}_{\text{an AND interaction } S}
+
\sum_{S \in \omegaor} \underbrace{\ior \cdot \triggeror}_{\text{an OR interaction } S}
+ b.
\end{aligned}
\end{equation}

where a tiny scalar $\epsilon$ ensures the fidelity of using the logical function $g(\cdot)$ to approximate the DNN function $v(\cdot)$ on all $2^n$ masked states of the input $\x$ in $\Psi=\{\x_S:S\subseteq N\}$.
In each masked state $\x_S$, only the variables in $S$ are retained, while other variables in $N\setminus S$ are masked\footnote{\label{fn:mask}The masking of the $i$-th variables is implemented as setting $x_i$ to a baseline values. Please see Appendix~\ref{subsec:apdx-interaction-settings} for details.}.
$\omegaand$ and $\omegaor$ denote the set of AND interactions and the set of OR interactions, respectively. Scalar weights $\iand$ and $\ior$, and a scalar bias $b$ are learned using the method of~\citep{chen2024defining} \textit{(please see Appendix~\ref{sec:apdx-optimize-pq} for details)}.

The binary trigger function $\triggerand$ detects the logical AND relationship over a subset $S\subseteq N$ of input variables.
It returns 1 if all variables in $S$ are present (\emph{i.e.}, not masked in $\x'$)\fnmark{fn:mask}, and 0 otherwise.
The binary trigger function $\triggeror$ detects the logical OR relationship over a subset $S\subseteq N$ of input variables.
It returns 1 if any variable in $S$ is present (\emph{i.e.}, not masked in $\x'$)\fnmark{fn:mask}, and 0 otherwise.

The fidelity of using the logical function $g(\cdot)$ to explain $v(\cdot)$ has been theoretically proven~\citep{chen2024defining} (\textit{see Appendix~\ref{proof:apdx-match} for details}), and supported by extensive empirical evidence~\citep{li2023does, zhou2024explaining, cheng2025revisiting}.

\color{black}

\textbf{Sparsity of interactions.} It is proven that the number of interactions encoded by a DNN for a given input is theoretically bounded~\cite{ren2024proving} and empirically observed to be relatively small (around 50-150).
\textit{Theoretical proofs and empirical validation are provided in Appendix~\ref{sec:apdx-condition-for-sparsity} and Appendix~\ref{sec:apdx-exp-sparsity}, respectively.}

\textit{A video demo of interaction-based explanations is provided in the supplementary materials.}

\subsection{Quantifying the changes in interactions during progressive parameter pruning}
\label{subsec:quantification}

In this study, we aim to move beyond the conventional studies~\citep{li2018visualizing, bartlett2017spectrally} that treat the DNN's generalization ability as a holistic property. Instead, we seek to investigate the underlying mechanisms that determine the generalization ability of DNNs.

To this end, we adopt parameter pruning as a representative lens to study the intricate mechanisms underlying changes in DNN generalization, beyond the broad empirical observation that pruning more parameters leads to more significant performance degradation. Why does pruning some parameters lead to a sharp drop in model performance, but pruning other parameters does not? This question requires a more fine-grained perspective.

\begin{myquote}
To this end, as mentioned in Section \ref{subsec:preliminaries}, \citet{ren2024proving} have proven that: (1) the DNN's output score for a given input can be faithfully decomposed into only 50-150 interactions, and (2) these interactions can well match the DNN's complex inference logic for this input with theoretically guaranteed high fidelity.
\end{myquote}

Therefore, we treat interactions as the primitive inference patterns that directly affect the generalization ability of a DNN. And we design a set of metrics to measure the changes in interactions caused by parameter pruning, so that \textit{we can establish a correlation between the changes in interactions and the changes in model performance after different pruning operations.}

\textbf{Preserved, removed, and emerged interactions after parameter pruning.\label{para:changes}}
We categorize the changed interactions after parameter pruning into two distinct types: \textit{removed interactions} and \textit{emerged interactions}. Let $\omegaand$ and $\omegaor$ denote the sets of AND interactions and OR interactions encoded by the original DNN $v(\cdot)$, respectively. Similarly, let $\omegaandp$ and $\omegaorp$ denote the sets of interactions encoded by the pruned DNN $\p(\cdot)$, whose parameters are pruned at a pruning ratio\footnote{The pruning ratio $\alpha$ is defined as the percentage of parameters that are pruned.\label{fn:pruning-ratio}} of $\alpha$.

\begin{figure}[!t]
  	\centering
	\includegraphics[width=0.9\textwidth]{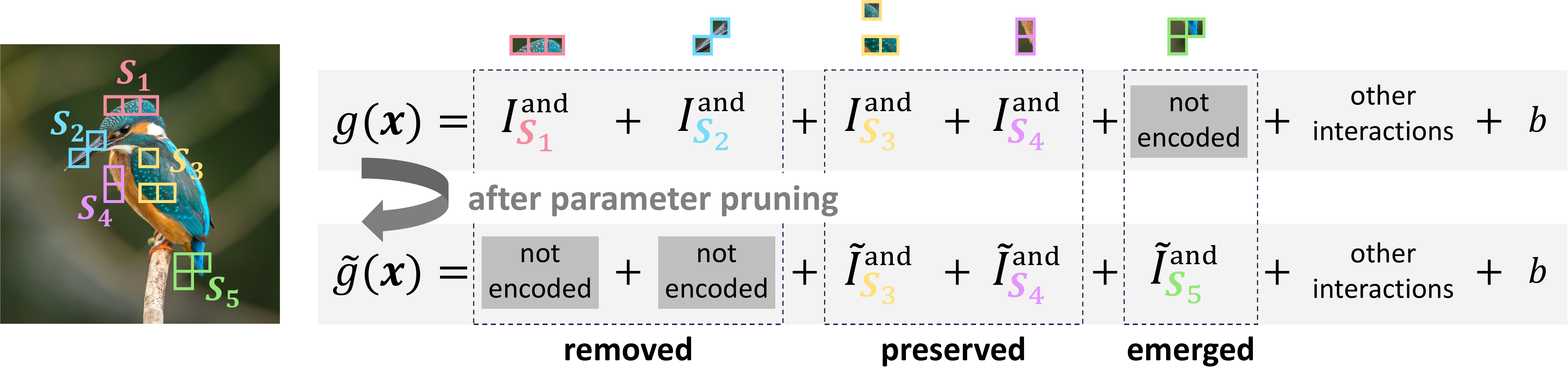}
  	\caption{Illustration of removed, preserved, and emerged interactions after pruning.}
  	\label{fig:changes}
\end{figure}

\begin{definition}
\label{def:changes}
For an AND interaction $S$ encoded by the original DNN $v(\cdot)$ for the input $\x$ ($S \in \omegaand$), if this interaction is also encoded by the pruned DNN $\p(\cdot)$ ($S \in \omegaandp$) and produces a consistent effect (\emph{i.e.,} {\small $\sign(\iand)=\sign(\iandp)$}), then this interaction is considered to be preserved after the pruning operation; otherwise, it is considered to be removed.
For an AND interaction $S$ encoded by the pruned DNN $\p(\cdot)$ for the input $\x$ ($S \in \omegaandp$), if this interaction is not a preserved interaction in the original DNN $v(\cdot)$, then this interaction is considered to have emerged during the pruning operation.
Thus, we can use the following binary metrics $\Pand$, $\Rand$ and $\Eand$ to identify the preserved, removed and emerged AND interactions after the pruning operation, respectively.

\begin{equation}
	\label{eq:changes}
	\begin{aligned}
	\Pand &= \mathbbm{1}(S\in\omegaand) \cdot \mathbbm{1}(S\in\omegaandp) \cdot \mathbbm{1}(\sign(\iand)=\sign(\iandp)), \\
	\Rand &= \mathbbm{1}(S\in\omegaand) \cdot \mathbbm{1}(\Pand=0), \\
	\Eand &= \mathbbm{1}(S\in\omegaandp) \cdot \mathbbm{1}(\Pand=0).
	\end{aligned}
\end{equation}

In the way, we obtain the relations {$\omegaand\!\!=\{S: \Pand\cdot\Rand\!\neq 0\}$} and {$\omegaandp\!\!=\{S: \Pand\cdot\Eand\!\neq 0\}$}. Metrics for the preserved, removed, emerged OR interactions, $\Por$, $\Ror$ and $\Eor$, are defined similarly.
\end{definition}

\begin{myquote}
	For example, as Figure \ref{fig:changes} shows, the interactions $S_1$ and $S_2$ are encoded only by the original DNN $v(\cdot)$ and is removed after the pruning operation. The interactions $S_3$ and $S_4$ are encoded by both the original DNN $v(\cdot)$ and the pruned DNN $\p(\cdot)$, and is preserved after the pruning operation. The interaction $S_5$ is encoded only by the pruned DNN $\p(\cdot)$, and has emerged during the pruning operation.
\end{myquote}


In this way, interactions provide a new perspective for understanding DNN's generalization ability. We can naturally consider that interactions that emerge after parameter pruning (without training) often represent non-generalizable noise patterns.
On the other hand, if many generalizable interactions are removed after pruning, then the DNN generalization is supposed to be significantly affected.


\textbf{Distribution of interactions over different orders (complexities).\label{para:complexity}}
To further characterize which interactions are more likely to be generalizable, we introduce interaction complexity as a key structural metric.
It is because \citet{zhou2024explaining, liu2023towards} have found that \textbf{complex interactions are less generalizable and more perturbation-sensitive than simple interactions, which will be verified in later experiments} in Section \ref{subsec:generization}.


Thus, we define the complexity of a single interaction as its order, given as $\order(S) = |S|$. The order of an interaction represents the number of variables involved in the interaction.
Furthermore, we define and quantify the distribution of interaction effects over different orders in Definition \ref{def:distribution}, so as to represent the overall complexity of all interactions encoded by the DNN.

\begin{definition}
\label{def:distribution}

\textbf{(Distribution of interactions over different orders)} Given the sets of AND/OR interactions $\omegaand$ and $\omegaor$, the distribution of interactions is represented by the total strength of all positive interactions over all the $n$ orders {\small $[{\bf I}^{(1),+}, {\bf I}^{(2),+},\ldots,{\bf I}^{(\operatorname{n}),+}]^T$}, and the total strength of all negative interactions over all the $n$ orders {\small $[{\bf I}^{(1),-}, {\bf I}^{(2),-},\ldots,{\bf I}^{(\operatorname{n}),-}]^T$}. The interaction strength of each $m$-th order is quantified as follows~\cite{he2025generalizability}.

\begin{equation}
	\label{eq:distribution}
	\begin{aligned}
	\dpos(\omegaand, \omegaor)&=\sum_{S \in\omegaand}{\triggerS \max(\iand,0)}+\sum_{S \in\omegaor}{\triggerS \max(\ior,0)}, \\
	\dneg(\omegaand, \omegaor)&=\sum_{S \in\omegaand}{\triggerS \min(\iand,0)}+\sum_{S \in\omegaor}{\triggerS \min(\ior,0)}.
	\end{aligned}
\end{equation}
\end{definition}

Thus, we can further use this definition to quantify the distribution of the removed interactions ({\small $\dposR = \dpos(\omegaandR, \omegaorR)$}, {\small $\dnegR = \dneg(\omegaandR, \omegaorR)$}), and the preserved interactions ({\small $\dposP = \dpos(\omegaandP, \omegaorP)$}, {\small $\dnegP = \dneg(\omegaandP, \omegaorP)$}), where the interaction sets ({\small $\omegaandR$}, {\small $\omegaorR$}) and ({\small $\omegaandP$}, {\small $\omegaorP$}) are determined by the metrics of {\small $\Rand$}, {\small $\Ror$} and {\small $\Pand$}, {\small $\Por$}.\footnote{
	${\omegaandR\!\!=\{S : R^{\typeand}_S\!\!=1\}},\ 
	{\omegaorR\!\!=\{S : R^{\typeor}_S\!\!=1\}};\ 
	{\omegaandP\!\!=\{S : P^{\typeand}_S\!\!=1\}},\
	{\omegaorP\!\!=\{S : P^{\typeor}_S\!\!=1\}}.$
}

\begin{figure}[!t]
  	\centering
  	\includegraphics[width=\textwidth]{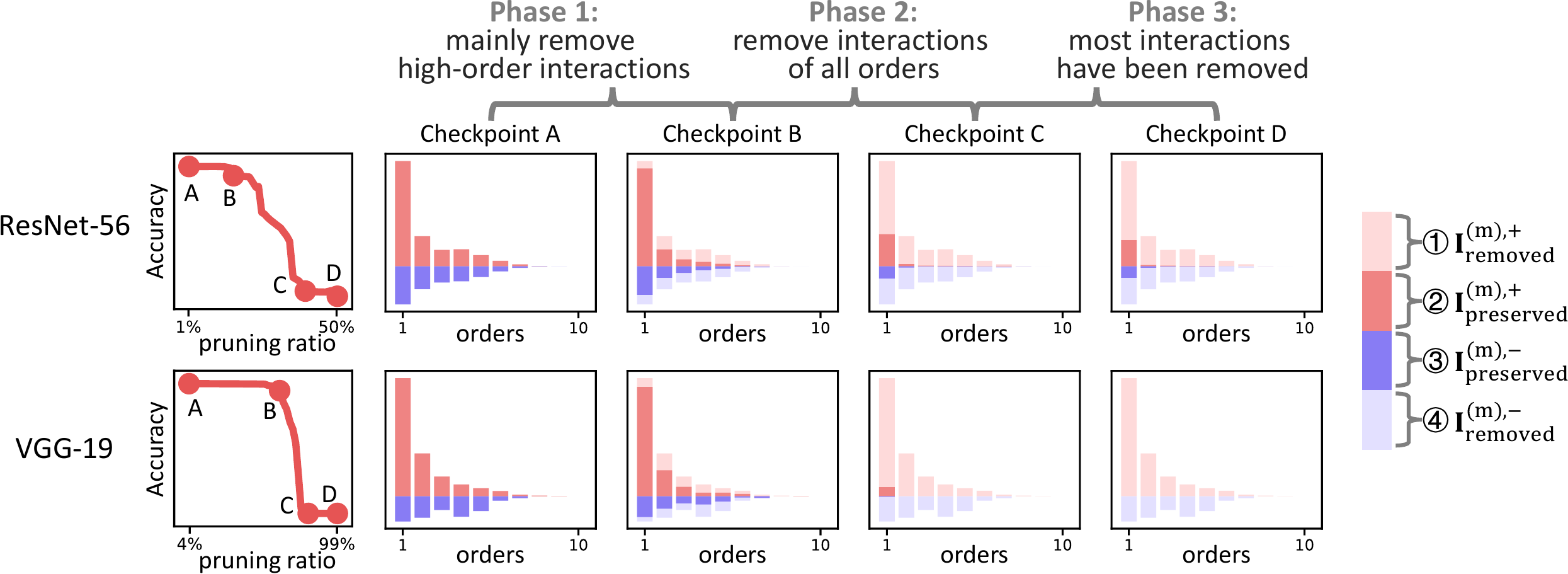}
  	\caption{Changes in the distributions of removed interaction ({\small$\dposR$} and {\small$\dnegR$}) and the distributions of preserved interaction ({\small$\dposP$} and {\small$\dnegP$}) when the pruning ratio increases. These changes can be divided into three phases. The four selected checkpoint pruning ratios correspond to the inflection points of the accuracy curve, and represent the boundaries of the three phases in the progressive parameter pruning process. \textit{Please see Appendix \ref{subsec:apdx-distribution-three-phase} for results on more pruning ratios.}
	}
  	\label{fig:phases}
\end{figure}

\textbf{Metric 1: Ratio of preserved low/high-order interactions.}
Considering the strong connection between the complexity and generalizability of interactions~\citep{zhou2024explaining, liu2023towards} (verified in Section~\ref{subsec:generization}), we measure the ratio of low-order ($1$-st to $3$-rd order) interactions that are preserved after pruning, denoted by $\rhol$, and the preserved ratio of high-order ($4$-th to $n$-th order) interactions $\rhoh$.

\begin{definition}
\label{def:rho}
\textbf{(Ratio of preserved interactions)}
The ratio of preserved low-order interactions to all low-order interactions $\rhol$ and the ratio of preserved high-order interactions to all high-order interactions $\rhoh$ are defined as follows.

\begin{equation}
	\label{eq:rho}
	\begin{aligned} 
	\rhol&=\frac
		{
			\sum_{S\in\omegaand:\conditionl}{|\iand| \!\cdot\! \Pand} + 
			\sum_{S\in\omegaor:\conditionl}{|\ior| \!\cdot\! \Por}
		}
		{
			\sum_{S\in\omegaand:\conditionl}{|\iand|} + 
			\sum_{S\in\omegaor:\conditionl}{|\ior|}
		} \\
	\rhoh&=\frac
		{
			\sum_{S\in\omegaand:\conditionh}{|\iand| \!\cdot\! \Pand} + 
			\sum_{S\in\omegaor:\conditionh}{|\ior| \!\cdot\! \Por}
		}
		{
			\sum_{S\in\omegaand:\conditionh}{|\iand|} + 
			\sum_{S\in\omegaor:\conditionh}{|\ior|}
		}
	\end{aligned}
\end{equation}
\end{definition}

Based on the interaction generalizability across different orders, we define interactions of orders 1-3 with high generalizability as low-order interactions, and interactions of other orders as high-order interactions. \textit{Please see Section~\ref{subsec:generization} and Appendix~\ref{sec:apdx-generalizable-distribution} for detailed discussions.}


\textbf{Metric 2: offsetting ratio of the removed interaction effects.\label{para:offsetting}}
The offsetting ratio of interaction effects is another typical metric to evaluate the representation efficiency of a DNN.
As Figure~\ref{fig:removed-interactions} shows, given an input sample, if lots of removed positive interaction effects and negative interaction effects offset each other, then such removed interactions can be considered as redundant noise-like patterns without producing a clear classification signal.
Therefore, we measure the ratio of offsetting interaction effects $\offset$ among all removed interaction effects to evaluate the changes in DNN representation quality after the pruning operation.

\begin{definition}
\label{def:offsetting}
\textbf{(Offsetting ratio of removed interaction effects)} Given the sets of AND/OR interactions $\omegaand$ and $\omegaor$, the offsetting ratio of all removed interactions $\offset$ is defined as follows~\cite{he2025generalizability}.

\begin{equation}
	\label{eq:offsetting}
	\offset = 1 - \frac{
		|\sum\nolimits_{S\in\omegaand}{\iand\cdot\Rand}+\sum\nolimits_{S\in\omegaor}{\ior\cdot\Ror}|
	}{
		\sum\nolimits_{S\in\omegaand}{|\iand\cdot\Rand|}+\sum\nolimits_{S\in\omegaor}{|\ior\cdot\Ror|}
	}
\end{equation}
\end{definition}

Additionally, the high-order interactions that are removed during progressive parameter pruning often exhibit a spindle-shaped distribution over different orders and greatly offset each other. Specifically, a half of removed high-order interactions produce positive effects and boost the classification score, while the other half produce negative effects and reduce the classification score.


\subsection{Three-phase dynamics of interactions during progressive pruning}
\label{subsec:three-phase}

\begin{figure}[!t]
  	\centering
  	\includegraphics[width=\textwidth]{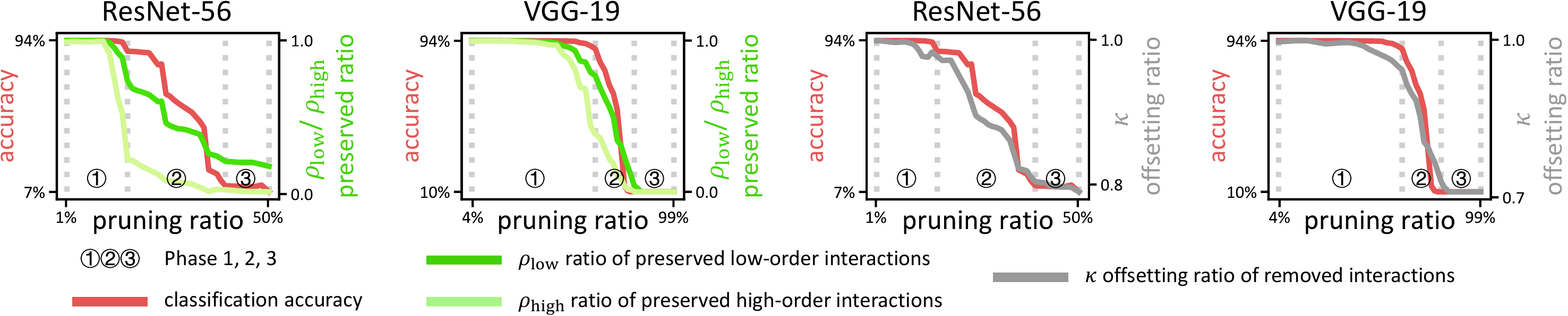}
  	\caption{(left) Changes in the ratio of preserved low/high-order interactions $\rhol$/$\rhoh$ and the classification accuracy when the pruning ratio increases. (right) Changes in the offsetting ratio of removed interactions $\offset$ and the classification accuracy when the pruning ratio increases.}
  	\label{fig:phases-rho}
\end{figure}

In order to explore the mechanistic interactions that cause generalization degradation, we use the metrics proposed in the previous section to continuously track and evaluate the changes in interactions when we progressively increase the pruning ratio.

We progressively increase the pruning ratio (from {\small 1.2\%} to {\small 50.2\%} for ResNet-56~\citep{resnet} and from {\small 3.6\%} to {\small 99.2\%} for VGG-19~\citep{vgg}), thereby incrementally removing more parameters to examine how parameter pruning degrade model performance.
Without loss of generality, we apply the generic method DepGraph~\citep{fang2023depgraph} to prune ResNet-56 and VGG-19 for image classification on the CIFAR-10 dataset~\citep{cifar}. \textit{Please see Appendix~\ref{subsec:apdx-pruning-setting-three-phase} for detailed settings.} In Section \ref{subsec:further-validation}, we will further verify our findings under other generic pruning settings.

\color{black}

In this way, we analyze how the ratio of preserved low-order/high-order interactions and the offsetting ratio of removed interactions are related to the classification accuracy.
Experimental results are mainly shown in the following three figures.
(1) Figure \ref{fig:phases} compares the changes in the distribution of all removed interactions (denoted by {\small $\dposR$} and {\small $\dnegR$}) and the distribution of all preserved interactions (denoted by {\small $\dposP$} and {\small $\dnegP$}) as the pruning ratio progressively increases.
(2) Figure \ref{fig:phases-rho} (left) compares the curve of classification accuracy, the curve of preserved low-order interaction ratio $\rhol$, and the curve of preserved high-order interaction ratio $\rhoh$. Besides, we also measure the offsetting ratio of the removed interactions $\offset$ when we progressively increase the pruning ratio.
(3) Figure \ref{fig:phases-rho} (right) shows the curve of offsetting ratio $\kappa$ after progressive parameter pruning.

\textbf{Three-phase dynamics.}

Just as the observation of three-phase interaction dynamics during the model-training process~\citep{he2025generalizability}, we find that the evolution of interactions during progressive parameter pruning can also be divided into three phases, although these interaction dynamics are measured under two fundamentally different experimental settings.


\textbullet\ \textit{Phase 1:} When the pruning ratio is low (below {\small 14.6\%} for ResNet-56 and below {\small 59.8\%} for VGG-19), pruning operations do not affect model performance (the accuracy remains around {\small 92.1\%}). Most removed interactions are high-order interactions (over a half of the high-order interactions are removed in this phase, \emph{i.e.}, {\small $\rhoh=0.3\pm 0.1$} while {\small $\rhol= 0.82\pm 0.03$}) and exhibit strong offsetting effects (nearly all removed interactions offset each other, \emph{i.e.}, {\small $\offset=0.96\pm 0.02$}). These offsetting interactions act like noise patterns and therefore are considered not generalizable.

\textbullet\ \textit{Phase 2:} When the pruning ratio increases (to {\small 39.7\%} for ResNet-56 and to {\small 77.4\%} for VGG-19), model performance starts to degrade (the accuracy drops to around {\small 10.3\%}). Low-order interactions start to be noticeably removed (over {\small 4/5} of the low-order interactions are removed in this phase, \emph{i.e.}, {\small $\rhoh= 0.01\pm0.01$} and {\small $\rhol= 0.1\pm0.1$} after this phase). And the removed interactions exhibit poorer offsetting effects than in the first phase (around {\small 3/4} of the removed interactions offset each other, \emph{i.e.}, {\small $\offset=0.75\pm0.05$}). It means that non-offsetting interactions, which actually contribute to the final prediction, start to be removed in this phase.

\textbullet\ \textit{Phase 3:} When the pruning ratio is even higher (to {\small 50.2\%} for ResNet-56 and to {\small 99.2\%} for VGG-19), model performance remains extremely low under excessive parameter pruning (the accuracy remains around {\small 10.0\%}). Both low-order and high-order interactions show only minor changes (nearly all interactions have already been removed, \emph{i.e.}, {\small $\rhoh= 0.01\pm0.01$} and {\small $\rhol= 0.1\pm0.1$}).

Note that the removal of the highly offsetting high-order interactions usually does not affect the test accuracy, which is supported by existing studies on interaction-based explanations: \citet{zhou2024explaining} have found that high-order interactions are less generalizable to testing samples than low-order interactions. And \citet{liu2023towards} have found that high-order interactions are more sensitive to feature perturbations. The poor generalizability of high-order interactions is also observed in Section~\ref{subsec:generization}.

In sum, when the pruning ratio is low (during Phase 1), parameter pruning mainly removes high-order interactions, which barely affects the performance of DNNs. In contrast, when the pruning ratio is high (during Phases 2 and 3), low-order interactions start to be removed, which explains the performance degradation of DNNs. Table~\ref{tab:phases} shows the analytical results of the three-phase dynamics from multiple perspectives. \textit{Detailed results on more pruning ratios can be found in Appendix~\ref{sec:apdx-three-phase-details}.}

\subsection{Specificity of removed interactions under generic pruning settings}
\label{subsec:further-validation}

\begin{figure}[!t]
  	\centering
  	\includegraphics[width=\textwidth]{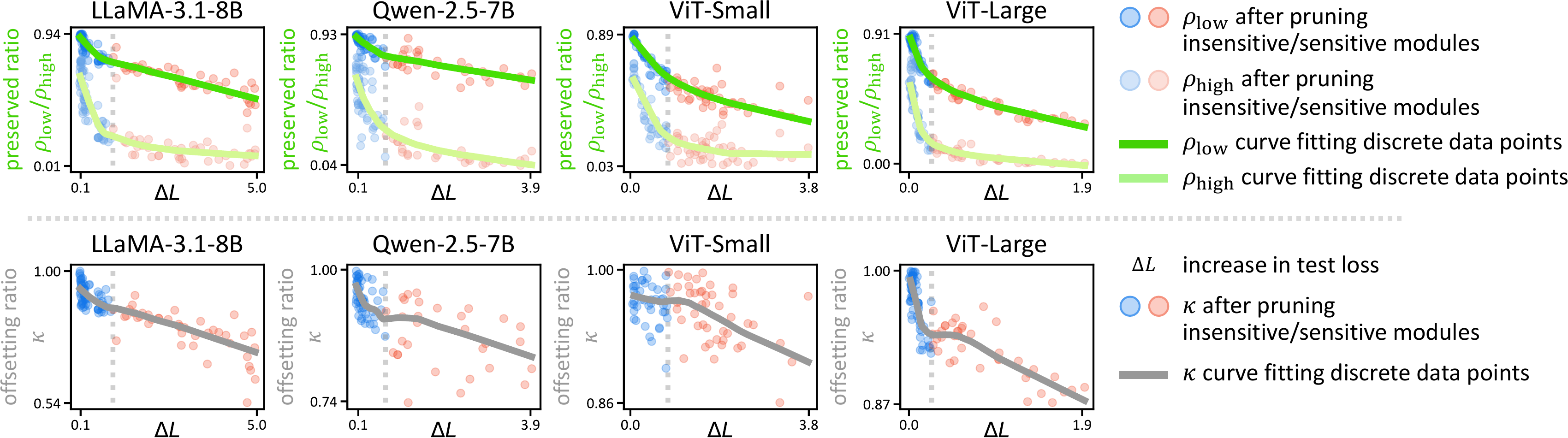}
  	\caption{(top) The ratio of preserved low/high-order interactions $\rhol$/$\rhoh$ after pruning each module. (bottom) The offsetting ratio of removed interactions $\offset$ after pruning.
	After pruning sensitive modules, low-order interactions are significantly removed, and the removed interactions exhibit poorer offsetting ratio.
	\textit{Please see Appendix~\ref{subsec:apdx-metric-module-pruning-full} for more results.}
	}
  	\label{fig:dynamics-module}
\end{figure}

In the previous subsection, we have identified a strong correlation between DNN performance degradation and the changes in interactions by using a classical parameter pruning method~\citep{fang2023depgraph}. However, compared to smaller models, large language models (LLMs) are usually much more sensitive to even mild pruning.\footnote{This also applies to other transformer-based models, for which no well-established pruning algorithm can remove parameters without affecting performance in the absence of post-hoc finetuning.}
In particular, the pruning strategies adopted by state-of-the-art LLM compression methods~\citep{frantar2023sparsegpt, sun2023wanda, ma2023llmpruner} are often coupled with post-hoc finetuning to recover performance.
Joint analysis of parameter pruning and post-hoc finetuning significantly complicates the narrative of this study.

Therefore, we extend our empirical studies to more general parameter-pruning settings, instead of state-of-the-art methods, and conduct experiments on more advanced and complex models. Without loss of generality, we apply two typical module-level pruning settings: (1) pruning a single residual block, and (2) pruning the hidden feature dimensions.
\textit{Detailed experimental settings are provided in Apendix~\ref{subsec:apdx-pruning-setting-module}.}


\textbf{Sensitive modules vs. insensitive modules.} Based on the above module settings, we categorize all pruned modules into two types: (1) sensitive modules, and (2) insensitive modules, according to the increase in test cross-entropy (CE) loss caused by pruning. Specifically, if pruning a module leads to a test CE loss increase greater than a predefined threshold $\tau$, the corresponding module is regarded as sensitive; otherwise, it is regarded as insensitive. In general, the threshold $\tau$ is empirical chosen to ensure that an increase in test CE loss greater than $\tau$ indicates substantial performance degradation. \textit{The exact values are reported in the Appendix~\ref{subsec:apdx-pruning-setting-module}.}



\textbf{Distinct changes in interactions after pruning sensitive modules and insensitive modules.}
We find that the pruning of sensitive modules and that of insensitive modules drives interaction representations in fully different directions, in terms of both interaction complexity and the offsetting ratio of interactions, thereby explaining the root cause of DNN performance degradation after pruning.
Specifically, for each pruned module {\small $\M$}, we measure the changes in interactions after pruning, including the ratios of preserved low-order interactions and high-order interactions (denoted by $\rhol$ and $\rhoh$ , respectively), and the offsetting ratio of removed interactions (denoted by $\offset$).

In Figure~\ref{fig:dynamics-module}, each dot represents the effect of pruning a specific module. The horizontal axis indicates the increase in test loss caused by the pruning operation, while the vertical axis shows the ratio of preserved low/high-order interactions $\rhol$/$\rhoh$ in the top row, and the offsetting ratio of removed interactions $\offset$ in the bottom row.


\textbf{We finding that} the removal of low-order interactions leads to performance degradation. Pruning insensitive modules (blue dots on the left with relatively small $\Delta \loss$) mainly removes high-order interactions (\emph{i.e.}, {\small $\rhoh=0.15\pm0.05$} while {\small $\rhol=0.7\pm0.1$}), and the removed interactions exhibit strong offsetting effects (\emph{i.e.}, {\small $\offset=0.92\pm0.03$}).
Pruning sensitive modules (red dots on the right with relatively large $\Delta \loss$) removes both low-order and high-order interactions are significantly removed (\emph{i.e.}, {\small $\rhoh=0.1\pm0.05$} and {\small $\rhol=0.4\pm0.2$}), and the removed interactions exhibit poorer offsetting effects than pruning insensitive modules (\emph{i.e.}, {\small $\offset=0.8\pm0.1$}).
Therefore, the removal of non-offsetting low-order interactions is the direct cause of DNN performance degradation.

\color{black}

\subsection{How do removed interactions correlate with DNN generalization degradation?}
\label{subsec:generization}

\begin{figure}[!t]
  	\centering
  	\includegraphics[width=0.9\textwidth]{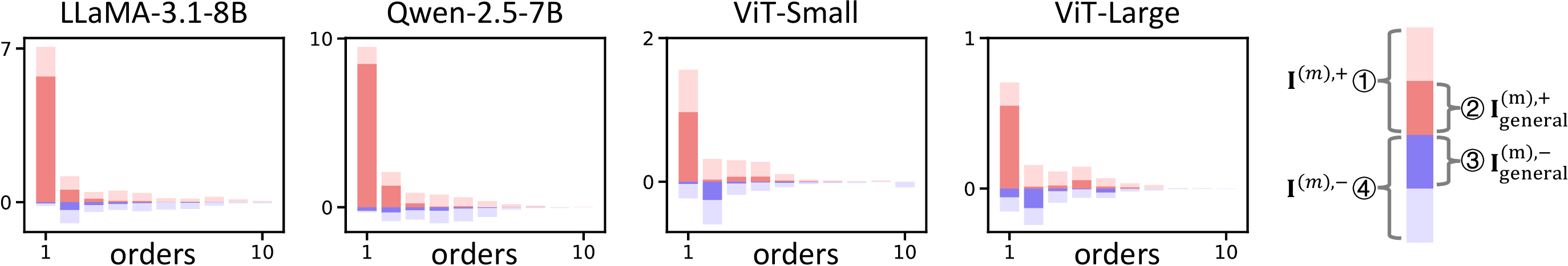}
  	\caption{The distributions of generalizable interactions (denoted by {\small $\dposG$} and {\small $\dnegG$}) and all interactions (denoted by {\small $\dpos$} and {\small $\dneg$}) over different orders. Low-order interactions exhibit stronger generalizability than high-order ones. \textit{Please see Appendix~\ref{sec:apdx-generalizable-distribution} for results on more DNNs.}}
  	\label{fig:generalizability}
\end{figure}

In the previous subsections, we have found the close connection between the changes in interactions and DNN performance degradation. Therefore, in this subsection, we aim to validate our finding by explicitly quantifying the generalizability of each removed interaction.

\textbf{Defining the generalizability of interactions.}
We follow~\citet{he2025generalizability} to define the generalizability of each interaction as follows.
Specifically, we use another DNN $\vref(\cdot)$, which is trained on testing/unseen samples and is referred to as a \textit{reference DNN}\footnote{Please see Appendix~\ref{subsec:apdx-reference} for detailed settings of reference DNNs.}.
Given an AND interaction $S$ encoded by the target DNN $v(\cdot)$ ($S\in\omegaand$), if this interaction is also encoded by the reference DNN $\vref(\cdot)$ ($S\in\omegaandref$), and produces a consistent effect (\emph{i.e.}, {\small $\sign(\iand)=\sign(\iandref)$}),\footnote{Here, {$\iandref$} denotes the interaction effect encoded by the reference DNN.} then we consider this interaction to be generalizable. The generalizable OR interactions are defined in a similar manner.
All such generalizable interactions can be considered transferable to testing samples, because all interactions encoded by the reference DNN are learned from the testing samples.
Thus, we use the following binary metrics to identify generalizable AND/OR interactions.

\begin{equation}
	\label{eq:generalization}
	\begin{aligned}
	\Gand&=\mathbbm{1}(S\in\omegaand)\cdot\mathbbm{1}(S\in\omegaandref)\cdot\mathbbm{1}(\sign(\iand)=\sign(\iandref)),\\
	\Gor&=\mathbbm{1}(S\in\omegaor)\cdot\mathbbm{1}(S\in\omegaorref)\cdot\mathbbm{1}(\sign(\ior)=\sign(\iorref))
	\end{aligned}
\end{equation}

\textbf{Verifying that low-order interactions exhibit stronger generalizability.}
Figure~\ref{fig:generalizability} shows the distribution of all interactions (denoted by {\small $\dpos$} and {\small $\dneg$} in Definition~\ref{def:distribution}) and the distribution of generalizable interactions (denoted by {\small $\dposG=\dpos(\omegaandG, \omegaorG)$} and {\small $\dnegG=\dneg(\omegaandG, \omegaorG)$} encoded by the DNN before pruning, where {\small $\omegaandG=\{S:\Gand\!\!=1\}$} and {\small $\omegaorG=\{S:\Gor\!\!=1\}$}).
We find that generalizable interactions are mainly concentrated at low orders.

\color{black}


\begin{figure}[!t]
  	\centering
  	\includegraphics[width=\textwidth]{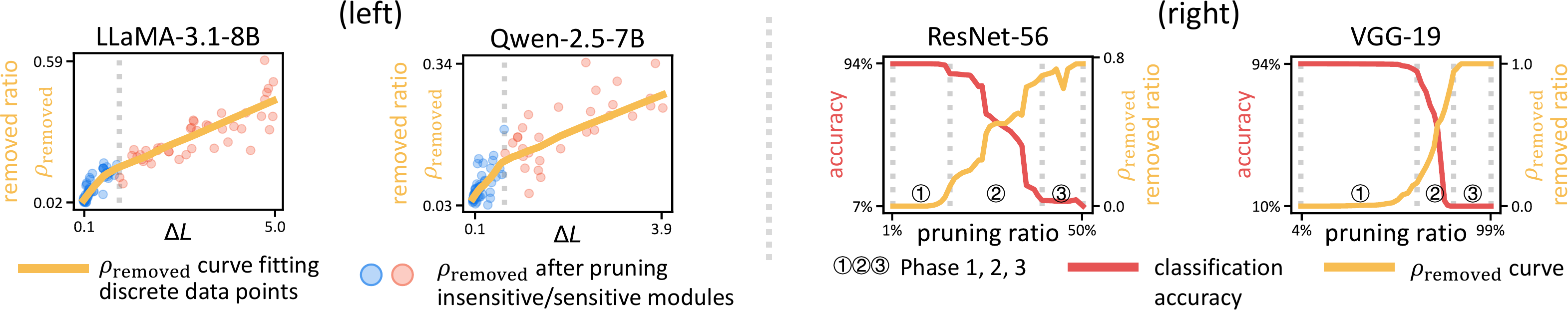}
  	\caption{(left) The ratio of removed generalizable interactions {\small $\rhoR$} after pruning each module. Pruning insensitive modules removes much less generalizable interactions than pruning sensitive modules. \textit{Please see Appendix~\ref{sec:apdx-module-pruning-details} for results on more DNNs.} (right) The ratio of removed generalizable interactions {\small $\rhoR$} when the pruning ratio progressively increases. The removal of generalizable interactions leads to DNN performance degradation.
	}
  	\label{fig:removed-generalizability}
\end{figure}

\textbf{Verifying that pruning sensitive modules removes much more generalizable interactions than pruning insensitive modules.} In order to better understand the different influence of pruning sensitive modules and pruning insensitive modules, we tentatively prune each different module {\small $\M$} from the DNN, and quantify the ratio of the generalizable interactions that are removed, denoted by {\small $\rhoR$}.

\begin{equation}
\rhoR = \frac{
	\sum_{S\in\omegaand}{\Rand\cdot\Gand\cdot\vert\iand\vert} + 
	\sum_{S\in\omegaor}{\Ror\cdot\Gor\cdot\vert\ior\vert}
	}{
	\sum_{S\in\omegaand}{\Gand\cdot\vert\iand\vert} + 
	\sum_{S\in\omegaor}{\Gor\cdot\vert\ior\vert}
	}
\end{equation}

For each DNN, the pruned modules include (1) a single residual block, and (2) the hidden feature dimensions in the DNN. \textit{Please see Appendix~\ref{subsec:apdx-pruning-setting-module} and Appendix~\ref{subsec:apdx-reference} for details about the pruned modules and the settings of reference DNNs.}

Figure~\ref{fig:removed-generalizability} (left) shows the ratio of removed generalizable interactions {\small $\rhoR$} after the pruning of each module.
Pruning insensitive modules barely removes any generalizable interactions compared to pruning sensitive modules.
Specifically, pruning insensitive modules only removes very few generalizable interactions (i.e., {\small $\rhoR= 0.1\pm0.02$}), whereas pruning sensitive modules significantly removes generalizable interactions (i.e., {\small $\rhoR= 0.35\pm0.05$}).

In addition, we extend the experiments in Section~\ref{subsec:three-phase} by measuring the ratio of removed generalizable interactions when we progressively increase the pruning ratio. As shown in Figure~\ref{fig:removed-generalizability} (right), when the pruning ratio is low (during Phase 1), both DNN performance and generalizable interactions remain unaffected after the pruning operation. When the pruning ratio is high (during Phase 2 and 3), DNN performance degrades significantly, and generalizable interactions are significantly removed.

In sum, these observations consistently indicate that the removal of generalizable interactions is a primary factor driving DNN performance degradation. We also find that generalizable interactions are mainly concentrated in low orders, which further supports the finding in previous subsections that the removal of low-order interactions is closely related to DNN performance degradation.
\section{Conclusion} \label{sec:conclusion}

In this paper, we focus on the scientific problem of identifying the underlying causes of DNN performance degradation after parameter pruning. 
Specifically, we have discovered that DNN performance degradation can be largely attributed to the removal of generalizable low-order interactions.
We validate our scientific hypothesis across twelve DNNs, including four LLMs and eight image classification models.
As a limitation, a rigorous mathematical proof explaining why low-order interactions are far more robust against pruning remains lacking.
Our findings further advance the understanding of parameter pruning and extend the boundary of existing knowledge in this area.

The validation of our scientific hypothesis is of considerable application value.
In preliminary follow-up work, we find that co-training two DNNs while penalizing non-transferable interactions between them can significantly boost the generalizability of interactions.
It enables the DNNs to maintain their performance at higher pruning ratios. \textit{Please see Appendix \ref{sec:apdx-practical-value} for details.}

\newpage
\bibliographystyle{unsrtnat}
\bibliography{arxiv.bib}

\newpage
\appendix



\section{Related work on interaction-based explanations}
\label{sec:apdx-related-interaction}

\textbf{Using interactions to explain the detailed inference logic encoded by a DNN.} Prior works on interaction-based explanations \citep{mobius, li2023does, tsai2023faith, sundararajan2020shapley} proposed to explain the inference logic of a DNN by quantifying the interactions among input variables encoded by the DNN. Building on this, \citet{ren2024proving} further proved that networks producing relatively smooth outputs across different input perturbations typically encode a small number of interactions. \citet{chen2024defining} developed an algorithm to extract interactions that were commonly encoded by multiple different neural networks. \citet{baseline} further improved a method to optimize the baseline value, which was used for interaction extraction.

\textbf{Using interactions to explain the generalization ability of a DNN.} Futhermore, interaction-based explanation research explains a DNN's generalization ability as the overall generalization ability of its compositional interactions.
\citet{zhou2024explaining} found that high-order interactions usually generalize worse to testing samples than lower-order interactions.
\citet{ren2021towards} further found that these higher-order interactions also showed poorer adversarial robustness.
\citet{ren2023bayesian} found that mean-field Bayesian neural networks typically struggled more than regular neural networks in modeling higher-order interactions.

Although interaction-based explanations offers a powerful strategy to explain the inference logic of DNNs and their generalization ability across different data domains, existing studies still leave a blank in the field of post-training interventional techniques. To this end, by investigating parameter pruning techniques, this paper bridges a crucial gap between interaction-based explanations theory and parameter pruning techniques, and offers a new perspective for understanding how interventional techniques, such as parameter pruning operations, reshape network representation quality and generalization ability.

\textbf{Comparison with mechanistic interpretability.} Mechanistic interpretability treats a neural network as a collection of distinct components rather than a single black box~\citep{bolei2019interpreting, kevin2022locating, wang2023interpretability}. It studies the roles of individual neurons, attention heads, or subnetworks, and how these parts work together to produce the model's overall behavior. Instead of physically dividing a neural network into sub-components, interaction-based explanations treats a neural network as a whole and explains the inference logic equivalently encoded by the network using a set of AND-OR interactions.

\section{Related work on parameter pruning}
\label{sec:apdx-related-pruning}

Parameter pruning has been widely studied as an effective approach for compressing deep neural networks (DNNs). Early works mainly focused on reducing model redundancy through weight pruning, quantization, and coding techniques to improve storage and inference efficiency. For example, \citet{han2016deep} proposed a classical pruning pipeline that combines parameter pruning, quantization, and Huffman coding for compact model representation. More recent studies have further extended pruning techniques to modern architectures and large language models (LLMs). Representative methods include one-shot pruning approaches such as SparseGPT \citep{frantar2023sparsegpt} and Wanda \citep{sun2023wanda}, as well as structural pruning methods that explicitly consider dependency relationships among network components \citep{fang2023depgraph, ma2023llmpruner}.

Most existing studies on parameter pruning primarily focus on developing more effective compression algorithms, pruning criteria, or acceleration strategies. Although these methods demonstrate strong empirical performance, their understanding of \emph{why} pruning causes performance degradation often remains relatively superficial, typically emphasizing the importance or sensitivity of the pruned parameters themselves. In contrast, our work investigates a different scientific problem: \textit{what internal factors lead to performance degradation in compressed DNNs}. Rather than proposing a new pruning algorithm, we aim to reveal the intrinsic mechanisms underlying pruning-induced performance degradation from the perspective of internal interactions within DNNs.

\newpage

\section{Practical value of our findings}
\label{sec:apdx-practical-value}

According to our findings, low-order generalizable interactions are removed at higher pruning ratios than high-order non-generalizable interactions. Based on this observation, and inspired by~\citet{zhangjp2026penalize}, we penalize the strength of all non-generalizable interactions during DNN training. This method encourages the DNN to encode more generalizable interactions. Specifically, we co-train two DNNs initialized with different random seeds, and explicitly penalize non-transferable interactions between them. Since the co-training procedure is performed entirely on the training set, it does not introduce additional information from the test data.

We trained VGG-16~\citep{vgg} and AlexNet~\citep{alexnet} on the training split of the TinyImageNet dataset~\citep{tinyimagenet}, with and without the proposed penalty.

\begin{figure}[H]
  	\centering
  	\includegraphics[width=0.86\textwidth]{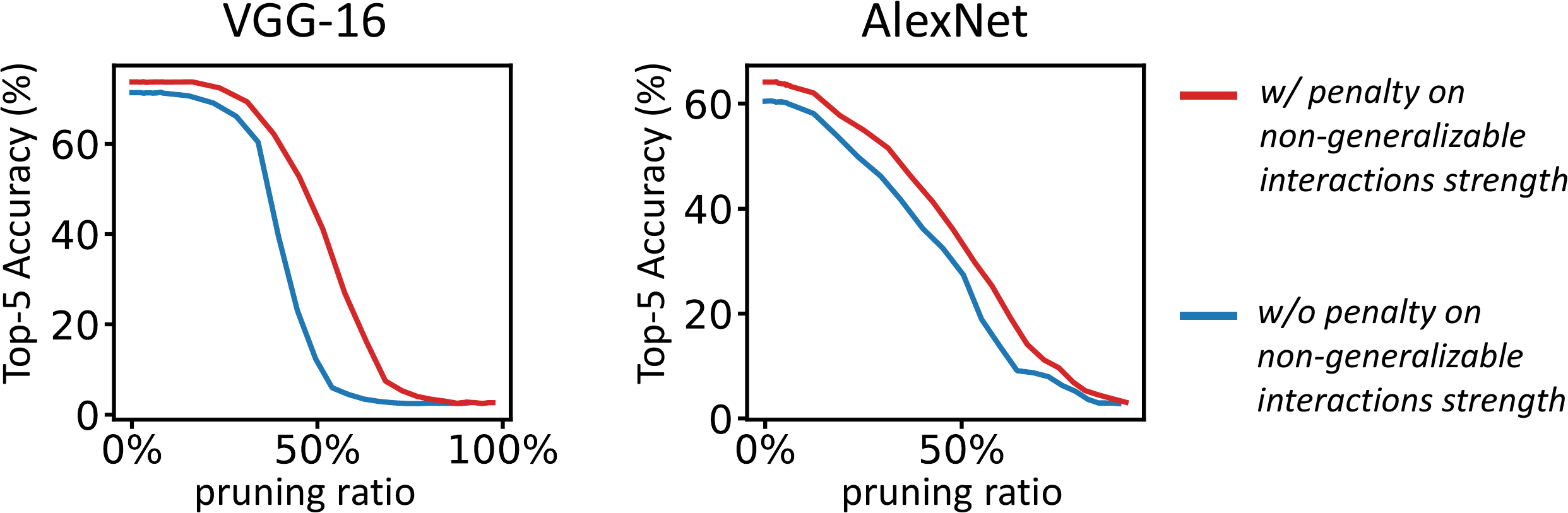}
	
  	\caption{
        Top-5 classification accuracy of pruned VGG-16 and pruned AlexNet on TinyImageNet under different pruning ratio.
    }
  	\label{fig:practical}
\end{figure}

As shown in Figure~\ref{fig:practical}, when pruned using the DepGraph method~\citep{fang2023depgraph}, DNNs trained with this penalty maintain their performance under higher pruning ratios compared to those trained without it.

\newpage

\section{Details of extracting the sparsest AND-OR interactions}
\label{sec:apdx-optimize-pq}

A method is proposed~\cite{chen2024defining, li2023defining} to simultaneously extract AND interactions $I^{{\typeand}}_T$ and OR interactions $I^{{\typeor}}_T$ from the network output. Given a masked sample $\boldsymbol{x}_L$, \cite{li2023defining} proposed to learn a decomposition $v(\boldsymbol{x}_L)=u^{{\typeand}}_L + u^{{\typeor}}_L$ towards the sparsest interactions. 
The component {$u^{{\typeand}}_L$} was explained by AND interactions, and the component {$u^{{\typeor}}_L$} was explained by OR interactions.
Specifically, they decomposed $v(\boldsymbol{x}_L)$ into $u^{{\typeand}}_L= 0.5 \cdot v(\boldsymbol{x}_L)+\gamma_L$ and $u^{{\typeor}}_L= 0.5 \cdot v(\boldsymbol{x}_L) -\gamma_L$, where $\{\gamma_L:L\subseteq N\}$ is a set of learnable variables that determine the decomposition. In this way, the AND interactions and OR interactions can be computed according to Equation~\ref{eq:logic}, \emph{i.e.}, $I^{{\typeand}}_T=\sum\nolimits_{L \subseteq T}(-1)^{|T|-|L|} u^{{\typeand}}_L$, and  $I^{{\typeor}}_T=-\sum\nolimits_{L \subseteq T}(-1)^{|T|-|L|} u^{{\typeor}}_{N \setminus L}$.

The parameters $\{\gamma_L\}$ were learned by minimizing the following LASSO-like loss to obtain sparse interactions:
\begin{equation}
\label{eq:loss-pq}
    \min_{\{\gamma_L\}} \sum_{T\subseteq N,\; T \neq \emptyset} \vert I^{{\typeand}}_T \vert + \vert I^{{\typeor}}_T \vert
\end{equation}

The following pseudocode \ref{alg:1} outlines the core procedure of extracting AND-OR
interactions.

\begin{algorithm}[!th] 
	\caption{Computing AND-OR interactions}
	\label{alg:1}
	\begin{algorithmic}[1] \label{algorithm}
		\STATE \textbf{Input:} Input sample $\x$, the DNN $v(\cdot)$
  
		\STATE \textbf{Output:} A set of interactions $I^{\typeand}_S$ and   $I^{\typeor}_S$

		\FOR{$S \subseteq N,\; S \neq \emptyset$} 
    \STATE For each masked sample $\x_S$, compute the confidence score $v(\x_S)$ based on Eq.~(\ref{eq:v});
		\ENDFOR

            \FOR{$S \subseteq N,\; S \neq \emptyset$} 
		\STATE Given $v(\x_S)$ for all combinations $S\subseteq N,\; S \neq \emptyset$, compute each AND interaction effect $I^{\typeand}_S$  and each OR interaction effect $I^{\typeor}_S$ via $\min_{\{\gamma_T\}}\sum_{S\subseteq N, S\ne \emptyset}[|I^{\typeand}_S| + |I^{\typeor}_S|]$;
		\ENDFOR

		\STATE return  $I^{\typeand}_S$, $I^{\typeor}_S$
	\end{algorithmic}
\end{algorithm}

Furthermore, in terms of evaluating the interaction transferability across multiple DNNs (e.g., the original DNN $v(\cdot)$ and the pruned DNN $\p(\cdot)$), \citet{chen2024defining} have found that people must solve the sensitivity of interactions to parameters $\{\gamma_L\}$, which hurts the stability of the evaluation. Thus, we follow the typical algorithm in \citep{chen2024defining} to extract a set of stable interactions that are shared by two DNNs, as follows.


\begin{equation}
\label{eq:loss-pq-shared}
\begin{aligned}
    \min_{\{\gamma_L, \tilde{\gamma}_L\}} \quad 
    & \sum_{T\subseteq N,\; T \neq \emptyset} \Big(
        \lvert I^{{\typeand}}_T \rvert 
      + \lvert I^{{\typeor}}_T \rvert 
      + \lvert \tilde{I}^{{\typeand}}_T \rvert 
      + \lvert \tilde{I}^{{\typeor}}_T \rvert
    \Big) \\
    & + \sum_{T\subseteq N,\; T \neq \emptyset} \Big(
        \lvert \min(0, I^{{\typeand}}_T, \tilde{I}^{{\typeand}}_T) \rvert 
      + \lvert \max(0, I^{{\typeand}}_T, \tilde{I}^{{\typeand}}_T) \rvert \\
    & \qquad\quad
      + \lvert \min(0, I^{{\typeor}}_T, \tilde{I}^{{\typeor}}_T) \rvert 
      + \lvert \max(0, I^{{\typeor}}_T, \tilde{I}^{{\typeor}}_T) \rvert
    \Big)
\end{aligned}
\end{equation}


\textbf{Removing small noises.} A small noise $\delta$ in the network output may significantly affect the extracted interactions, especially for high-order interactions. Thus, Li et al. \cite{li2023defining} proposed to learn to remove a small noise term $\delta_T$ from the computation of AND-OR interactions.
Specifically, the decomposition was rewritten as {$u^{{\typeand}}_L=0.5 (v(\boldsymbol{x}_L) -\delta_L) +\gamma_L$} and {$u^{{\typeor}}_L=0.5 (v(\boldsymbol{x}_L)-\delta_L) -\gamma_L$}.
Thus, the parameters {$\{\delta_L\}$} and {$\{\gamma_L\}$} are simultaneously learned by minimizing the loss function in Eq.~(\ref{eq:loss-pq}).
The values of {$\{\delta_L\}$} were constrained in $[-\zeta, \zeta]$ where {$\zeta=0.02\cdot \vert v(\boldsymbol{x})-v(\boldsymbol{x}_\emptyset) \vert$}.
\newpage

\section{Experimental detail}
\label{sec:experimental_setting}

\subsection{Training settings} 

In this paper, we followed~\citet{fang2023depgraph} and trained ResNet-56~\citep{resnet} and VGG-19~\citep{vgg} on the CIFAR-10 dataset~\citep{cifar} for 200 epochs using stochastic gradient descent (SGD) with a learning rate of $0.1$, momentum of $0.9$, and a weight decay of $5 \times 10^{-4}$. The learning rate was decayed by a factor of $0.1$ at epochs 120, 150, and 180. We used a batch size of 128 in all experiments.

All experiments were conducted on a compute node equipped with dual Intel Xeon Silver 4310 CPUs (48 logical cores) and a combination of four NVIDIA A800 and two NVIDIA A100 GPUs (each with 80GB memory).

\subsection{\label{subsec:apdx-pruning-setting-three-phase}Pruning settings in Section \ref{subsec:three-phase}}

We used DepGraph~\citep{fang2023depgraph} to prune ResNet-56~\citep{resnet} and VGG-19~\citep{vgg} on the CIFAR-10 dataset~\citep{cifar}.
For ResNet-56, we progressively increase the pruning ratio from 1.2\% to 50.2\%. And for VGG-19, we progressively increase the pruning ratio from 3.6\% to 99.2\%.
All other pruning settings followed the original settings in~\citep{fang2023depgraph}, respectively. No post-pruning finetuning was performed.

\subsection{\label{subsec:apdx-pruning-setting-module}Pruning settings in Section \ref{subsec:further-validation}}
We prune four LLMs (including LLaMA-7B~\citep{llama1}, LLaMA-3.1-8B~\citep{llama3}, DeepSeek-R1-Distill-LLaMA-8B~\citep{deepseek}, Qwen-2.5-7B~\citep{qwen}).
Additionally, we conduct experiments on models from the ResNet and ViT families, including ResNet-50/101/152~\citep{resnet} and ViT-Small/Base/Large~\citep{vit}, respectively. All models are taken from the timm library~\citep{timm} and pretrained on ImageNet-1k~\citep{imagenet}. For ResNet models, we use the a1 / a1h pretrained variants~\citep{timm_resnet}, and for ViT models we use the augreg2-in21k-ft-in1k pretrained variants~\citep{timm_vit}.

In each experiment, we prune a single module from the original model. We apply two typical module-level pruning settings:
\textit{(1) Residual block pruning.} We prune a single residual block by skipping its computation on all hidden features, while keeping the rest of the network intact. For the ResNet family, each block consists of three convolutional layers. For Transformer-based models, including ViT family and large language models (LLMs), each block consists of a self-attention layer followed by a multi-layer perceptron (MLP).
\textit{(2) Hidden feature dimension pruning.} Rather than pruning within a single residual block, we remove selected dimensions from all hidden features, and apply the same dimensionality reduction to all subsequent computations (including, in ResNet models, convolutional channel dimensions and corresponding normalization layers; and in transformer-based models, the hidden size of all attention and MLP projections, along with normalization layers), ensuring consistency across the entire network. The pruned dimensions are randomly selected for each experiment, with 3\% to 15\% of the total hidden dimensionality being removed.

\textbf{Threshold selection for module sensitivity.}
Based on the above module settings, we categorize all pruned modules into two types: (1) sensitive modules, and (2) insensitive modules, based on the increase of test cross-entropy loss caused by the pruning.
If pruning a module leads to a significant increase in test loss greater than a predefined threshold $\tau$, then the corresponding module is regarded as a sensitive module; otherwise, it is regarded as an insensitive module.

The threshold $\tau$ is selected according to the increase in test cross-entropy (CE) loss induced by parameter pruning. Empirically, we consider a test CE loss increase of roughly 0.5 as indicating substantial degradation. Because different DNN architectures exhibit different CE loss magnitudes, $\tau$ is adjusted slightly on a per-model basis to ensure a consistent notion of significant degradation across models. Table~\ref{tab:tau-setting} lists the resulting threshold values.

\begin{table}[ht]
\centering
\caption{Threshold $\tau$ used for different DNNs.}
\label{tab:tau-setting}
\resizebox{0.4\linewidth}{!}
{\begin{tabular}{l c}
\toprule
Model & $\tau$ \\
\midrule
LLaMA-7B & 0.60 \\
LLaMA-3.1-8B & 1.00 \\
DeepSeek-R1-Distill-LLaMA-8B & 0.70 \\
Qwen-2.5-7B & 0.70 \\
ResNet-50 & 0.50 \\
ResNet-101 & 0.45 \\
ResNet-152 & 0.30 \\
ViT-Small & 0.80 \\
ViT-Base & 1.00 \\
ViT-Large & 0.25 \\
\bottomrule
\end{tabular}}
\end{table}


\subsection{Details about how to calculate interactions for different DNNs}
\label{subsec:apdx-interaction-settings}

\textbf{\textbullet \ For experiments on image classification models,} since the computational cost of interactions was intolerable, we applied a sampling-based approximation method to calculate AND-OR interactions. Specifically, we considered the feature map after the low-layer as intermediate-layer features of DNNs. We uniformly split each intermediate-layer feature map into $s \times s$ patches, and sampled 10 patches based on the highest significance values computed via Integrated Gradients~\citep{integratedgradients}. For ResNet-56 and VGG-19 on CIFAR-10, $s=8$. For ResNet family and ViT family on ImageNet-1k, $s=7$. These selected patches were considered as input variables for the corresponding intermediate-layer feature.
We replace variables with a baseline vector determined by following the method in \citep{chen2024defining} to mask the variables when computing interactions. The duration of the experiments ranges from 4 to 8 hours.

\textbf{\textbullet \ For experiments on large language models,} we manually collected 100 sentences for interaction extraction, which cover topics such as corporate restructuring, science and technology, economic policy, global trade, geopolitics, public health, political developments, cryptocurrency collapse, monetary policy, and climate and energy equity. The test cross-entropy loss analysis is performed on WikiText2 dataset~\citep{wikitext}. And the cross-entropy loss is averaged over all tokens.
We considered the one or multiple token embeddings that correspond to a single word as an input variable. And we randomly sampled 10 words for each sentence, which must have a specific meaning and not be stop words, to calculate interactions.
We replace variables with a baseline vector determined by following the method in \citep{chen2024defining} to mask the variables when computing interactions. The duration of the experiments ranges from 6 to 8 hours.

Specifically, we empirically considered the first 4 convolutional layer of VGG-19 as the low layers, and considered all the other 13 layers\footnote{VGG-19 for CIFAR-10~\citep{cifar} has been simplified compared to the original VGG-19 architecture designed for ImageNet, by replacing the multi-layer perceptron (MLP) classifier with a single fully connected layer and adapting the input resolution. The total number of layers, including both convolutional and fully connected layers, remains 17 in the simplified VGG-19 model for CIFAR-10.} as the high layers. For ResNet-56, we consider the first 3 convolution layers as low layers. For ResNet family, we consider the first 2 residual stages as low layers. For ViT family, we consider the first convolution layer as low layers. Typically, we compute the mean distribution of interactions over 100 samples for ResNet-56 on the CIFAR-10 dataset. We compute the mean distribution of interactions over 100 samples for all models.

\subsection{Settings of reference DNNs\label{subsec:apdx-reference}}

According to~\citet{he2025generalizability}, we need to train reference DNNs from scratch on the testing samples. However, for large-scale models, it is usually infeasible to determine which data samples were used for training and which were not. More importantly, retraining the model from scratch is impractical. Therefore, we use other classic models that are independently trained as the reference DNNs.

We use ResNet-56 and VGG-19 as reference models for each other. For the ResNet family, we adopt ViT-Large as the reference model, while for the ViT family, we use ResNet-152 as the reference model. For LLMs, we pair Qwen-2.5-7B with DeepSeek-R1-Distill-LLaMA-8B, and LLaMA-3.1-8B with LLaMA-7B as mutual reference models.

To ensure comparability across architectures, we align the spatial (or token) locations of sampled variables. Specifically, for ResNet-56 and VGG-19, we partition each image into $8\times 8$ patches, while for the ResNet and ViT families we use $7\times 7$ patches. For LLMs, words are naturally segmented into tokens. This unified partitioning ensures that the sampled variables are aligned across models, making the comparing between intersection sets mathematically well-defined.
\subsection{Licenses of Used Assets}
\label{subsec:licenses}

We list the licenses and terms of use for all assets used in this paper, including datasets and pre-trained models.

\textbf{Datasets.}
\begin{itemize}
    \item CIFAR-10 — publicly available dataset without an explicit license; used in accordance with standard academic usage and citation requirements
    \item ImageNet-1k — used under the official Terms of Access for non-commercial research; images are collected from third-party sources and may be subject to individual licenses
    \item WikiText-2 — Creative Commons Attribution-ShareAlike license (CC BY-SA), derived from Wikipedia text
\end{itemize}

\textbf{Pre-trained Models.}
\begin{itemize}
    \item LLaMA-7B — released under Meta's original LLaMA research license with restricted access and non-commercial use conditions
    \item LLaMA-3.1-8B — released under the LLaMA 3.1 Community License Agreement, subject to its terms of use and acceptable use policy
    \item Qwen-2.5-7B — Apache License 2.0
    \item DeepSeek-R1-Distill-LLaMA-8B — released under the MIT License, while incorporating components subject to the LLaMA license
    \item ResNet-50/101/152 — implemented using the timm library under the Apache License 2.0
    \item ViT-Small/Base/Large — implemented using the timm library under the Apache License 2.0
\end{itemize}

All assets are used in accordance with their respective licenses and terms of use.
Detailed license terms can be found on the official websites of the respective providers.

\newpage

\newpage
\section{Detailed results of the three-phase dynamics of interactions during progressive parameter pruning}
\label{sec:apdx-three-phase-details}

We provide additional results on interaction distributions after each pruning operation, and further summarize the changes in interactions during parameter pruning in a table for clarity.

\subsection{The distributions of interactions during progressive parameter pruning}
\label{subsec:apdx-distribution-three-phase}

We provide additional results on interaction distributions of ResNet-56 and VGG-19 after each pruning operation.

\subsubsection{The distribution of interactions encoded by ResNet-56 during progressive parameter pruning}

\begin{figure}[H]
  	\centering
  	\includegraphics[width=0.85\textwidth]{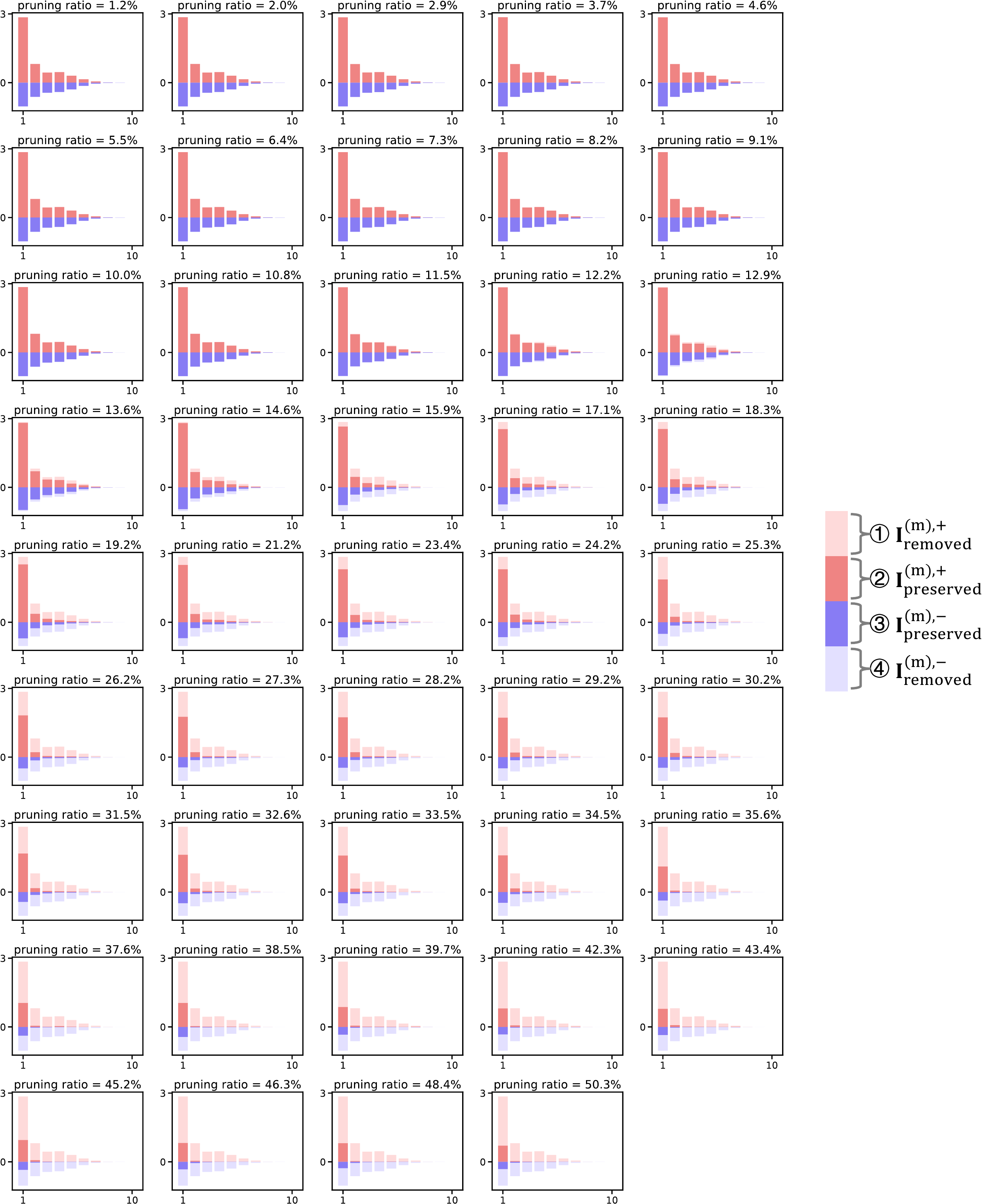}
  	\caption{The distributions of removed interactions ({\small $\dposR$} and {\small $\dnegR$}) and preserved interactions ({\small $\dposP$} and {\small $\dnegP$}) over different orders encoded by pruned ResNet-56 at different pruning ratios. The distributions are arranged in ascending order of the pruning ratio.}
  	\label{fig:details-resnet56}
\end{figure}

\subsubsection{The distribution of interactions encoded by VGG-19 during progressive parameter pruning}

\begin{figure}[H]
  	\centering
  	\includegraphics[width=0.85\textwidth]{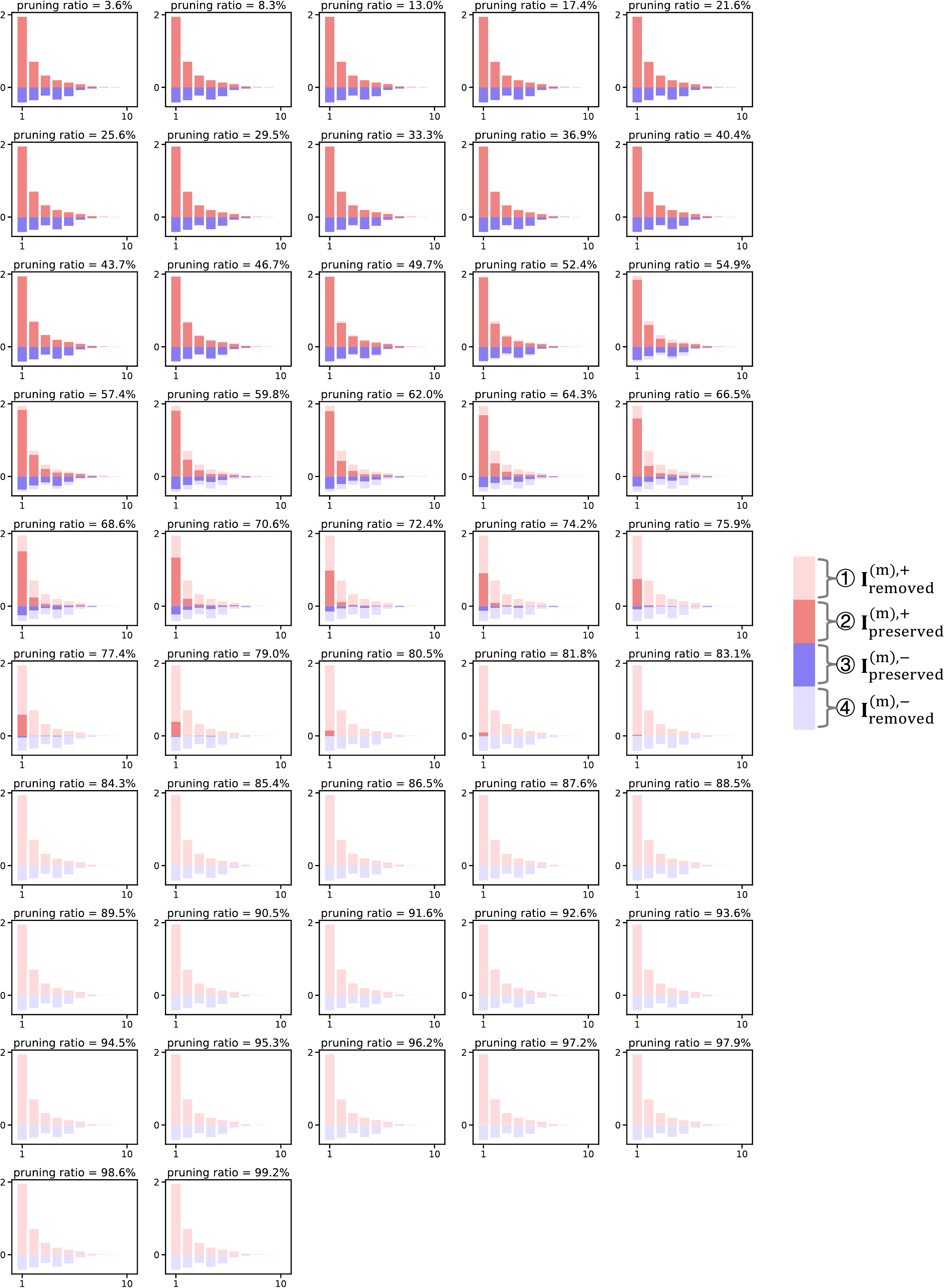}
  	\caption{The distributions of removed interactions ({\small $\dposR$} and {\small $\dnegR$}) and preserved interactions ({\small $\dposP$} and {\small $\dnegP$}) over different orders encoded by pruned VGG-19 at different pruning ratios. The distributions are arranged in ascending order of the pruning ratio.}
  	\label{fig:details-vgg19}
\end{figure}

\newpage
\subsection{Quantify the changes in interactions during progressive parameter pruning}
\label{subsec:apdx-metric-three-phase}

We quantify the changes in interactions after pruning from different perspectives, including the ratio of preserved low-order/high-order interactions $\rhol$/$\rhoh$, the offsetting ratio of removed interaction effects $\offset$, and the ratio of removed generalizable interactions $\rhoR$. For completeness, we include in this appendix some content that is identical to the main paper to ensure the appendix is self-contained and easy to follow.

\subsubsection{The ratio of preserved low-order/high-order interactions}

\begin{figure}[H]
  	\centering
  	\includegraphics[width=\textwidth]{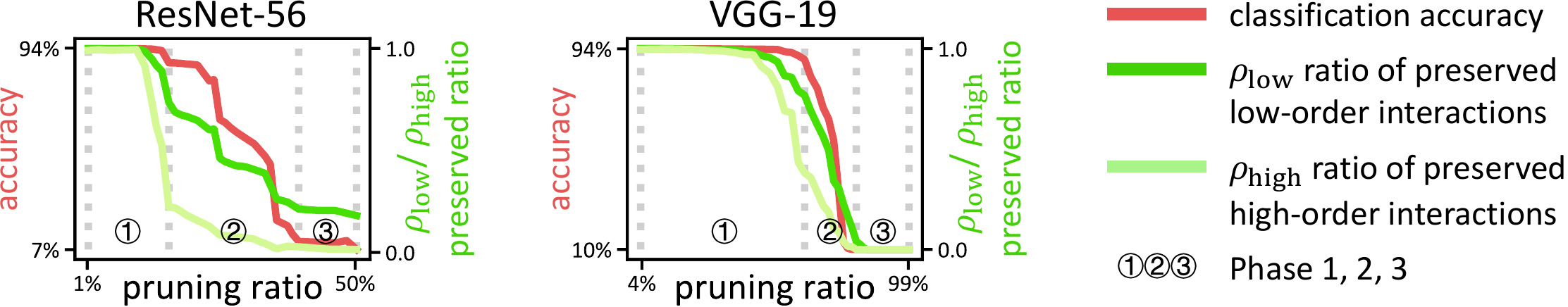}
  	\caption{The ratio of preserved low-order/high-order interactions $\rhol$/$\rhoh$ during progressive parameter pruning on different DNNs.}
  	\label{fig:rho-three-phase-full}
\end{figure}

Figure~\ref{fig:rho-three-phase-full} shows the ratio of preserved low-order/high-order interactions $\rhol$/$\rhoh$ during progressive parameter pruning on different DNNs.

\subsubsection{The offsetting ratio of removed interactions}

\begin{figure}[H]
  	\centering
  	\includegraphics[width=\textwidth]{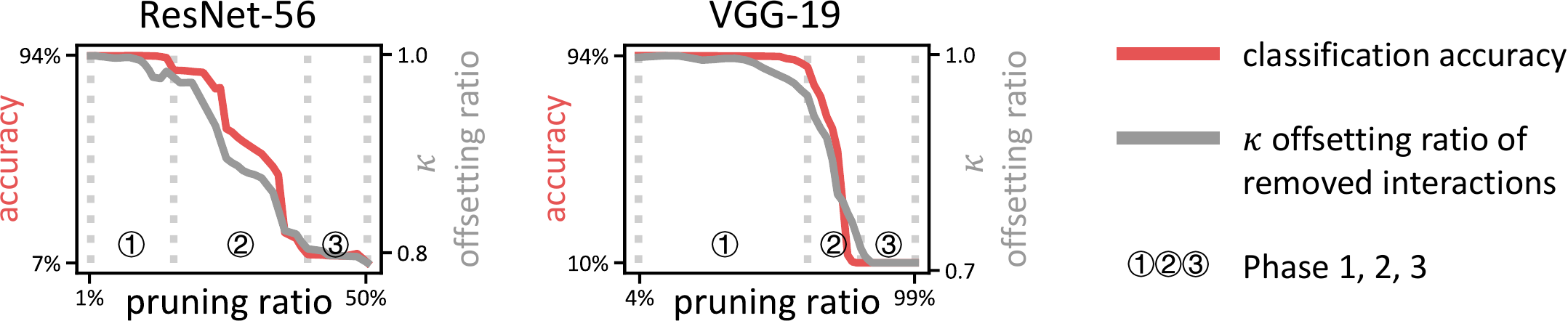}
  	\caption{The offsetting ratio of removed interactions $\offset$ during progressive parameter pruning on different DNNs.}
  	\label{fig:kappa-three-phase-full}
\end{figure}

Figure~\ref{fig:kappa-three-phase-full} shows the offsetting ratio of removed interactions $\offset$ during progressive parameter pruning on different DNNs.

\subsubsection{The ratio of removed generalizable interactions}

\begin{figure}[H]
  	\centering
  	\includegraphics[width=\textwidth]{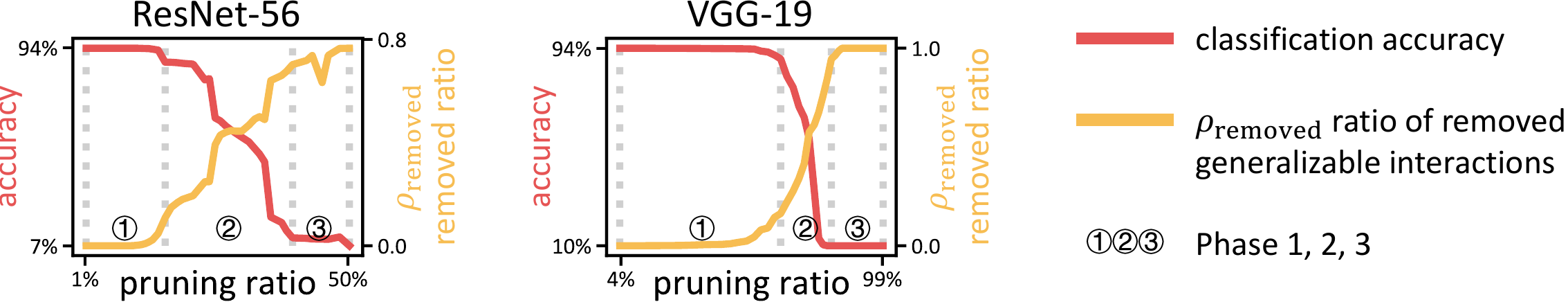}
  	\caption{The ratio of removed generalizable interactions $\rhoR$ during progressive parameter pruning on different DNNs.}
  	\label{fig:rho_removed-three-phase-full}
\end{figure}

Figure~\ref{fig:rho_removed-three-phase-full} shows the ratio of removed generalizable interactions $\rhoR$ during progressive parameter pruning on different DNNs.

\subsection{Analysis of three-phase dynamics of interactions during parameter pruning}
\label{subsec:apdx-analysis-three-phase}

Table~\ref{tab:phases} summarizes the changes in interactions during progressive parameter pruning from different perspectives.

\begin{table}[H]
\caption{Analysis of three-phase dynamics of interactions from different perspectives.}
\label{tab:phases}
\centering
\renewcommand{\arraystretch}{1.5} 
\resizebox{0.86\linewidth}{!}{
\begin{tabular}{llll}
\toprule
\textbf{Perspectives} & \textbf{Phase 1} & \textbf{Phase 2} & \textbf{Phase 3} \\
\midrule
Testing accuracy & barely affected & significantly decreased & remains low \\
Low-order interactions & slightly removed & significantly removed & most have been removed \\
High-order interactions & significantly removed & significantly removed & most have been removed \\
Offsetting of the removed & extremely high & comparatively low & comparatively low \\
Generalizability of the removed & slightly increase & significantly increase & most have been removed \\
\bottomrule
\end{tabular}
}
\end{table}
\newpage

\newpage
\section{Detailed results of pruning sensitive/insensitive modules}
\label{sec:apdx-module-pruning-details}

We provide additional results on interaction distributions after each pruning operation, and further quantify the changes in interactions after pruning on more DNNs.

\subsection{The distributions of interactions after the pruning of each module}
\label{subsec:apdx-distribution-module-pruning-full}

We provide additional results on interaction distributions of four LLMs, three ResNet models and three ViT models after each pruning operation.

\newcommand{\dwidth}{0.75\textwidth}

\subsubsection{The distribution of interactions encoded by LLaMA-7B after the pruning of each module}

\begin{figure}[H]
  	\centering
  	\includegraphics[width=\dwidth]{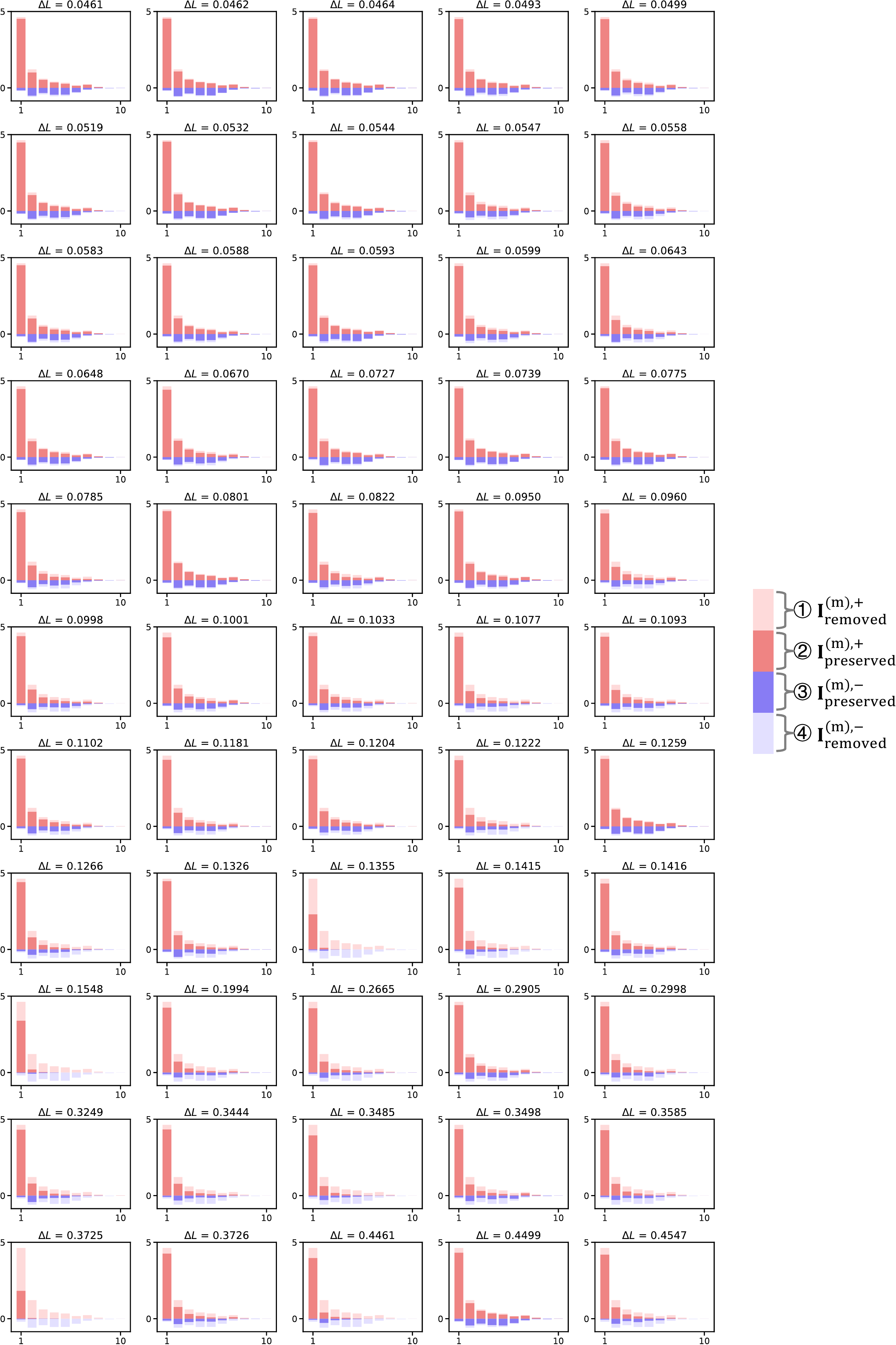}
  	\caption{The distributions of removed interactions ({\small $\dposR$} and {\small $\dnegR$}) and preserved interactions ({\small $\dposP$} and {\small $\dnegP$}) over different orders encoded by LLaMA-7B after pruning each sensitive/insensitive module. The distributions are arranged in ascending order of the increase in test loss after pruning. For clarity, this figure presents the first of two subsets of results.}
  	\label{fig:distribution-llama1-0}
\end{figure}

\begin{figure}[H]
  	\centering
  	\includegraphics[width=\dwidth]{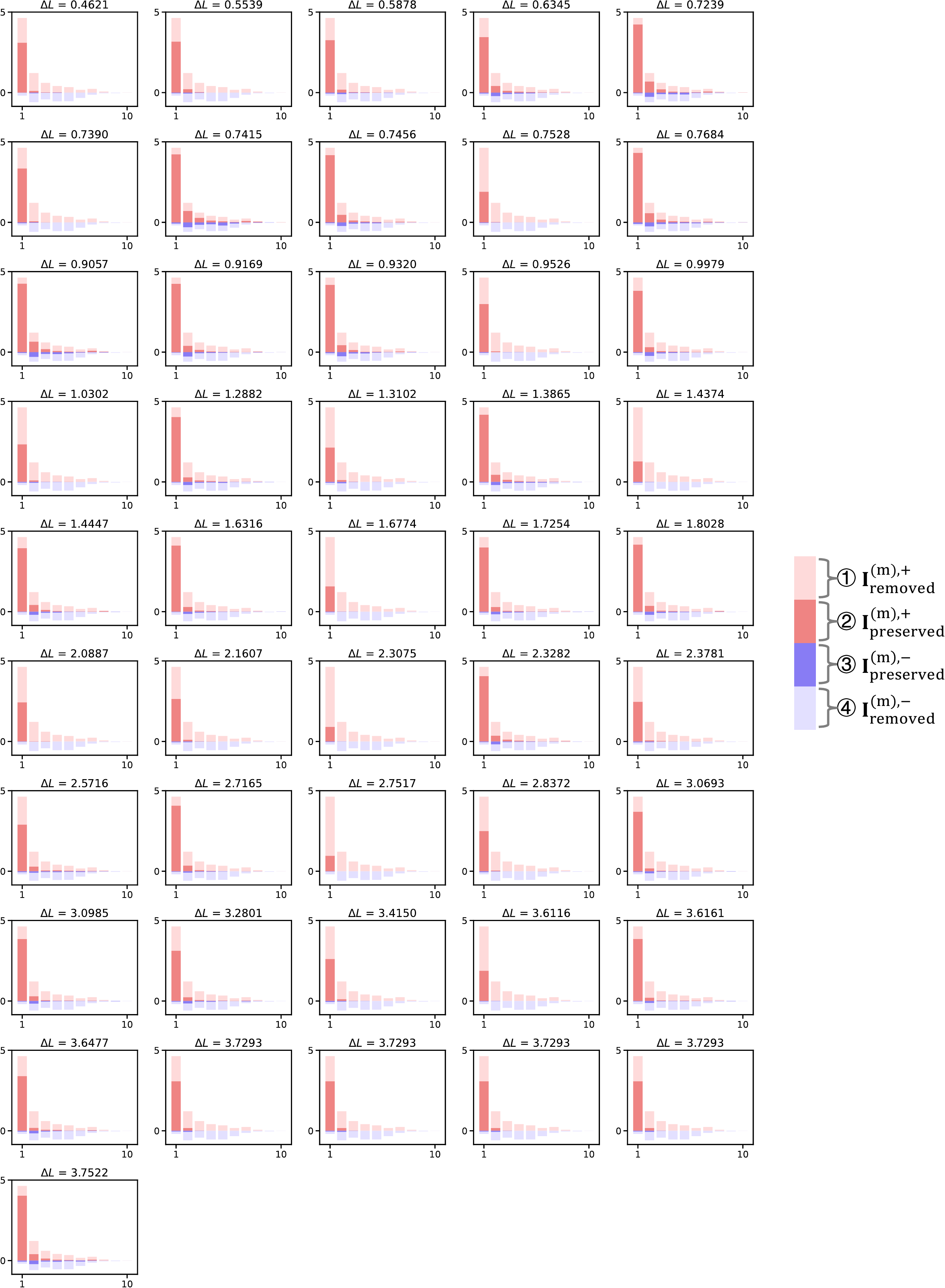}
  	\caption{The distributions of removed interactions ({\small $\dposR$} and {\small $\dnegR$}) and preserved interactions ({\small $\dposP$} and {\small $\dnegP$}) over different orders encoded by LLaMA-7B after pruning each sensitive/insensitive module. The distributions are arranged in ascending order of the increase in test loss after pruning. For clarity, this figure presents the second of two subsets of results.}
  	\label{fig:distribution-llama1-1}
\end{figure}

\subsubsection{The distribution of interactions encoded by LLaMA-3.1-8B after the pruning of each module}

\begin{figure}[H]
  	\centering
  	\includegraphics[width=\dwidth]{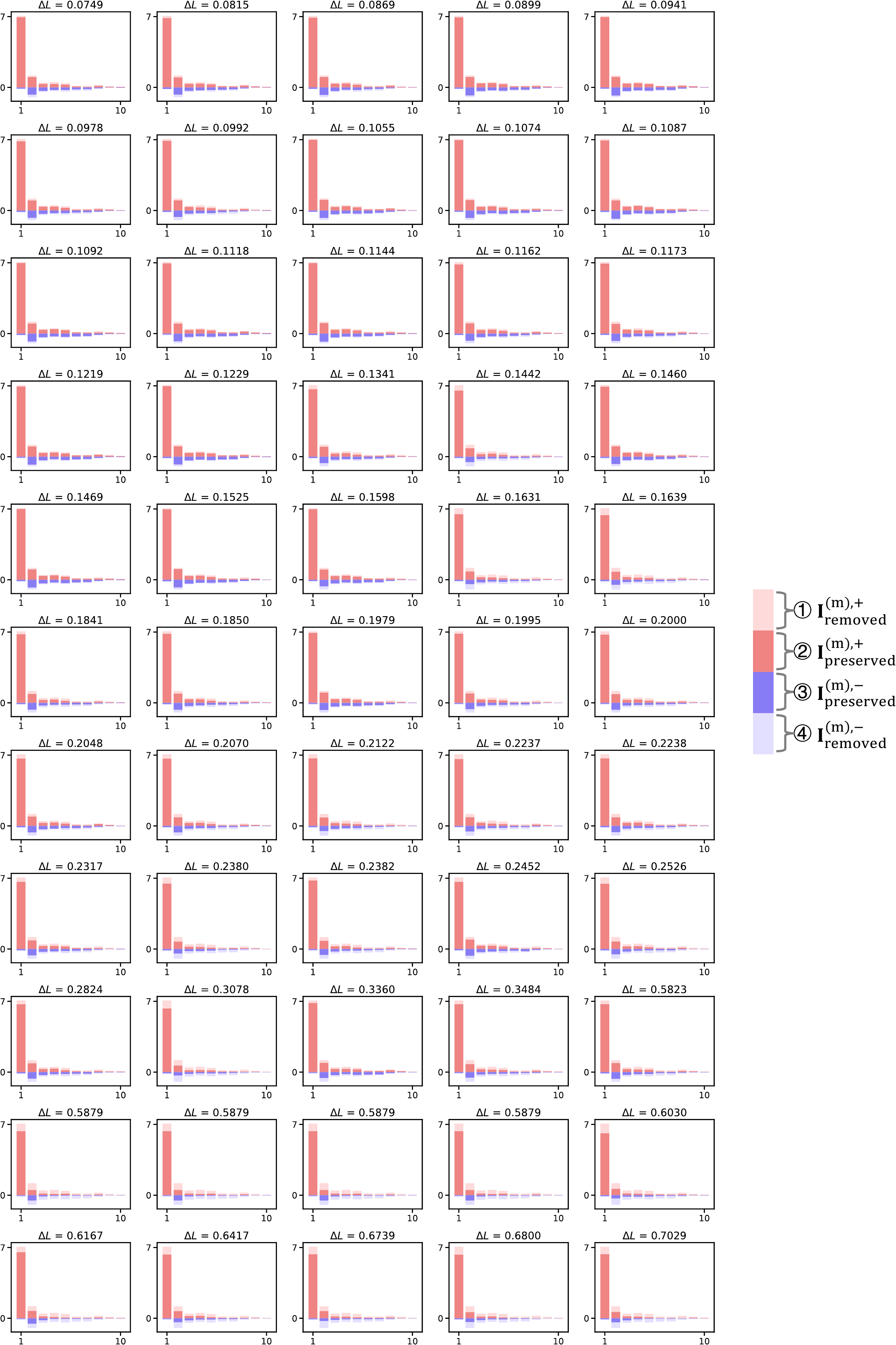}
  	\caption{The distributions of removed interactions ({\small $\dposR$} and {\small $\dnegR$}) and preserved interactions ({\small $\dposP$} and {\small $\dnegP$}) over different orders encoded by LLaMA-3.1-8B after pruning each sensitive/insensitive module. The distributions are arranged in ascending order of the increase in test loss after pruning. For clarity, this figure presents the first of two subsets of results.}
  	\label{fig:distribution-llama3-0}
\end{figure}

\begin{figure}[H]
  	\centering
  	\includegraphics[width=\dwidth]{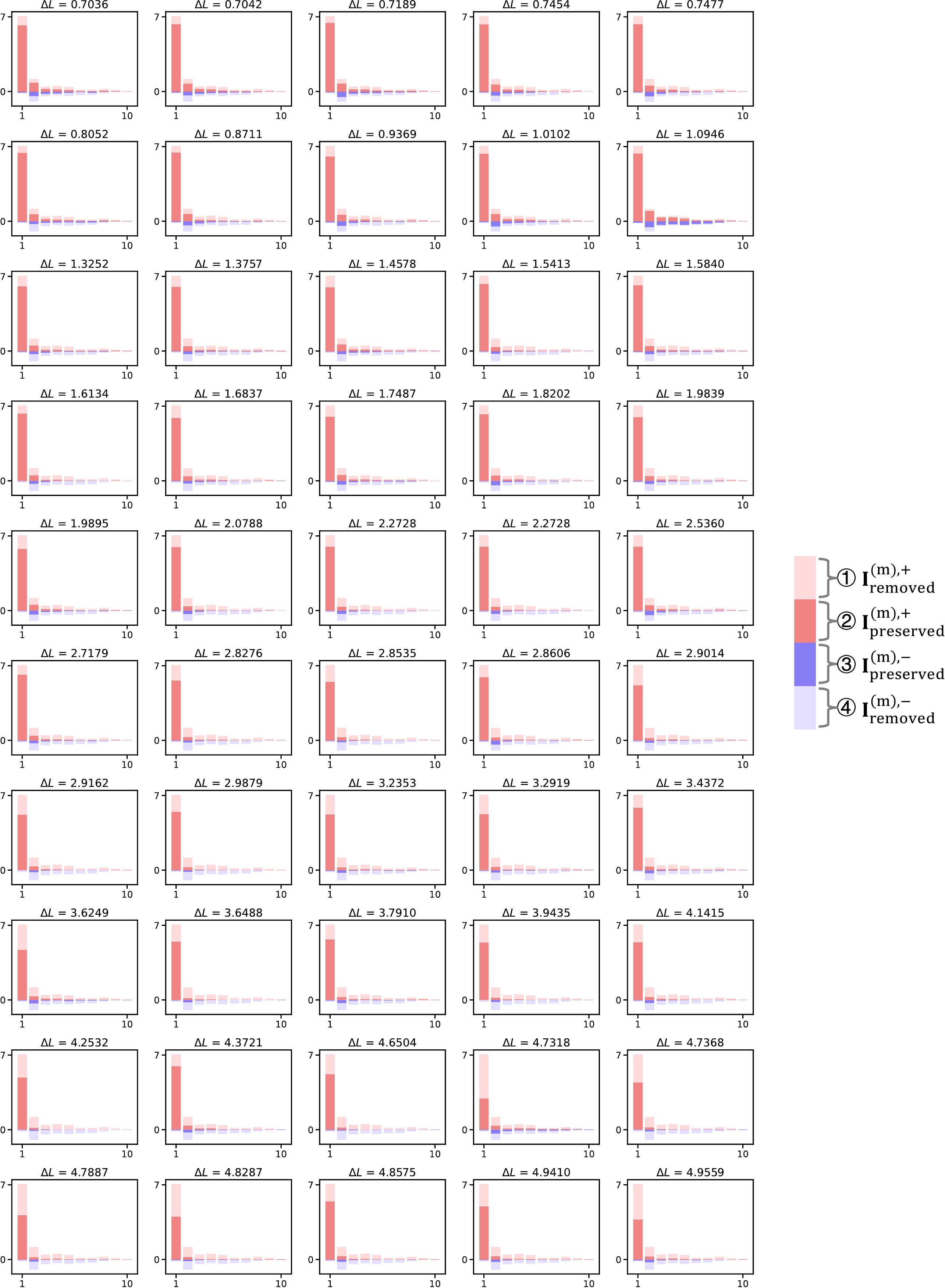}
  	\caption{The distributions of removed interactions ({\small $\dposR$} and {\small $\dnegR$}) and preserved interactions ({\small $\dposP$} and {\small $\dnegP$}) over different orders encoded by LLaMA-3.1-8B after pruning each sensitive/insensitive module. The distributions are arranged in ascending order of the increase in test loss after pruning. For clarity, this figure presents the second of two subsets of results.}
  	\label{fig:distribution-llama3-1}
\end{figure}

\subsubsection{The distribution of interactions encoded by DeepSeek-R1-Distill-LLaMA-8B after the pruning of each module}

\begin{figure}[H]
  	\centering
  	\includegraphics[width=\dwidth]{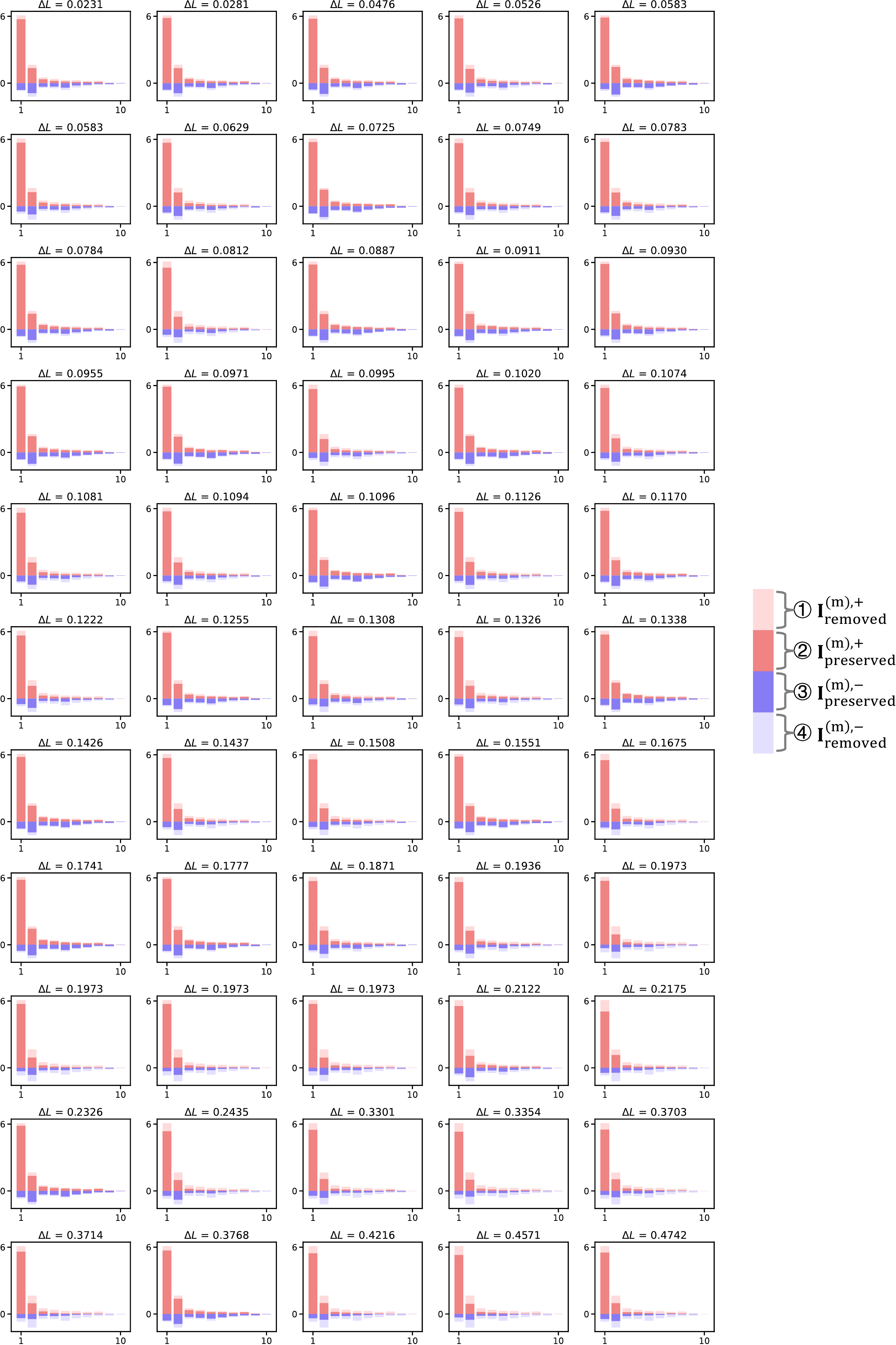}
  	\caption{The distributions of removed interactions ({\small $\dposR$} and {\small $\dnegR$}) and preserved interactions ({\small $\dposP$} and {\small $\dnegP$}) over different orders encoded by DeepSeek-R1-Distill-LLaMA-8B after pruning each sensitive/insensitive module. The distributions are arranged in ascending order of the increase in test loss after pruning. For clarity, this figure presents the first of two subsets of results.}
  	\label{fig:distribution-deepseek-0}
\end{figure}

\begin{figure}[H]
  	\centering
  	\includegraphics[width=\dwidth]{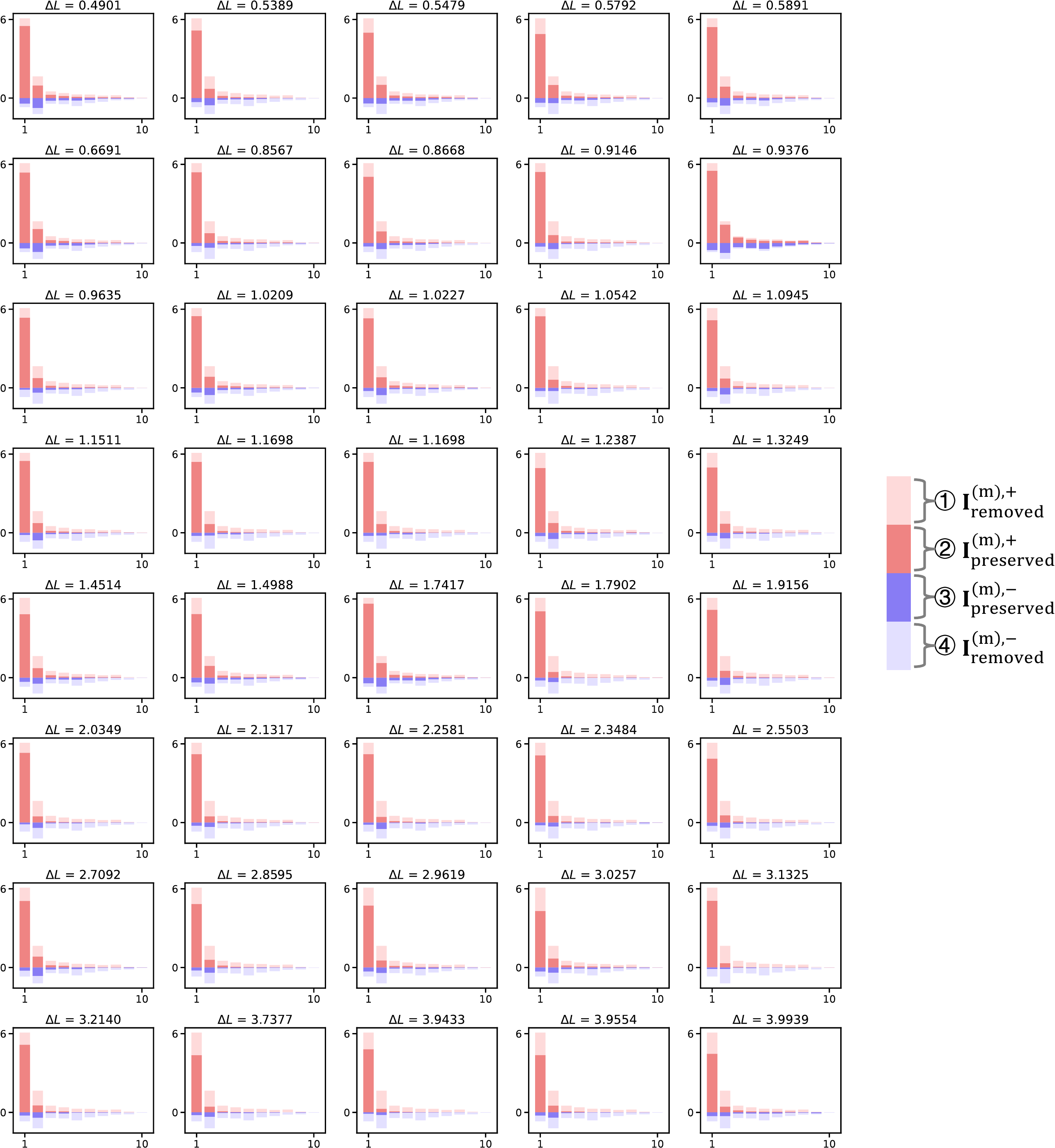}
  	\caption{The distributions of removed interactions ({\small $\dposR$} and {\small $\dnegR$}) and preserved interactions ({\small $\dposP$} and {\small $\dnegP$}) over different orders encoded by DeepSeek-R1-Distill-LLaMA-8B after pruning each sensitive/insensitive module. The distributions are arranged in ascending order of the increase in test loss after pruning. For clarity, this figure presents the second of two subsets of results.}
  	\label{fig:distribution-deepseek-1}
\end{figure}

\subsubsection{The distribution of interactions encoded by Qwen-2.5-7B after the pruning of each module}

\begin{figure}[H]
  	\centering
  	\includegraphics[width=\dwidth]{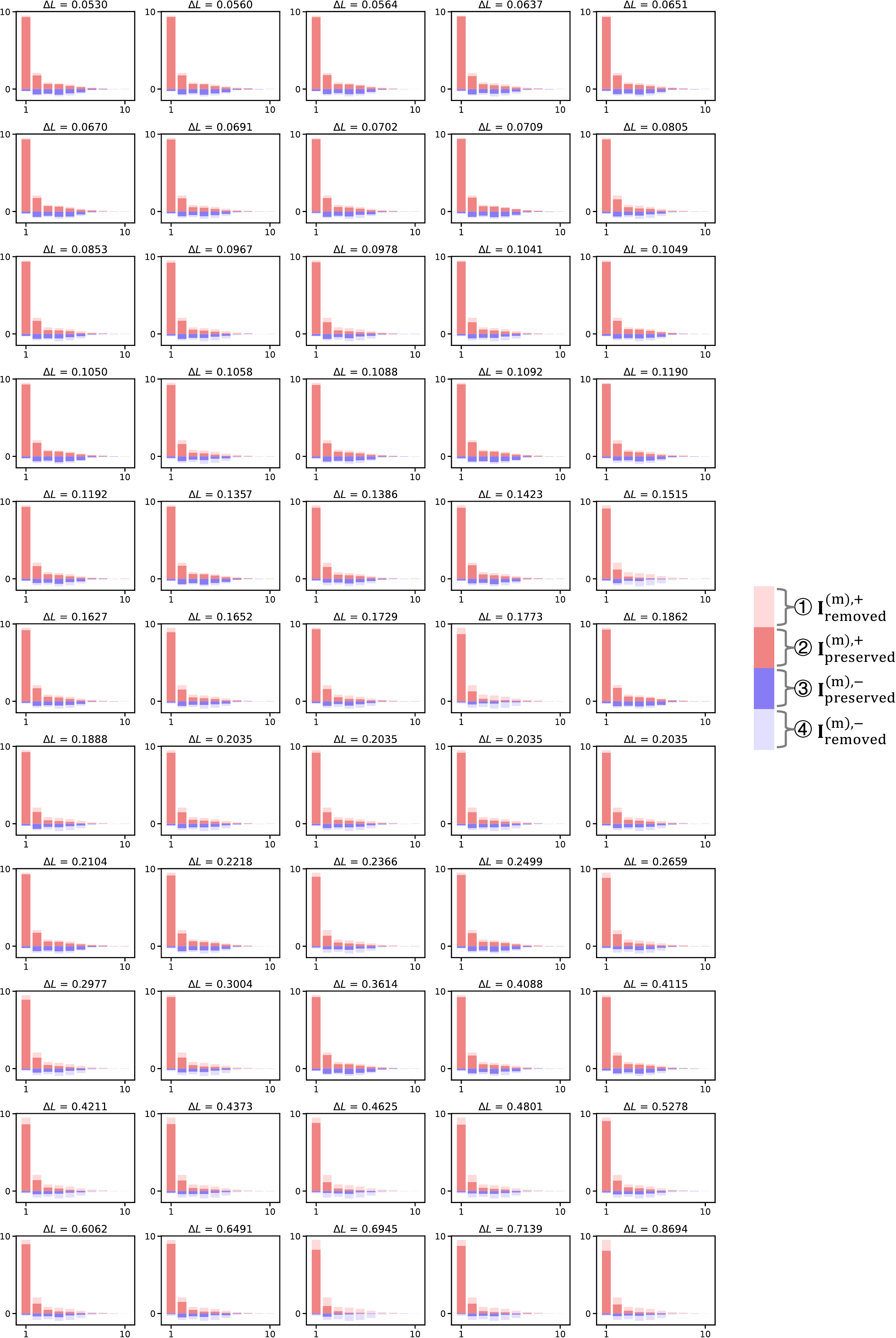}
  	\caption{The distributions of removed interactions ({\small $\dposR$} and {\small $\dnegR$}) and preserved interactions ({\small $\dposP$} and {\small $\dnegP$}) over different orders encoded by Qwen-2.5-7B after pruning each sensitive/insensitive module. The distributions are arranged in ascending order of the increase in test loss after pruning. For clarity, this figure presents the first of two subsets of results.}
  	\label{fig:distribution-qwen-0}
\end{figure}

\begin{figure}[H]
  	\centering
  	\includegraphics[width=\dwidth]{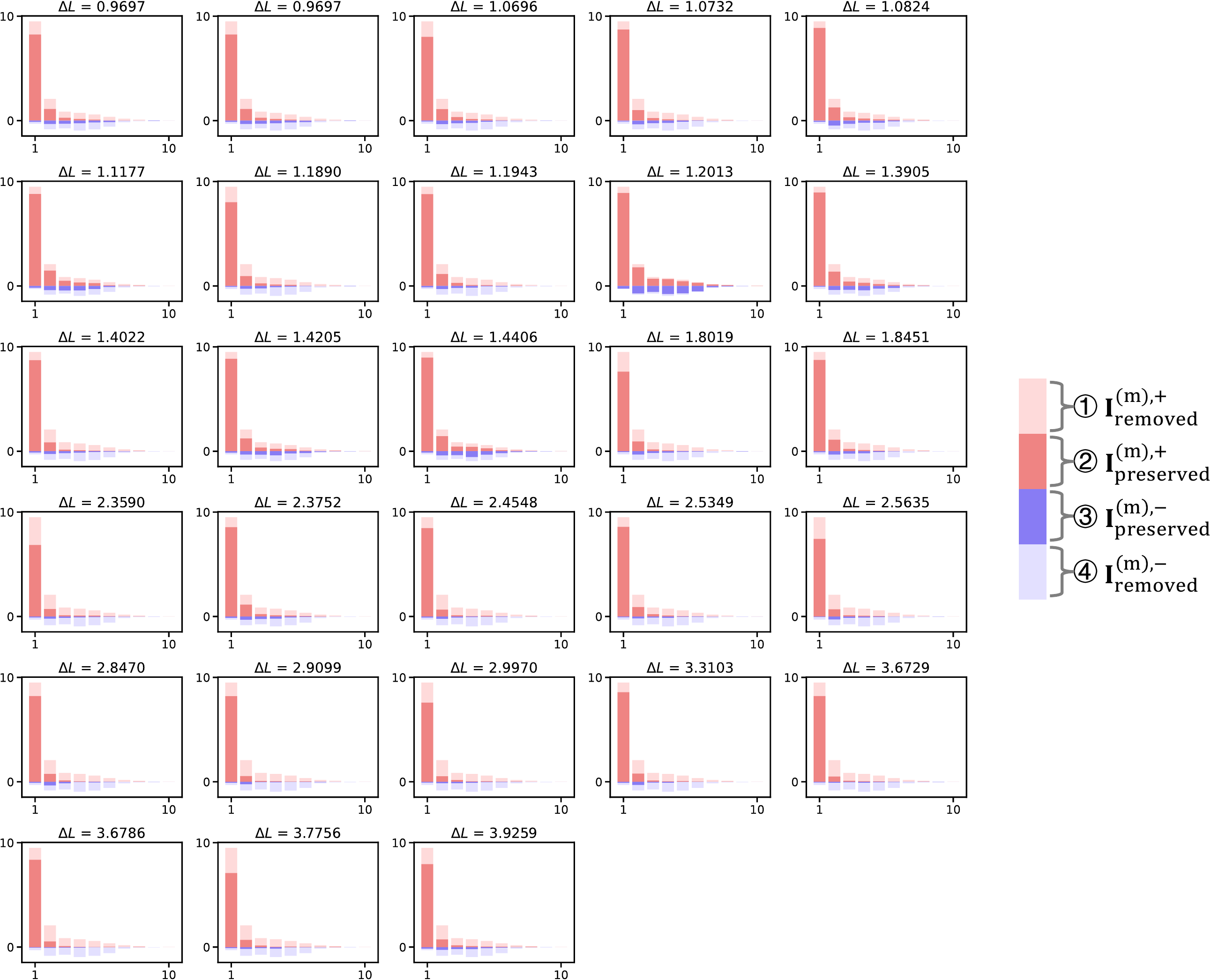}
  	\caption{The distributions of removed interactions ({\small $\dposR$} and {\small $\dnegR$}) and preserved interactions ({\small $\dposP$} and {\small $\dnegP$}) over different orders encoded by Qwen-2.5-7B after pruning each sensitive/insensitive module. The distributions are arranged in ascending order of the increase in test loss after pruning. For clarity, this figure presents the second of two subsets of results.}
  	\label{fig:distribution-qwen-1}
\end{figure}

\subsubsection{The distribution of interactions encoded by ResNet-50 after the pruning of each module}

\begin{figure}[H]
  	\centering
  	\includegraphics[width=\dwidth]{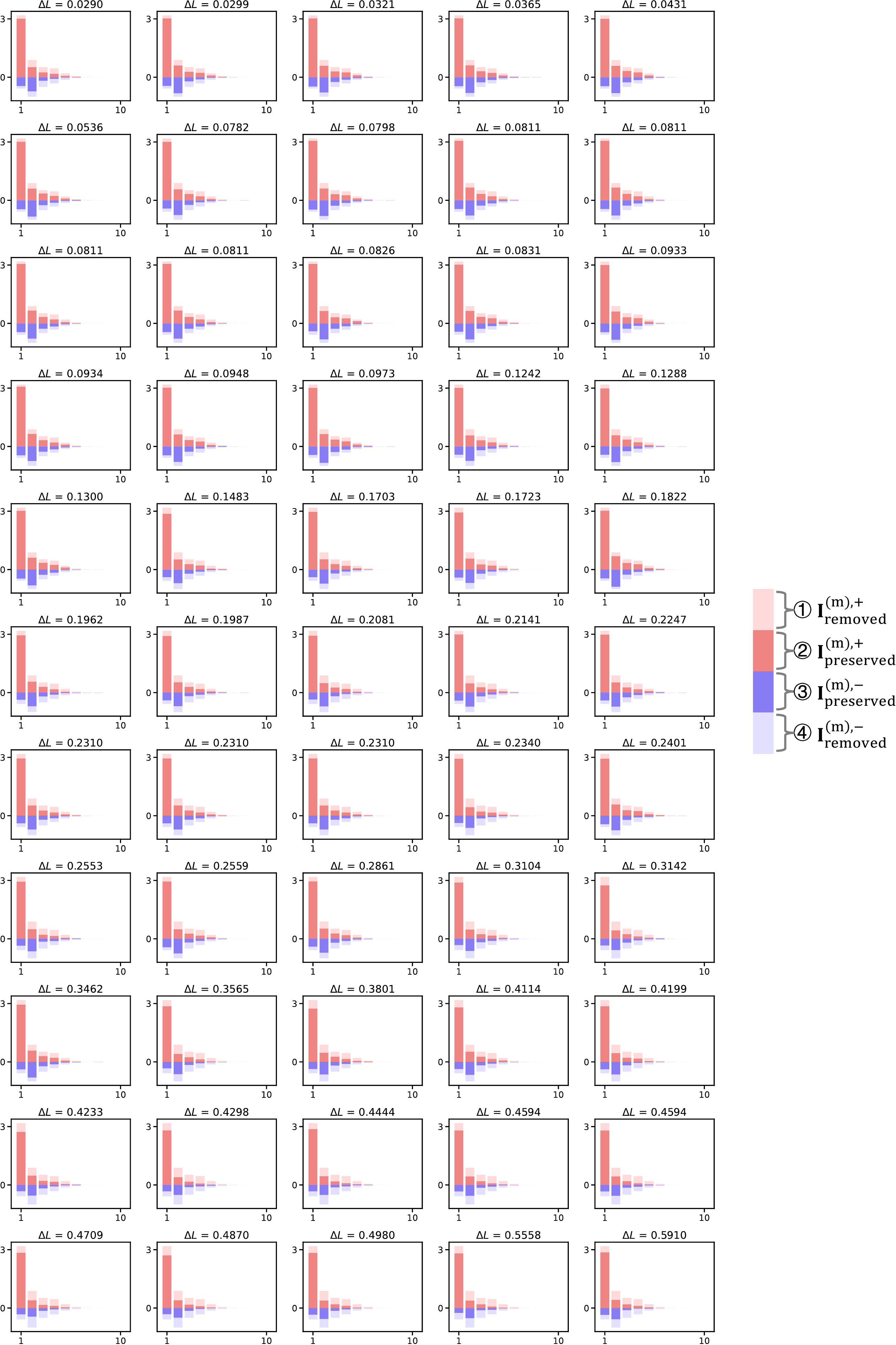}
  	\caption{The distributions of removed interactions ({\small $\dposR$} and {\small $\dnegR$}) and preserved interactions ({\small $\dposP$} and {\small $\dnegP$}) over different orders encoded by ResNet-50 after pruning each sensitive/insensitive module. The distributions are arranged in ascending order of the increase in test loss after pruning. For clarity, this figure presents the first of two subsets of results.}
  	\label{fig:distribution-resnet50-0}
\end{figure}

\begin{figure}[H]
  	\centering
  	\includegraphics[width=\dwidth]{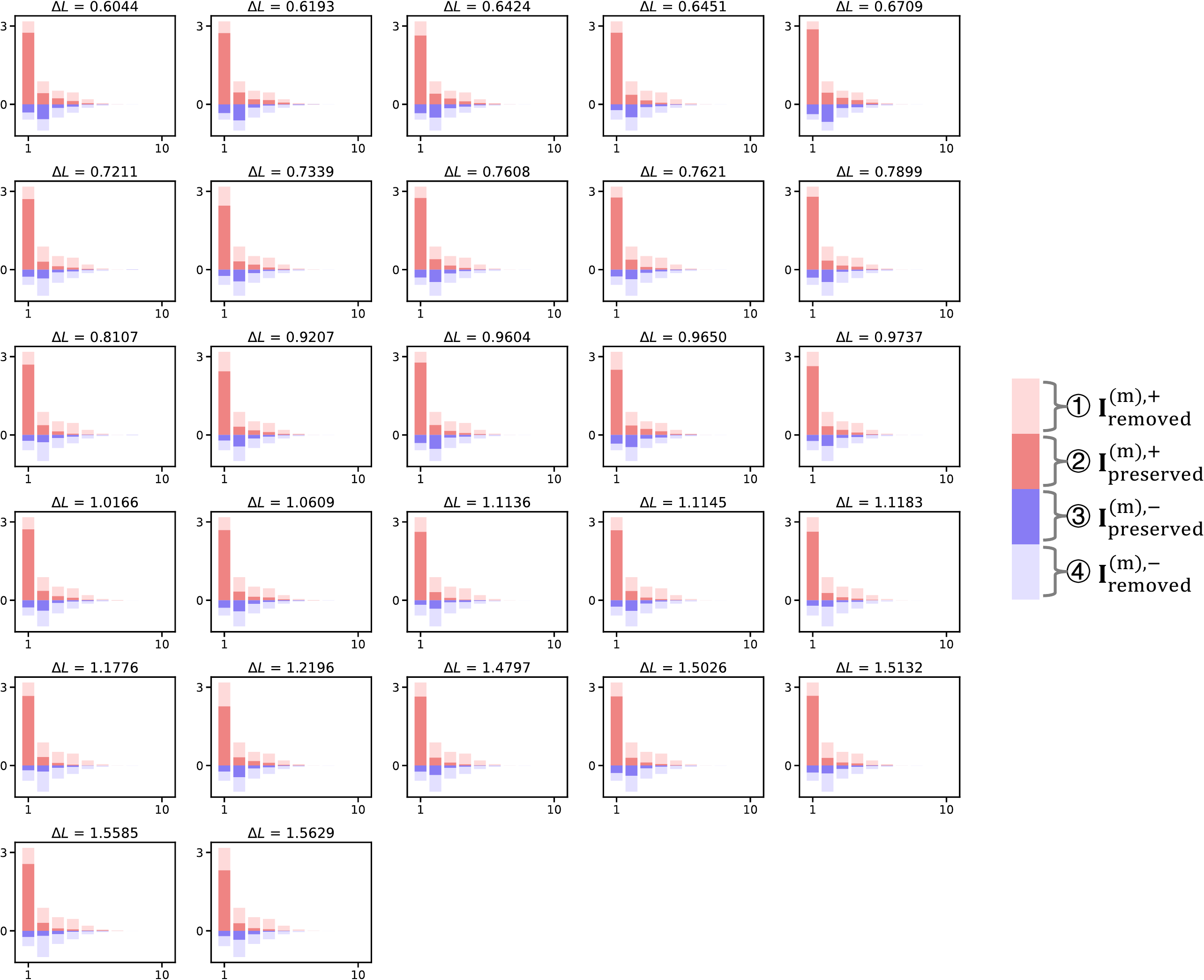}
  	\caption{The distributions of removed interactions ({\small $\dposR$} and {\small $\dnegR$}) and preserved interactions ({\small $\dposP$} and {\small $\dnegP$}) over different orders encoded by ResNet-50 after pruning each sensitive/insensitive module. The distributions are arranged in ascending order of the increase in test loss after pruning. For clarity, this figure presents the second of two subsets of results.}
  	\label{fig:distribution-resnet50-1}
\end{figure}

\subsubsection{The distribution of interactions encoded by ResNet-101 after the pruning of each module}

\begin{figure}[H]
  	\centering
  	\includegraphics[width=\dwidth]{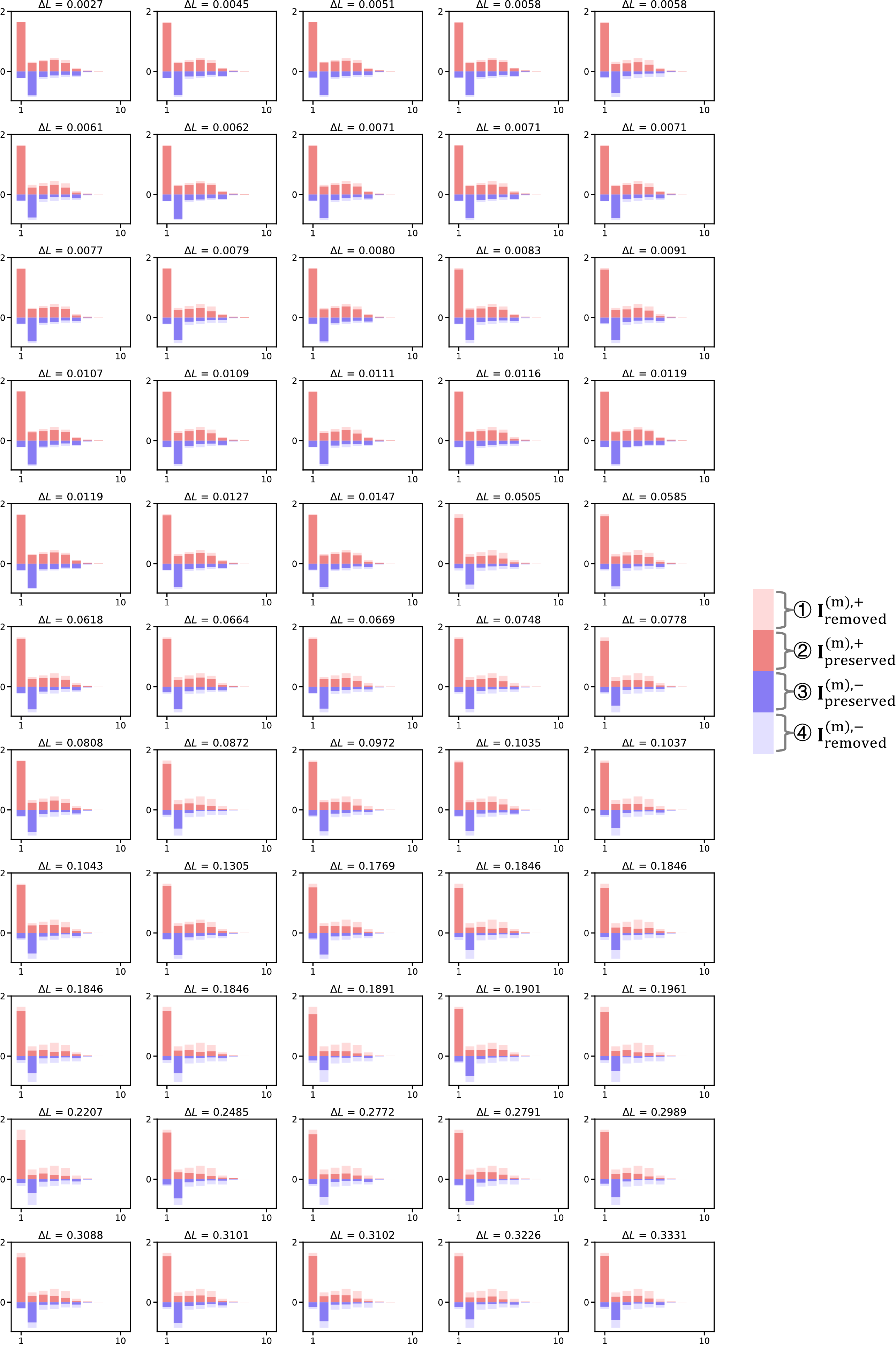}
  	\caption{The distributions of removed interactions ({\small $\dposR$} and {\small $\dnegR$}) and preserved interactions ({\small $\dposP$} and {\small $\dnegP$}) over different orders encoded by ResNet-101 after pruning each sensitive/insensitive module. The distributions are arranged in ascending order of the increase in test loss after pruning. For clarity, this figure presents the first of two subsets of results.}
  	\label{fig:distribution-resnet101-0}
\end{figure}

\begin{figure}[H]
  	\centering
  	\includegraphics[width=\dwidth]{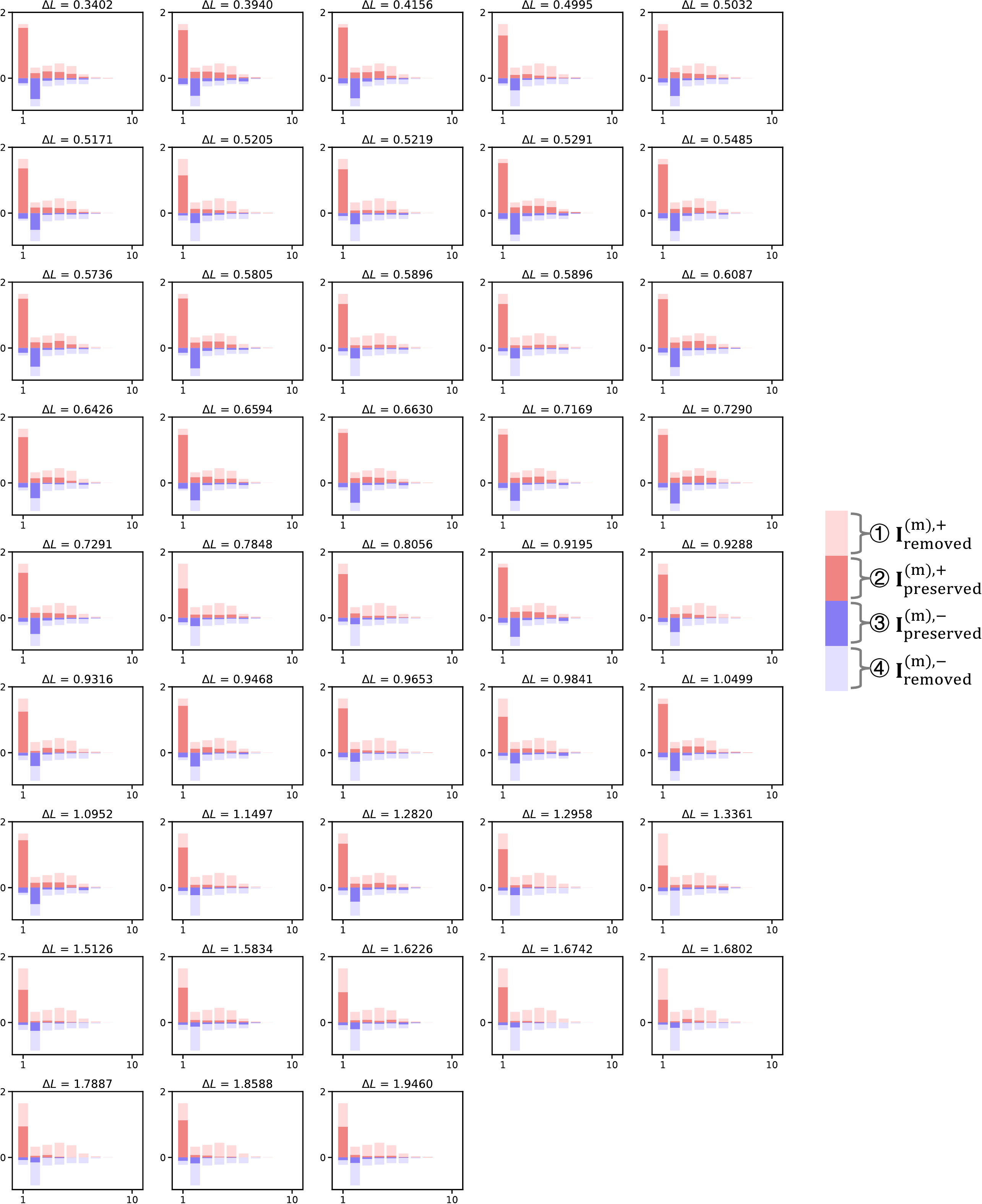}
  	\caption{The distributions of removed interactions ({\small $\dposR$} and {\small $\dnegR$}) and preserved interactions ({\small $\dposP$} and {\small $\dnegP$}) over different orders encoded by ResNet-101 after pruning each sensitive/insensitive module. The distributions are arranged in ascending order of the increase in test loss after pruning. For clarity, this figure presents the second of two subsets of results.}
  	\label{fig:distribution-resnet101-1}
\end{figure}

\subsubsection{The distribution of interactions encoded by ResNet-152 after the pruning of each module}

\begin{figure}[H]
  	\centering
  	\includegraphics[width=\dwidth]{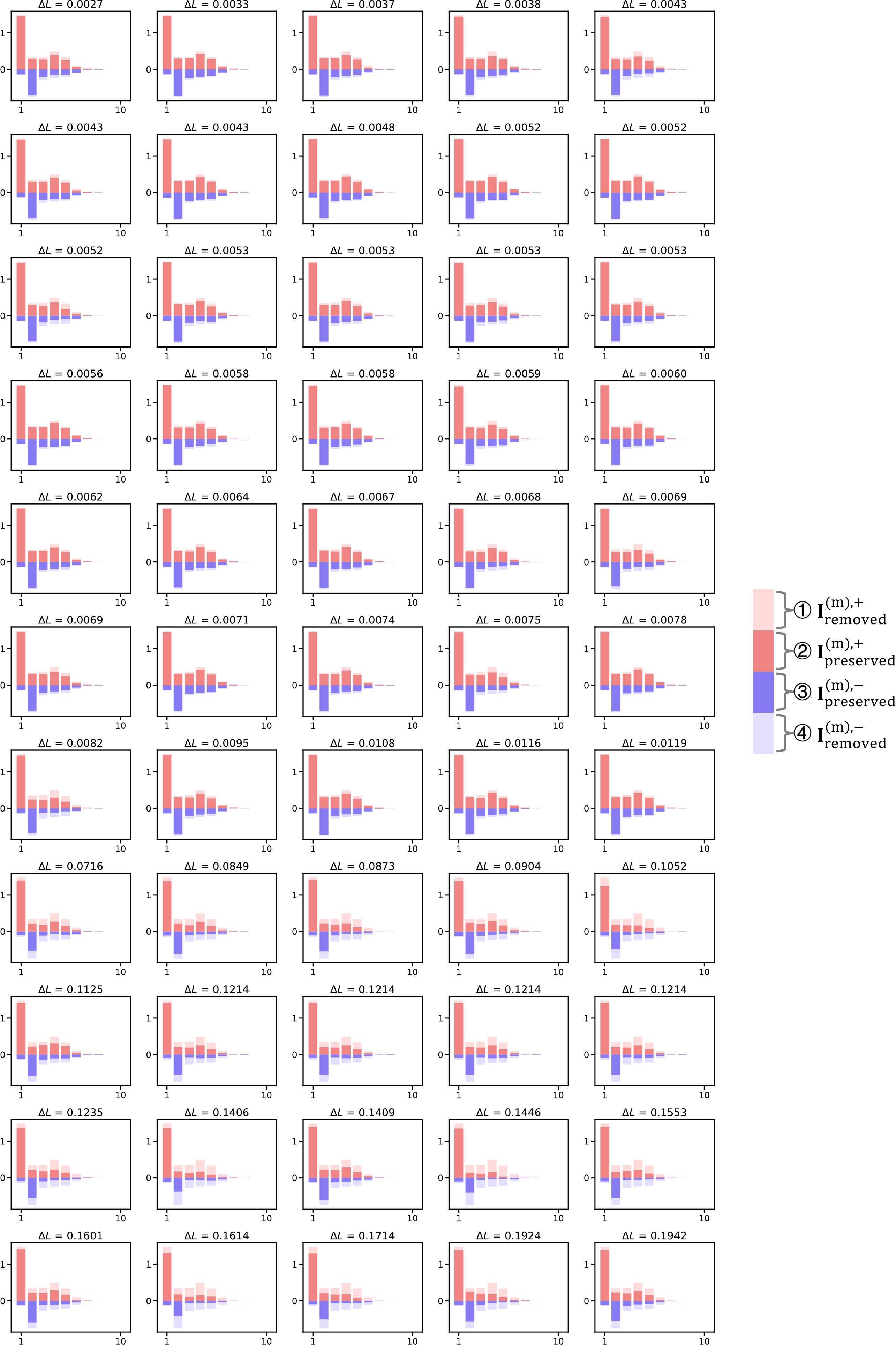}
  	\caption{The distributions of removed interactions ({\small $\dposR$} and {\small $\dnegR$}) and preserved interactions ({\small $\dposP$} and {\small $\dnegP$}) over different orders encoded by ResNet-152 after pruning each sensitive/insensitive module. The distributions are arranged in ascending order of the increase in test loss after pruning. For clarity, this figure presents the first of two subsets of results.}
  	\label{fig:distribution-resnet152-0}
\end{figure}

\begin{figure}[H]
  	\centering
  	\includegraphics[width=\dwidth]{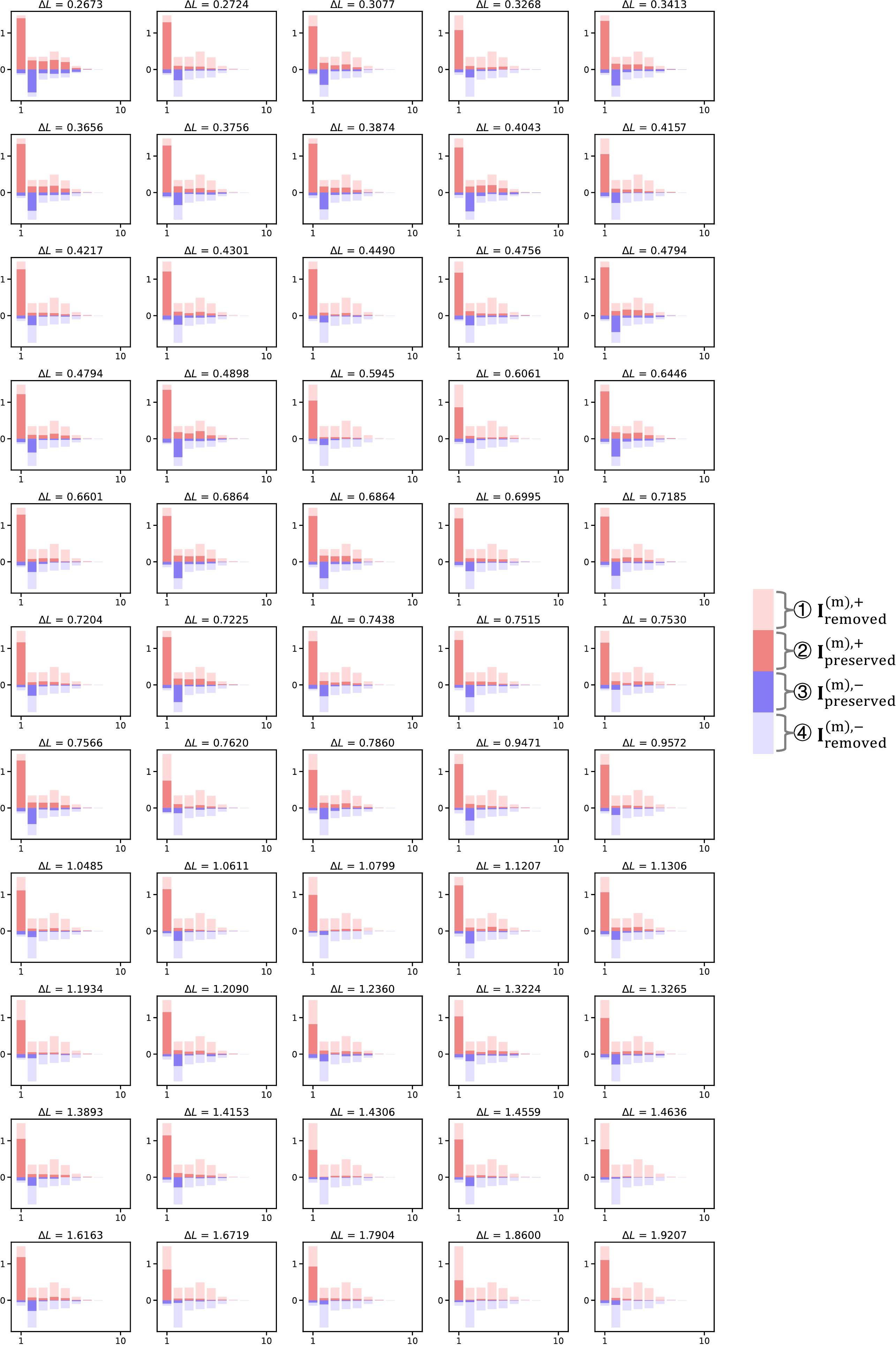}
  	\caption{The distributions of removed interactions ({\small $\dposR$} and {\small $\dnegR$}) and preserved interactions ({\small $\dposP$} and {\small $\dnegP$}) over different orders encoded by ResNet-152 after pruning each sensitive/insensitive module. The distributions are arranged in ascending order of the increase in test loss after pruning. For clarity, this figure presents the second of two subsets of results.}
  	\label{fig:distribution-resnet152-1}
\end{figure}

\subsubsection{The distribution of interactions encoded by ViT-Small after the pruning of each module}

\begin{figure}[H]
  	\centering
  	\includegraphics[width=\dwidth]{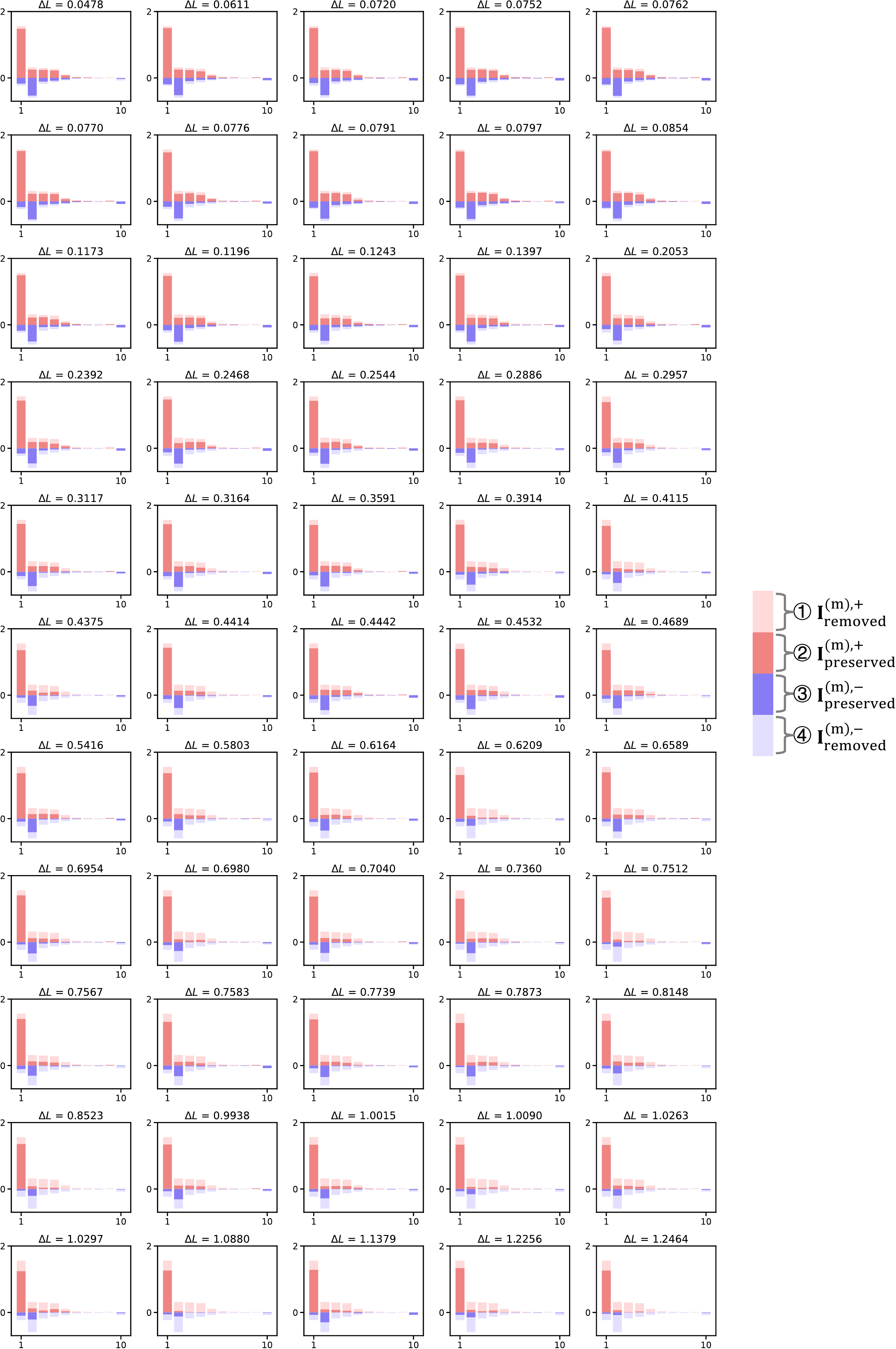}
  	\caption{The distributions of removed interactions ({\small $\dposR$} and {\small $\dnegR$}) and preserved interactions ({\small $\dposP$} and {\small $\dnegP$}) over different orders encoded by ViT-Small after pruning each sensitive/insensitive module. The distributions are arranged in ascending order of the increase in test loss after pruning. For clarity, this figure presents the first of two subsets of results.}
  	\label{fig:distribution-vit-small-0}
\end{figure}

\begin{figure}[H]
  	\centering
  	\includegraphics[width=\dwidth]{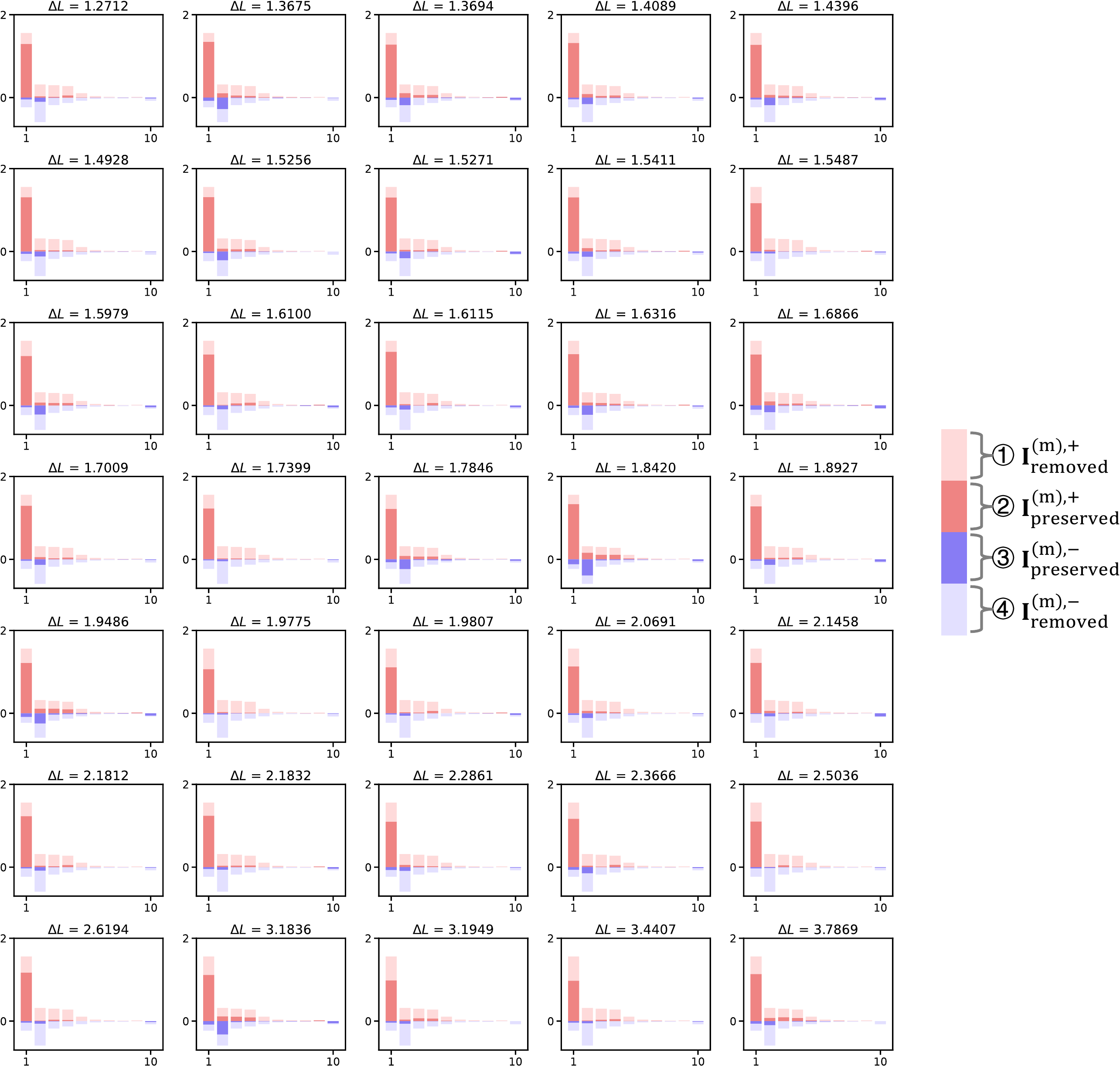}
  	\caption{The distributions of removed interactions ({\small $\dposR$} and {\small $\dnegR$}) and preserved interactions ({\small $\dposP$} and {\small $\dnegP$}) over different orders encoded by ViT-Small after pruning each sensitive/insensitive module. The distributions are arranged in ascending order of the increase in test loss after pruning. For clarity, this figure presents the second of two subsets of results.}
  	\label{fig:distribution-vit-small-1}
\end{figure}

\subsubsection{The distribution of interactions encoded by ViT-Base after the pruning of each module}

\begin{figure}[H]
  	\centering
  	\includegraphics[width=\dwidth]{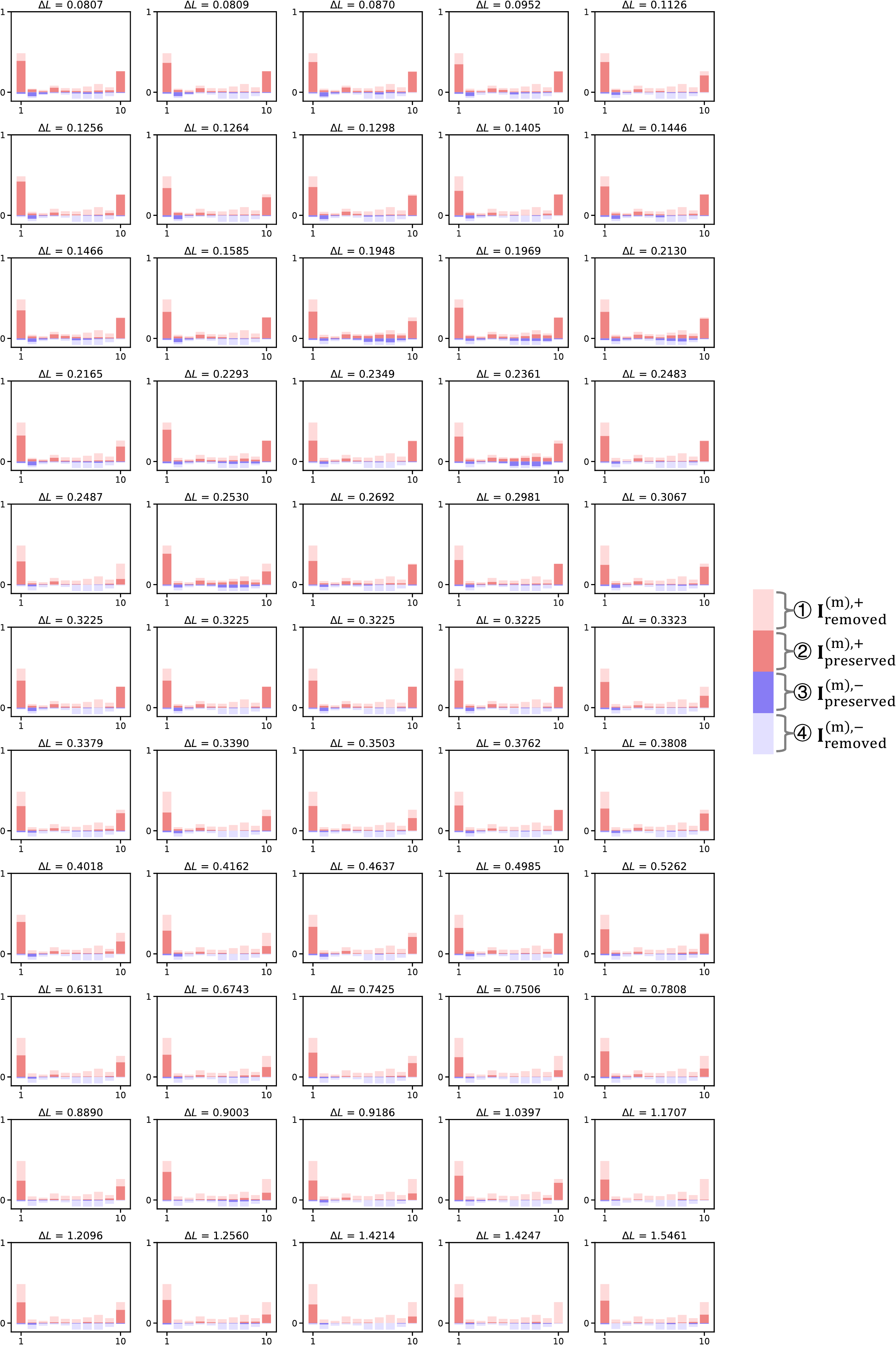}
  	\caption{The distributions of removed interactions ({\small $\dposR$} and {\small $\dnegR$}) and preserved interactions ({\small $\dposP$} and {\small $\dnegP$}) over different orders encoded by ViT-Base after pruning each sensitive/insensitive module. The distributions are arranged in ascending order of the increase in test loss after pruning. For clarity, this figure presents the first of two subsets of results.}
  	\label{fig:distribution-vit-base-0}
\end{figure}

\begin{figure}[H]
  	\centering
  	\includegraphics[width=\dwidth]{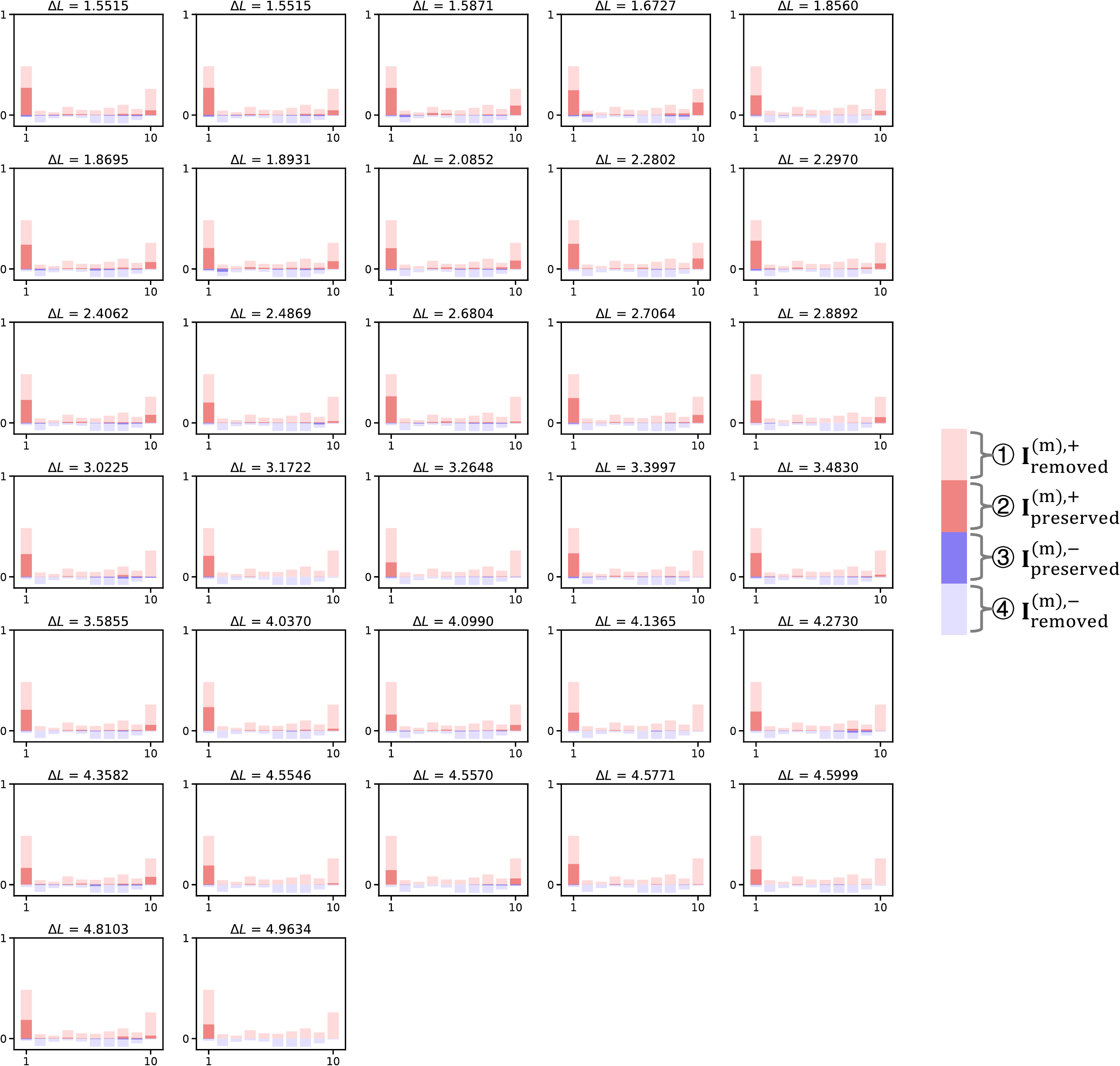}
  	\caption{The distributions of removed interactions ({\small $\dposR$} and {\small $\dnegR$}) and preserved interactions ({\small $\dposP$} and {\small $\dnegP$}) over different orders encoded by ViT-Base after pruning each sensitive/insensitive module. The distributions are arranged in ascending order of the increase in test loss after pruning. For clarity, this figure presents the second of two subsets of results.}
  	\label{fig:distribution-vit-base-1}
\end{figure}

\subsubsection{The distribution of interactions encoded by ViT-Large after the pruning of each module}

\begin{figure}[H]
  	\centering
  	\includegraphics[width=\dwidth]{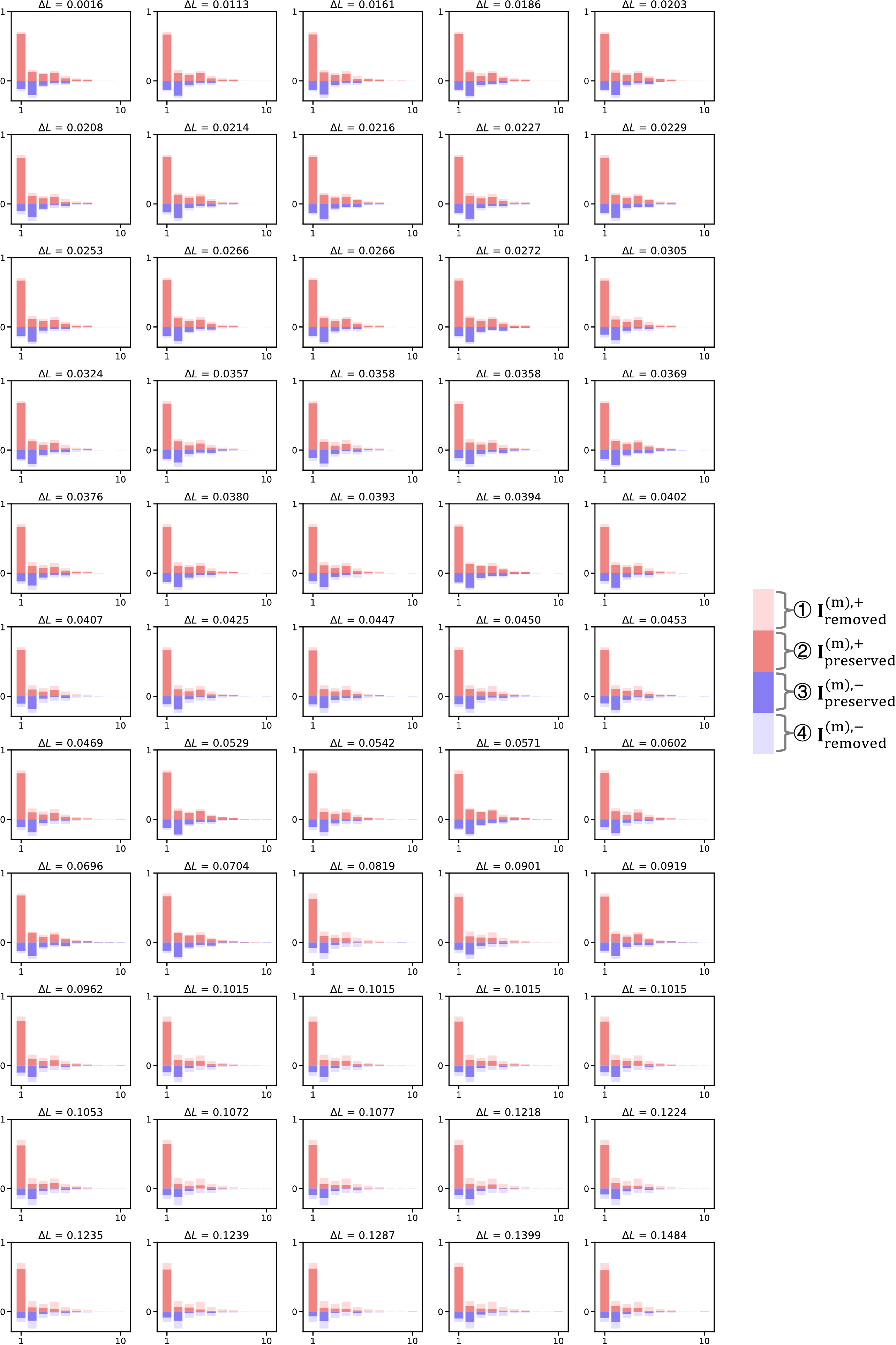}
  	\caption{The distributions of removed interactions ({\small $\dposR$} and {\small $\dnegR$}) and preserved interactions ({\small $\dposP$} and {\small $\dnegP$}) over different orders encoded by ViT-Large after pruning each sensitive/insensitive module. The distributions are arranged in ascending order of the increase in test loss after pruning. For clarity, this figure presents the first of two subsets of results.}
  	\label{fig:distribution-vit-large-0}
\end{figure}

\begin{figure}[H]
  	\centering
  	\includegraphics[width=\dwidth]{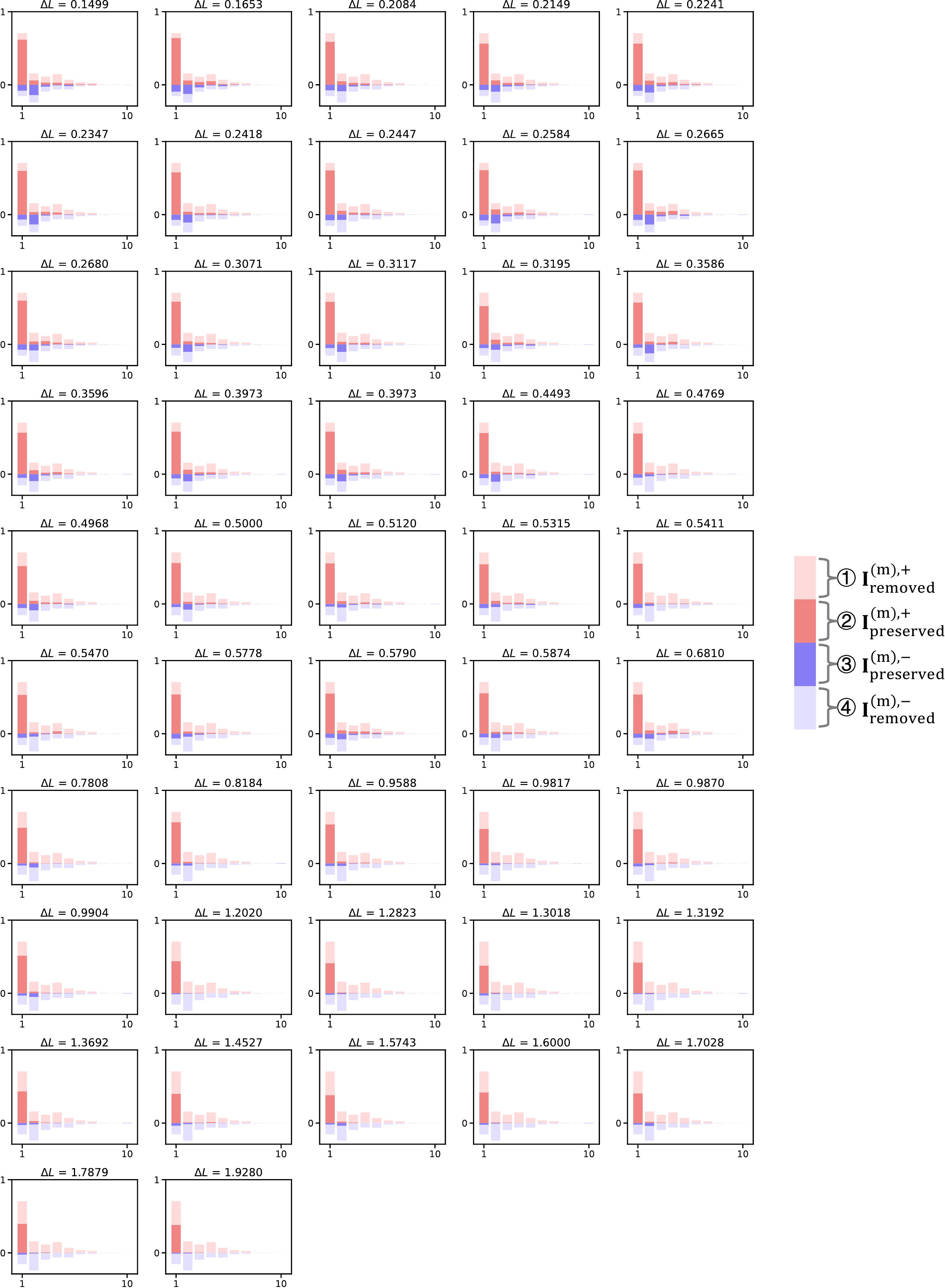}
  	\caption{The distributions of removed interactions ({\small $\dposR$} and {\small $\dnegR$}) and preserved interactions ({\small $\dposP$} and {\small $\dnegP$}) over different orders encoded by ViT-Large after pruning each sensitive/insensitive module. The distributions are arranged in ascending order of the increase in test loss after pruning. For clarity, this figure presents the second of two subsets of results.}
  	\label{fig:distribution-vit-large-1}
\end{figure}

\newpage
\subsection{Quantify the changes in interactions after the pruning of each module}
\label{subsec:apdx-metric-module-pruning-full}

We provide additional results quantifying the changes in interactions after pruning each module from multiple perspectives, including the ratio of preserved low-order/high-order interactions $\rhol$/$\rhoh$, the offsetting ratio of removed interaction effects $\offset$, and the ratio of removed generalizable interactions $\rhoR$.
In addition, the reported curves are obtained by applying the LOWESS smoothing algorithm~\citep{cleveland1979robust} to the individual data points.

\subsubsection{The ratio of preserved low-order/high-order interactions}

\begin{figure}[H]
  	\centering
  	\includegraphics[width=\textwidth]{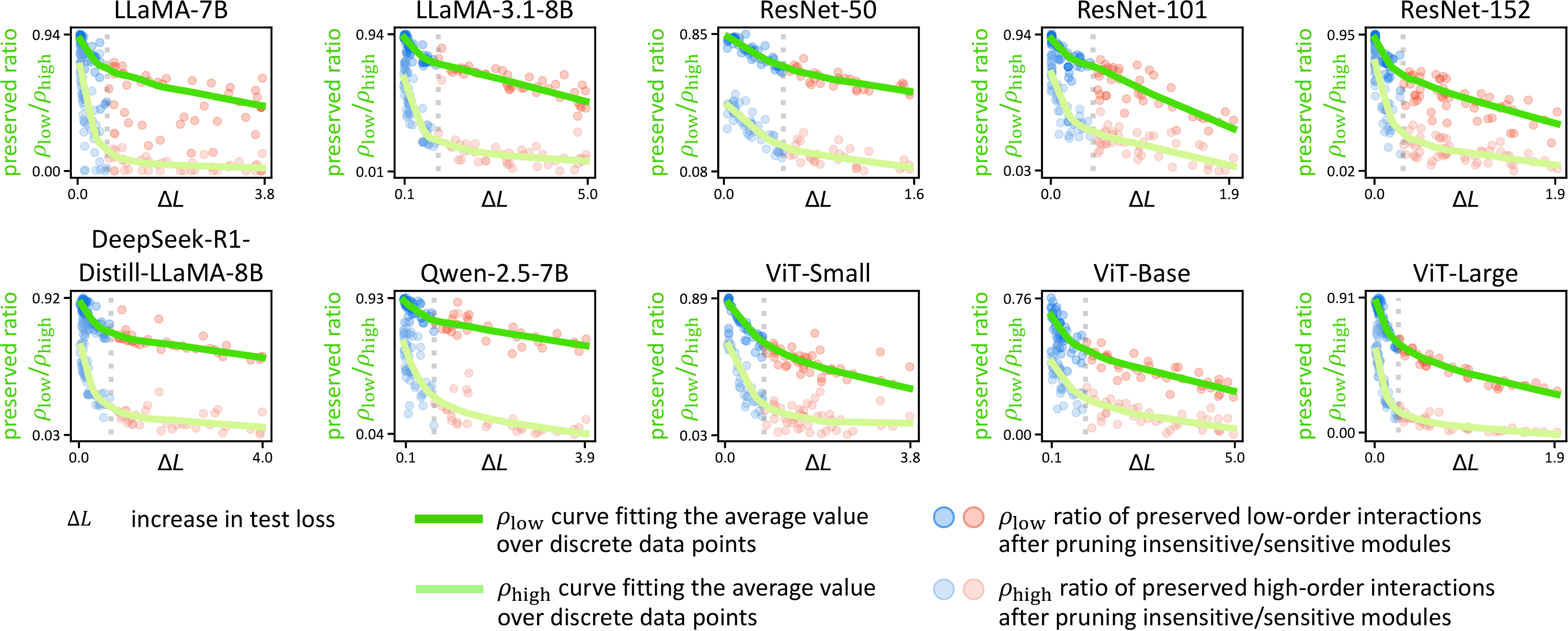}
  	\caption{The ratio of preserved low-order/high-order interactions $\rhol$/$\rhoh$ after pruning sensitive/insensitive modules on different DNNs.}
  	\label{fig:rho-module-pruning-full}
\end{figure}

Figure~\ref{fig:rho-module-pruning-full} shows the ratio of preserved low-order/high-order interactions $\rhol$/$\rhoh$ after pruning sensitive/insensitive modules on different DNNs.

\subsubsection{The offsetting ratio of removed interactions}

\begin{figure}[H]
  	\centering
  	\includegraphics[width=\textwidth]{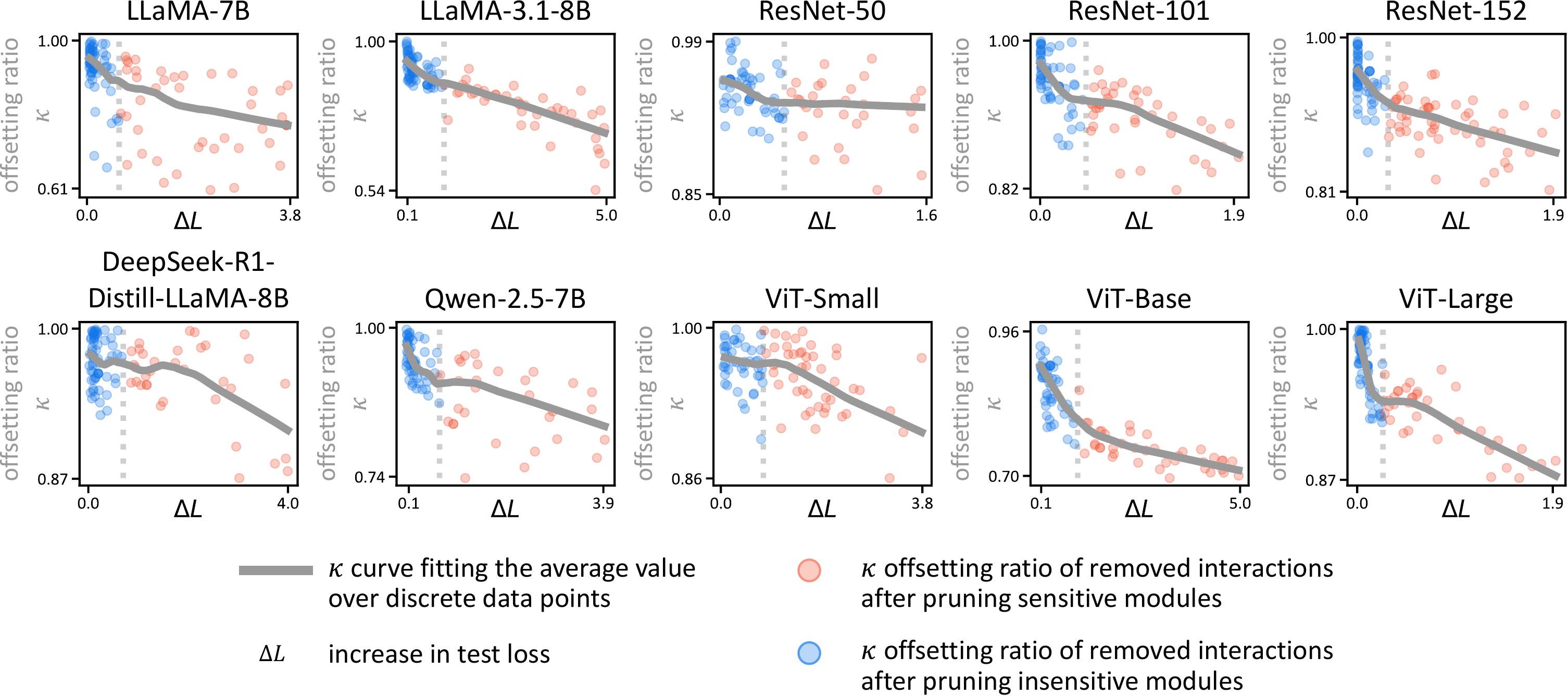}
  	\caption{The offsetting ratio of removed interactions $\offset$ after pruning sensitive/insensitive modules on different DNNs.}
  	\label{fig:kappa-module-pruning-full}
\end{figure}

Figure~\ref{fig:kappa-module-pruning-full} shows the offsetting ratio of removed interactions $\offset$ after pruning sensitive/insensitive modules on different DNNs.

\subsubsection{The ratio of removed generalizable interactions}

\begin{figure}[H]
  	\centering
  	\includegraphics[width=\textwidth]{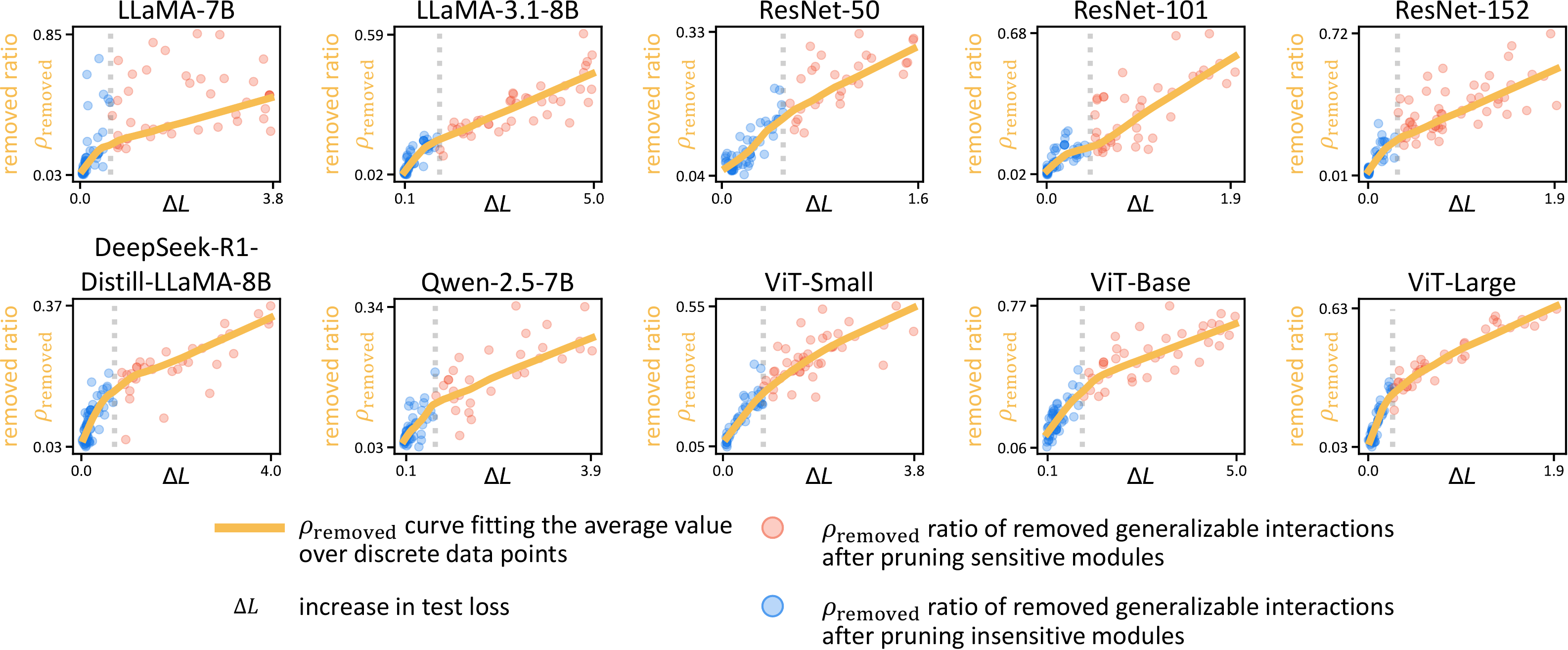}
  	\caption{The ratio of removed generalizable interactions $\rhoR$ after pruning sensitive/insensitive modules on different DNNs.}
  	\label{fig:rho_removed-module-pruning-full}
\end{figure}

Figure~\ref{fig:rho_removed-module-pruning-full} shows the ratio of removed generalizable interactions $\rhoR$ after pruning sensitive/insensitive modules on different DNNs.

\newpage

\newpage

\section{Distributions of generalizable interaction}
\label{sec:apdx-generalizable-distribution}

\begin{figure}[H]
  	\centering
  	\includegraphics[width=0.9\textwidth]{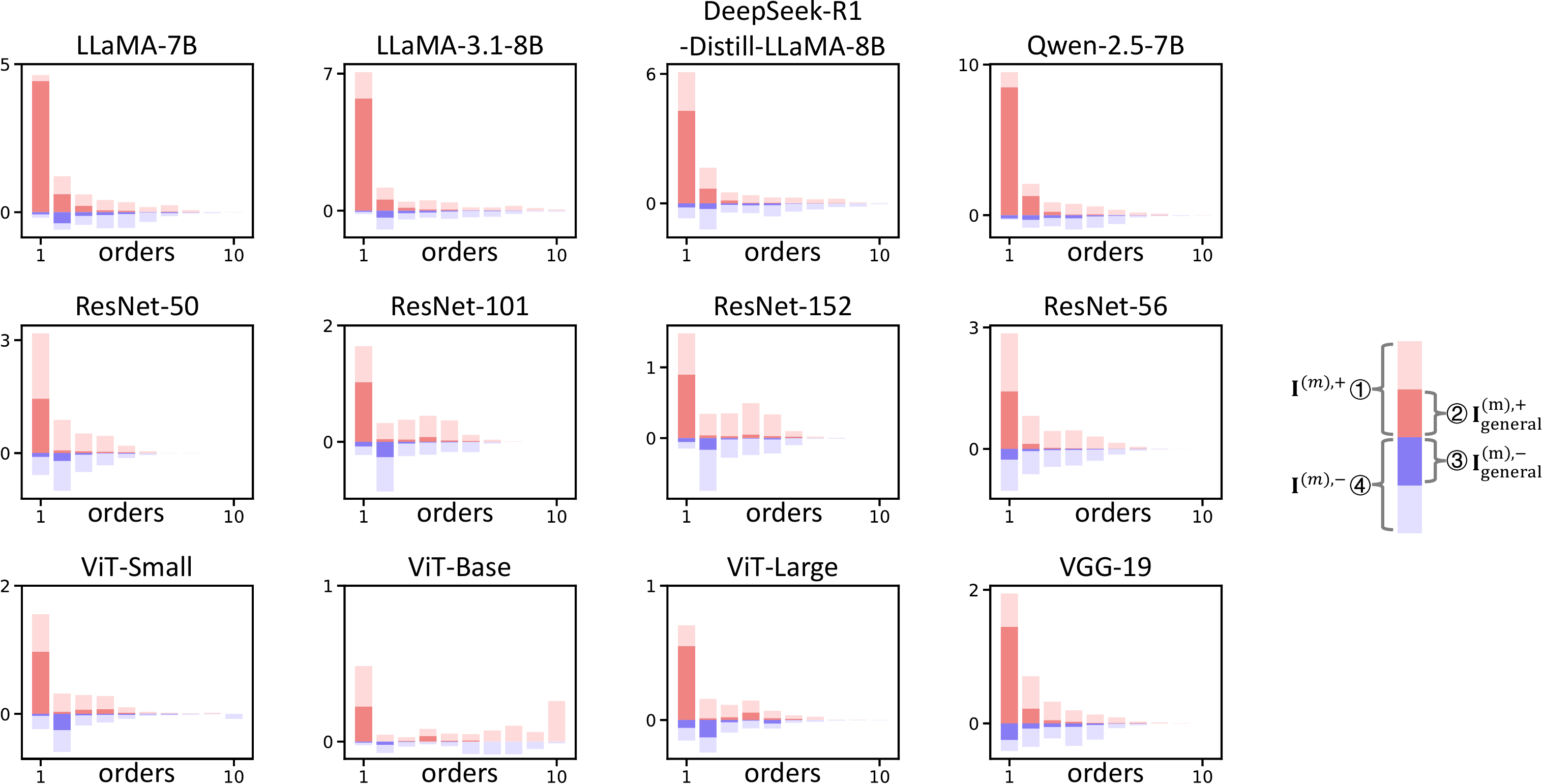}
  	\caption{The distributions of interactions and generalizable interactions {\small $\dpos$}, {\small $\dneg$}, {\small $\dposG$} and {\small $\dnegG$} encoded by twelve DNNs. Interactions of different orders exhibit different generalizability. Low-order interactions exhibit stronger generalizability than high-order ones.}
  	\label{fig:generalization-full}
\end{figure}

Figure~\ref{fig:generalization-full} shows the distributions of interactions and generalizable interactions {\small $\dpos$}, {\small $\dneg$}, {\small $\dposG$} and {\small $\dnegG$} encoded by twelve DNNs. Generalizable interactions are primarily concentrated at lower orders.

\textbf{Categorizing low-order and high-order interactions.}
In Figure~\ref{fig:generalization-full}, we observe a clear disparity in the generalizability of interactions across different orders. Specifically, nearly all generalizable interactions are concentrated within the 1st-3rd orders, whereas interactions of orders 4 to n are almost entirely non-generalizable. Motivated by this pronounced difference, we categorize interactions of orders 1-3 as low-order interactions, and those of orders 4 to n as high-order interactions.

\section{Offsetting of removed interaction effects}

\begin{figure}[H]
  	\centering
  	\includegraphics[width=0.7\textwidth]{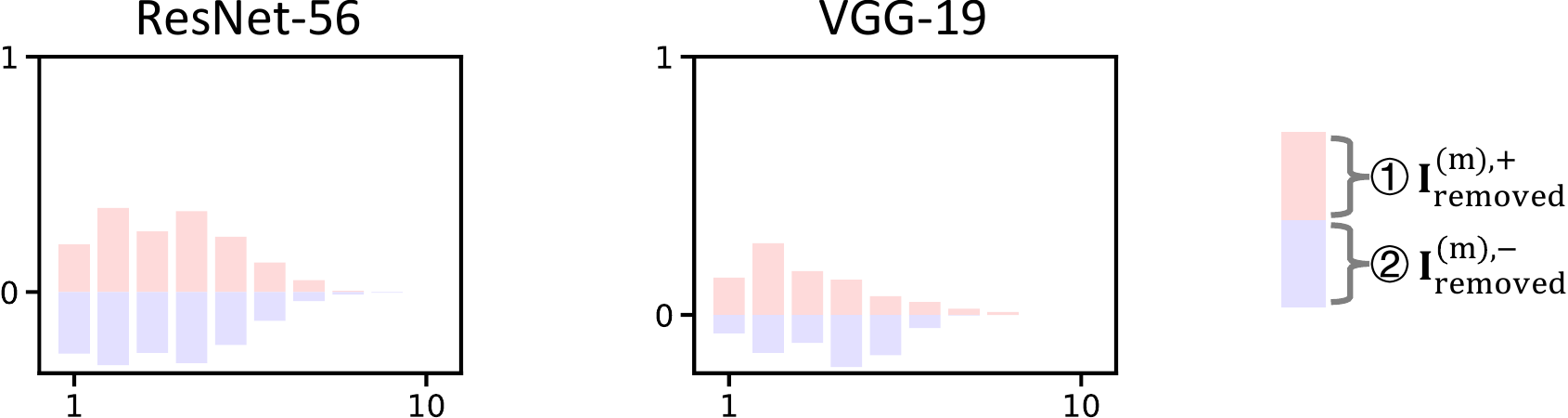}
  	\caption{The distributions of removed interactions {\small $\dposR$}, {\small $\dnegR$} at the end of Phase 1. Interactions of different orders exhibit different generalizability. Removed interactions exhibit strong offsetting effects.}
  	\label{fig:removed-interactions}
\end{figure}

Figure~\ref{fig:removed-interactions} shows the distributions of removed interactions {\small $\dposR$}, {\small $\dnegR$} encoded by ResNet-56 and VGG-19. Removed interactions exhibit strong offsetting effects.

\newpage


\section{Properties of the AND interaction}
\label{sec:apdx-property-harsanyi}

The Harsanyi interaction \cite{harsanyi} (referred to as the AND interaction in this work) has been a conventional metric for measuring the effect of the AND relationship that a DNN encodes among input variables. In this section, we introduce several desirable axioms that the AND interaction {\small$I^{\text{and}}_T$} adheres to. These properties further underscore the reliability of using AND interactions to explain the inference score of a DNN.

(1) \textit{Efficiency axiom} (proven by \cite{harsanyi}). The output score of a model can be decomposed into interaction effects of different patterns, \emph{i.e.} {\small $v(\boldsymbol{x})=\sum_{T\subseteq N}I^{\text{and}}_T$}.

(2) \textit{Linearity axiom}. If we merge output scores of two models $v_1$ and $v_2$ as the output of model $v$, \emph{i.e.} {\small $\forall S\subseteq N,~ v(\boldsymbol{x}_S)=v_1(\boldsymbol{x}_S)+v_2(\boldsymbol{x}_S)$}, then their interaction effects {\small $I^{\text{and}}_{T, v_1}$} and {\small $I^{\text{and}}_{T, v_2}$} can also be merged as {\small $\forall T\subseteq N, I^{\text{and}}_{T, v}=I^{\text{and}}_{T, v_1} + I^{\text{and}}_{T, v_2}$}.

(3) \textit{Dummy axiom}. If a variable {\small $i\in N$} is a dummy variable, \emph{i.e.} {\small $\forall S\subseteq N\setminus\{i\}, v(\boldsymbol{x}_{S\cup\{i\}})=v(\boldsymbol{x}_S)+v(\boldsymbol{x}_{\{i\}})$}, then it has no interaction with other variables, {\small $\forall \ \emptyset\not= T\subseteq N\setminus\{i\}$, $I^{\text{and}}_{T\cup\{i\}}=0$}.

(4) \textit{Symmetry axiom}. If input variables {\small $i,j\in N$} cooperate with other variables in the same way, {\small $\forall S\subseteq N\setminus\{i,j\}, v(\boldsymbol{x}_{S\cup\{i\}})=v(\boldsymbol{x}_{S\cup\{j\}})$}, then they have same interaction effects with other variables, {\small $\forall T\subseteq N\setminus\{i,j\}, I^{\text{and}}_{T\cup\{i\}} = I^{\text{and}}_{T\cup\{j\}}$}.

(5) \textit{Anonymity axiom}. For any permutations $\pi$ on {\small $N$}, we have {\small $\forall T \!\subseteq\! N, I^{\text{and}}_{T, v}=I^{\text{and}}_{\pi T, \pi v}$}, where {\small $\pi T \overset{\text{def}}{=} \{\pi(i) | i \in T\}$}, and the new model {\small $\pi v$} is defined by {\small $(\pi v)(\boldsymbol{x}_{\pi S}) = v(\boldsymbol{x}_S)$}. This indicates that interaction effects are not changed by permutation.

(6) \textit{Recursive axiom}. The interaction effects can be computed recursively. For {\small $i\in N$} and {\small $T\subseteq N\setminus\{i\}$}, the interaction effect of the pattern {\small $T\cup\{i\}$} is equal to the interaction effect of {\small $T$} with the presence of $i$ minus the interaction effect of $T$ with the absence of $i$, \emph{i.e.} {\small $\forall T\!\subseteq\! N\!\setminus\!\{i\}, I^{\text{and}}_{T\cup \{i\}}=I^{\text{and}}_{T, i\text{ present}} - I^{\text{and}}_T$}. {\small $I^{\text{and}}_{T, i\text{ present}}$} denotes the interaction effect when the variable $i$ is always present as a constant context, \emph{i.e.} {\small $I^{\text{and}}_{T, i\text{ present}}=\sum_{L\subseteq T} (-1)^{|T|-|L|}\cdot v(\boldsymbol{x}_{L\cup\{i\}})$}.

(7) \textit{Interaction distribution axiom}. This axiom characterizes how interactions are distributed for ``interaction functions''~\cite{sundararajan2020shapley}. An interaction function {\small $v_T$} parameterized by a subset of variables {\small $T$} is defined as follows. {\small $\forall S\subseteq N$}, if {\small $T\subseteq S$}, {\small$v_T(\boldsymbol{x}_S)=c$} ; otherwise, {\small $v_T(\boldsymbol{x}_S)=0$}. The function {\small$v_T$} models pure interaction among the variables in {\small$T$}, because only if all variables in {\small$T$} are present, the output value will be increased by {\small$c$}. The interactions encoded in the function {\small$v_T$} satisfies {\small $I^{\text{and}}_T=c$}, and {\small $\forall S\neq T$}, {\small $I^{\text{and}}_S=0$}.
\newpage

\section{Common conditions for sparse interactions}
\label{sec:apdx-condition-for-sparsity}

\cite{ren2024proving} have proved three sufficient conditions for the sparsity of AND interactions.

\textbf{Condition 1.} \textit{The DNN does not encode extremely high-order interactions: {\small$\forall \ T\in \{T\subseteq N \mid \vert T\vert \ge M+1\}, \ I^{\text{\rm and}}_T =0$}.}

Condition 1 is common because extremely high-order interactions usually represent very complex and over-fitted patterns, which are unlikely to be learned by a well-trained DNN in real scenarios.

\textbf{Condition 2.} \textit{Let {\small$\bar{u}^{(k)}\overset{\text{\rm def}}{=}\mathbb{E}_{|S|=k}[v(\boldsymbol{x}_S)-v(\boldsymbol{x}_\emptyset)]$} denote the average classification confidence of the DNN over all masked samples $\boldsymbol{x}_S$ with $k$ unmasked input variables. This average classification confidence monotonically increases when $k$ increases: $\forall \ k' \le k$, {\small$\bar{u}^{(k')} \le \bar{u}^{(k)}$}.}

Condition 2 implies that a well-trained DNN is likely to have higher average classification confidence for less masked input samples.

\textbf{Condition 3.} \textit{Given the average classification confidence $\bar{u}^{(k)}$ of samples with $k$ unmasked input variables, there is a polynomial lower bound for the average classification confidence with $k' (k'\le k)$ unmasked input variables: {\small $\forall \ k' \le k, \ \bar{u}^{(k')} \ge (\frac{k'}{k})^p \ \bar{u}^{(k)}$}, where $p>0$ is a constant.}

Condition 3 suggests that the classification confidence of the DNN remains relatively stable even when presented with masked input samples. In real-world applications, the classification or detection of masked or occluded samples frequently occurs. As a result, a well-trained DNN typically develops the ability to classify such masked inputs by leveraging local information, which can be derived from the visible portions of the input. Consequently, the model should not produce a substantially reduced confidence score for masked samples.
\newpage

\section{Empirical validation of sparsity property}
\label{sec:apdx-exp-sparsity}

As defined by Equation~\ref{eq:logic}, the AND-OR logical model comprises a set of AND interactions and a set of OR interactions, denoted as $\omegaand$ and $\omegaor$. The primary objective of this experiment is to demonstrate the sparsity property of these interactions, that is, to validate that both $\omegaand$ and $\omegaor$ are sufficiently small relative to the total possible $2^n$ interactions.

\begin{figure}[h]
\centering
\includegraphics[width=0.8\linewidth]{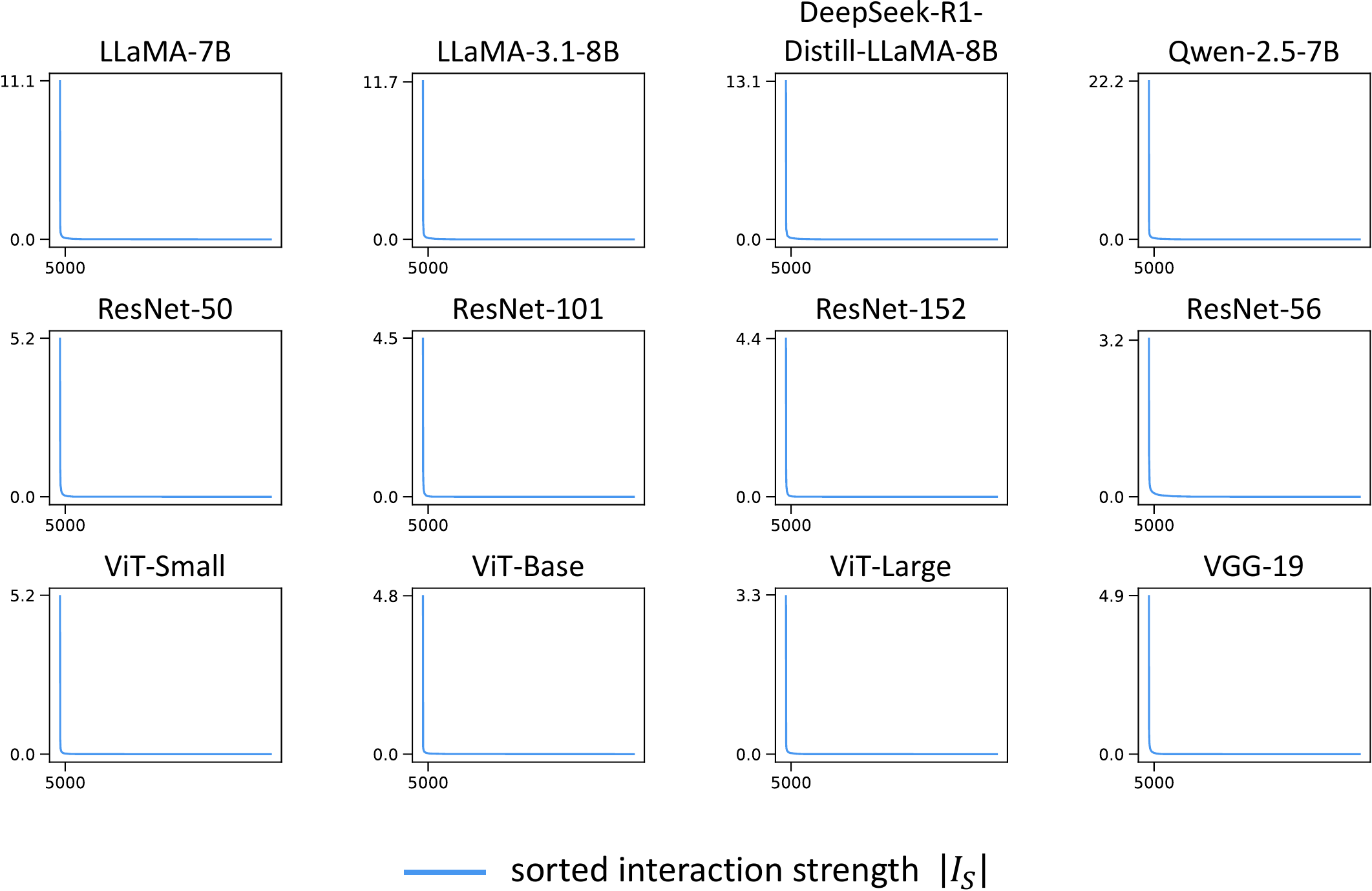}
\caption{Sorted interaction strength distribution on different input samples. Most interactions exhibit negligible strength.}
\label{fig:apdx-spa}
\end{figure}

We conducted experiments to validate the sparsity property of interactions. Given a well-trained DNN and an input sample $\x$, we calculated AND interactions {\small$I_S^\text{AND}$} and OR interactions {\small$I_S^\text{OR}$} for all {\small$2^{n}$} possible subsets. To this end, we employed the same network architecture and dataset described in Appendix~\ref{sec:experimental_setting}. Figure~\ref{fig:apdx-spa} shows the strength of all AND-OR interactions $\vert I_S \vert$ extracted from 100 different samples in descending order. The results demonstrate that the size of the interaction set is relatively small, typically no more than 5000 for all 100 sample (\emph{i.e.}, 50 for each sample), across different networks and tasks.

\newpage

\section{Proof of Equation \ref{eq:logic}}
\label{proof:apdx-match}

\begin{proof} \textbf{(1) Universal matching property of AND interactions.}

We will prove that output component \( u^{\text{AND}}_S \) on all \( 2^n \) masked samples \( \{\boldsymbol{x}_S:S\subseteq N\} \) could be universally explained by all interactions in \( S\subseteq N \), \emph{i.e.}, \( \forall \emptyset \neq S\subseteq N, u^{\text{AND}}_S = \sum_{\emptyset \neq T\subseteq S} I^{\text{AND}}_T + v(\boldsymbol{x}_\emptyset) \). In particular, we define \( u^{\text{AND}}_\emptyset = v(\boldsymbol{x}_\emptyset) \) (\emph{i.e.}, we attribute output on an empty sample to AND interactions).

Specifically, the AND interaction is defined as \( I^{\text{AND}}_T = \sum\nolimits_{L \subseteq T} (-1)^{|T|-|L|} u^{\text{AND}}_L \). 
To compute the sum of AND interactions \( \sum_{\emptyset \neq T\subseteq S} I^{\text{AND}}_T = \sum\nolimits_{\emptyset \neq T \subseteq S} \sum\nolimits_{L \subseteq T} (-1)^{\vert T \vert - \vert L \vert} u^{\text{AND}}_L \), we first exchange the order of summation of the set \( L\subseteq T\subseteq S \) and the set \( T \supseteq L \). 
That is, we compute all linear combinations of all sets \( T \) containing \( L \) with respect to the model outputs \( u^{\text{AND}}_L \) given a set of input variables \( L \), \emph{i.e.}, \( \sum\nolimits_{T: L \subseteq T \subseteq S} (-1)^{|T|-|L|} u^{\text{AND}}_L \). 
Then, we compute all summations over the set \( L\subseteq S \).

In this way, we can compute them separately for different cases of \( L\subseteq T\subseteq S \). In the following, we consider the cases (1) \( L = S = T \), and (2) \( L\subseteq T\subseteq S, L\ne S \), respectively.

(1) When \( L=S=T \), the linear combination of all subsets \( T \) containing \( L \) with respect to the model output \( u^{\text{AND}}_L \) is \( (-1)^{|S|-|S|} u^{\text{AND}}_L = u^{\text{AND}}_L \).

(2) When \( L\subseteq T\subseteq S, L\ne S \), the linear combination of all subsets \( T \) containing \( L \) with respect to the model output \( u^{\text{AND}}_L \) is \( \sum\nolimits_{T: L \subseteq T \subseteq S} (-1)^{|T|-|L|} u^{\text{AND}}_L \). For all sets \( T: S\supseteq T\supseteq L \), let us consider the linear combinations of all sets \( T \) with number \( |T| \) for the model output \( u^{\text{AND}}_L \), respectively. Let \( m := |T| - |L| \), (\( 0\le m\le |S|-|L| \)), then there are a total of \( C_{|S|-|L|}^{m} \) combinations of all sets \( T \) of order \( |T| \). Thus, given \( L \), accumulating the model outputs \( u^{\text{AND}}_L \) corresponding to all \( T\supseteq L \), then \( \sum\nolimits_{T: L \subseteq T \subseteq S} (-1)^{|T|-|L|} u^{\text{AND}}_L = u^{\text{AND}}_L \cdot \underbrace{\sum\nolimits_{m=0}^{\vert S \vert - \vert L \vert} C_{|S|-|L|}^m (-1)^m}_{=0} = 0 \). Please see the complete derivation of the following formula.

\begin{equation}\begin{aligned}
    \sum\nolimits_{\emptyset \neq T \subseteq S} I^{\text{AND}}_T
    = &  \sum\nolimits_{\emptyset \neq T \subseteq S} \sum\nolimits_{L \subseteq T} (-1)^{\vert T \vert - \vert L \vert} u^{\text{AND}}_L \\
    = & \sum\nolimits_{L \subseteq S} \sum\nolimits_{T: L \subseteq T \subseteq S} (-1)^{\vert T \vert - \vert L \vert} u^{\text{AND}}_L  - u^{\text{AND}}_\emptyset \\
    = & \underbrace{u^{\text{AND}}_S}_{L = S} + \sum\nolimits_{L \subseteq S, L \neq S} u^{\text{AND}}_L \cdot \underbrace{\sum\nolimits_{m=0}^{\vert S \vert - \vert L \vert} C_{|S|-|L|}^m (-1)^m}_{=0}   - u^{\text{AND}}_\emptyset \\
     = & u^{\text{AND}}_S  - u^{\text{AND}}_\emptyset  = u^{\text{AND}}_S  - v(\boldsymbol{x}_\emptyset)
\end{aligned}\end{equation}

Thus, we have \( \forall \emptyset \neq S\subseteq N, u^{\text{and}}_S = \sum_{\emptyset \neq T\subseteq S} I^{\text{and}}_T + v(\boldsymbol{x}_\emptyset) \).

\textbf{(2) Universal matching property of OR interactions.}

According to the definition of OR interactions, we will derive that \( \forall S\subseteq N, u^{\text{OR}}_S = \sum_{T:T\cap S\neq \emptyset} I^{\text{OR}}_T \), 
where we define \( u^{\text{OR}}_\emptyset = 0 \) (recall that in Step (1), we attribute the output on empty input to AND interactions).

Specifically, the OR interaction is defined as \( I^{\text{OR}}_T = -\sum\nolimits_{L \subseteq T} (-1)^{|T|-|L|} u^{\text{OR}}_{N\setminus L} \).
Similar to the above derivation of the universal matching theorem of AND interactions, to compute the sum of OR interactions \( \sum\nolimits_{T:T \cap S \neq \emptyset} I^{\text{OR}}_T = \sum\nolimits_{T:T \cap S \neq \emptyset} \left[- \sum\nolimits_{L \subseteq T} (-1)^{\vert T \vert - \vert L \vert} u^{\text{OR}}_{N \setminus L} \right] \), we first exchange the order of summation of the set \( L\subseteq T \subseteq N \) and the set \( T:T \cap S \neq \emptyset \). That is, we compute all linear combinations of all sets \( T \) containing \( L \) with respect to the model outputs \( u^{\text{OR}}_{N \setminus L} \) given a set of input variables \( L \), \emph{i.e.}, \( \sum\nolimits_{T: T \cap S \neq \emptyset, T \supseteq L} (-1)^{\vert T \vert - \vert L \vert} u^{\text{OR}}_{N \setminus L} \). Then, we compute all summations over the set \( L\subseteq N \).

In this way, we can compute them separately for different cases of \( L\subseteq T\subseteq N, T \cap S \neq \emptyset \). In the following, we consider the cases (1) \( L = N \setminus S \), (2) \( L=N \), (3) \( L \cap S \neq \emptyset, L \neq N \), and (4) \( L \cap S=\emptyset, L \neq N \setminus S \), respectively.

(1) When \( L = N \setminus S \), the linear combination of all subsets \( T \) containing \( L \) with respect to the model output \( u^{\text{OR}}_{N \setminus L} \) is \( \sum\nolimits_{T: T \cap S \neq \emptyset, T \supseteq L} (-1)^{\vert T \vert - \vert L \vert} u^{\text{OR}}_{N \setminus L} = \sum\nolimits_{T: T \cap S \neq \emptyset, T \supseteq L} (-1)^{\vert T \vert - \vert L \vert} u^{\text{OR}}_S \). For all sets \( T: T\supseteq L, T \cap S \neq \emptyset \) (then \( T \neq N \setminus S, T \neq L \)), let us consider the linear combinations of all sets \( T \) with number \( |T| \) for the model output \( u^{\text{OR}}_S \), respectively. Let \( |T'| := |T| - |L| \), (\( 1\le |T'|\le |S| \)), then there are a total of \( C_{|S|}^{|T'|} \) combinations of all sets \( T' \) of order \( |T'| \). 
Thus, given \( L \), accumulating the model outputs \( u^{\text{OR}}_S \) corresponding to all \( T\supseteq L \), then \( \sum\nolimits_{T: T \cap S \neq \emptyset, T \supseteq L} (-1)^{\vert T \vert - \vert L \vert} u^{\text{OR}}_{N \setminus L} = u^{\text{OR}}_S \cdot \underbrace{\sum\nolimits_{|T'|=1}^{\vert S \vert } C_{|S|}^{|T'|} (-1)^{|T'|}}_{=-1} = -u^{\text{OR}}_S \).

(2) When \( L=N \) (then \( T=N \)), the linear combination of all subsets \( T \) containing \( L \) with respect to the model output \( u^{\text{OR}}_{N \setminus L} \) is \( \sum\nolimits_{T: T \cap S \neq \emptyset, T \supseteq L} (-1)^{\vert T \vert - \vert L \vert} u^{\text{OR}}_{N \setminus L} = (-1)^{\vert N \vert - \vert N \vert} u^{\text{OR}}_\emptyset = u^{\text{OR}}_\emptyset \).

(3) When \( L \cap S \neq \emptyset, L \neq N \), the linear combination of all subsets \( T \) containing \( L \) with respect to the model output \( u^{\text{OR}}_{N \setminus L} \) is \( \sum\nolimits_{T: T \cap S \neq \emptyset, T \supseteq L} (-1)^{\vert T \vert - \vert L \vert} u^{\text{OR}}_{N \setminus L} \). For all sets \( T: T\supseteq L, T \cap S \neq \emptyset \), let us consider the linear combinations of all sets \( T \) with number \( |T| \) for the model output \( u^{\text{OR}}_S \), respectively. Let us split \( |T| - |L| \) into \( |T'| \) and \( |T''| \), \emph{i.e.}, \( |T| - |L| = |T'| + |T''| \), where \( T'=\{i|i\in T, i\notin L, i\in N\setminus S\} \), \( T''=\{i|i\in T, i\notin L, i\in S\} \) (then \( 0\le|T''|\le|S|-|S\cap L| \)) and \( |T'| + |T''| + |L| = |T| \). In this way, there are a total of \( C_{|S|-|S\cap L|}^{|T''|} \) combinations of all sets \( T'' \) of order \( |T''| \). Thus, given \( L \), accumulating the model outputs \( u^{\text{OR}}_{N\setminus L} \) corresponding to all \( T\supseteq L \), then \( \sum\nolimits_{T: T \cap S \neq \emptyset, T \supseteq L} (-1)^{\vert T \vert - \vert L \vert} u^{\text{OR}}_{N \setminus L} = u^{\text{OR}}_{N \setminus L} \cdot \sum_{T' \subseteq N\setminus S \setminus L} \underbrace{\sum\nolimits_{\vert T'' \vert = 0}^{\vert S \vert-\vert S \cap L \vert} C_{\vert S \vert - \vert S \cap L \vert}^{\vert T''\vert } (-1)^{\vert T' \vert + \vert T'' \vert} }_{=0} = 0 \).

(4) When \( L \cap S=\emptyset, L \neq N \setminus S \), the linear combination of all subsets \( T \) containing \( L \) with respect to the model output \( u^{\text{OR}}_{N \setminus L} \) is \( \sum\nolimits_{T: T \cap S \neq \emptyset, T \supseteq L} (-1)^{\vert T \vert - \vert L \vert} u^{\text{OR}}_{N \setminus L} \). Similarly, let us split \( |T| - |L| \) into \( |T'| \) and \( |T''| \), \emph{i.e.}, \( |T| - |L| = |T'| + |T''| \), where \( T'=\{i|i\in T, i\notin L, i\in N\setminus S\} \), \( T''=\{i|i\in T, i\in S\} \) (then \( 0\le|T''|\le|S| \)) and \( |T'| + |T''| + |L| = |T| \). In this way, there are a total of \( C_{|S|}^{|T''|} \) combinations of all sets \( T'' \) of order \( |T''| \). Thus, given \( L \), accumulating the model outputs \( u^{\text{OR}}_{N\setminus L} \) corresponding to all \( T\supseteq L \), then \( \sum\nolimits_{T: T \cap S \neq \emptyset, T \supseteq L} (-1)^{\vert T \vert - \vert L \vert} u^{\text{OR}}_{N \setminus L} = u^{\text{OR}}_{N \setminus L} \cdot \sum_{T' \subseteq N\setminus S \setminus L} \underbrace{\sum\nolimits_{\vert T'' \vert = 0}^{\vert S \vert} C_{\vert S \vert }^{\vert T''\vert } (-1)^{\vert T' \vert + \vert T'' \vert} }_{=0} = 0 \). 

Please see the complete derivation of the following formula.
\small{

\begin{equation}
    \begin{aligned}
        \sum\nolimits_{T:T \cap S \neq \emptyset} I^{\text{OR}}_T
        &= \sum\nolimits_{T:T \cap S \neq \emptyset} \left[- \sum\nolimits_{L \subseteq T} (-1)^{\vert T \vert - \vert L \vert} u^{\text{OR}}_{N \setminus L} \right]\\
        &= - \sum\nolimits_{L \subseteq N} \sum\nolimits_{T: T \cap S \neq \emptyset, T \supseteq L} (-1)^{\vert T \vert - \vert L \vert} u^{\text{OR}}_{N \setminus L} \\
        &=  - \left[\sum_{\vert T' \vert = 1}^{\vert S \vert} C_{\vert S \vert}^{\vert T' \vert} (-1)^{\vert T' \vert} \right] \cdot \underbrace{u^{\text{OR}}_S}_{L=N\setminus S} - \underbrace{u^{\text{OR}}_\emptyset}_{L=N} \\
        &\quad- \sum_{L \cap S \neq \emptyset, L \neq N} \left[\sum_{T' \subseteq N\setminus S \setminus L} \left( \sum_{\vert T'' \vert = 0}^{\vert S \vert-\vert S \cap L \vert} C_{\vert S \vert - \vert S \cap L \vert}^{\vert T''\vert } (-1)^{\vert T' \vert + \vert T'' \vert} \right) \right]\cdot u^{\text{OR}}_{N \setminus L}  \\
        &\quad- \sum_{L \cap S=\emptyset, L \neq N \setminus S} \left[ \sum_{T' \subseteq N\setminus S \setminus L} \left( \sum_{\vert T'' \vert=0}^{\vert S \vert} C_{\vert S \vert}^{\vert T'' \vert} (-1)^{\vert T' \vert + \vert T'' \vert}\right) \right] \cdot u^{\text{OR}}_{N \setminus L}  \\
        &=  - (-1) \cdot u^{\text{OR}}_S - u^{\text{OR}}_\emptyset - \sum_{L \cap S \neq \emptyset, L \neq N} \left[\sum_{T' \subseteq N\setminus S \setminus L} 0 \right]\cdot u^{\text{OR}}_{N \setminus L}  \\
        &\quad- \sum_{L \cap S=\emptyset, L \neq N \setminus S}\left[\sum_{T' \subseteq N\setminus S \setminus L} 0 \right] \cdot u^{\text{OR}}_{N \setminus L}  \\
        &= u^{\text{OR}}_S - u^{\text{OR}}_\emptyset\\
        &= u^{\text{OR}}_S
    \end{aligned}
\end{equation}
}

\textbf{(3) Universal matching property of AND-OR interactions.}

With the universal matching theorem of AND interactions and the universal matching theorem of OR interactions, we can easily get \( v(\boldsymbol{x}_S) = u^{\text{and}}_S + u^{\text{OR}}_S 
= v(\boldsymbol{x}_\emptyset) + \sum_{\emptyset \neq T\subseteq S} I^{\text{and}}_T + \sum_{T: T\cap S \neq \emptyset} I^{\text{OR}}_T \), thus, we obtain the universal matching theorem of AND-OR interactions.

\end{proof}

\newpage

\section{Illustration of And-Or logical model}
\label{sec:apdx-demo-model}

The following four figures show the logical models that mathematically represent the inference logic of DeepSeek-R1-Distill-LLaMA-8B model and Qwen-2.5-7B on two input prompts, respectively.

\begin{figure}[H]
  	\centering
  	\includegraphics[width=0.86\textwidth]{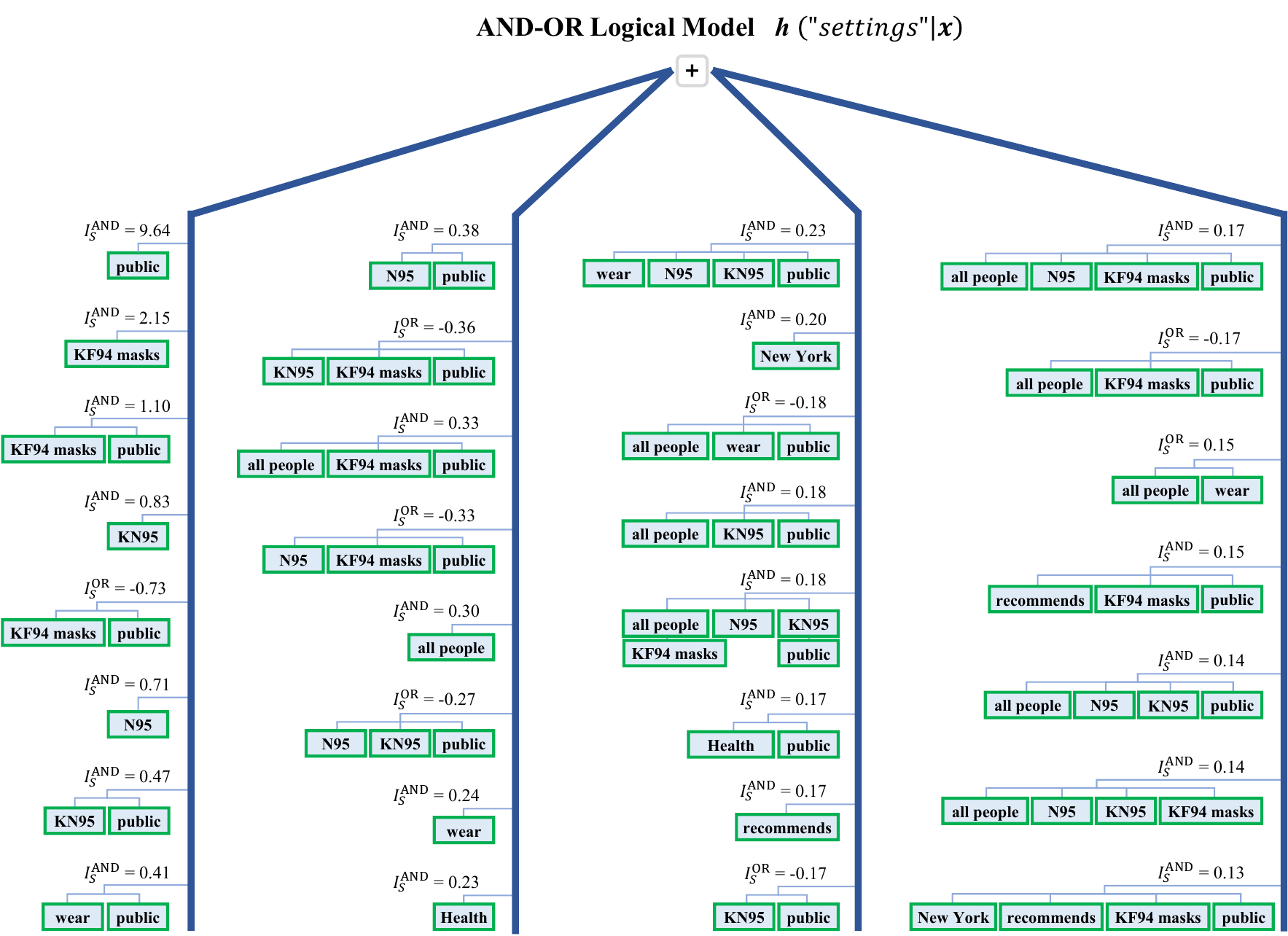}
  	\caption{The logical model representing the inference logic of the DeepSeek-R1-Distill-LLaMA-8B model on the input prompt ``New York Department of Health recommends that all people should wear N95, KN95, or KF94 masks in all public.'' The predicted next word is ``settings.''}
  	\label{fig:aog_0}
\end{figure}

\begin{figure}[H]
  	\centering
  	\includegraphics[width=0.86\textwidth]{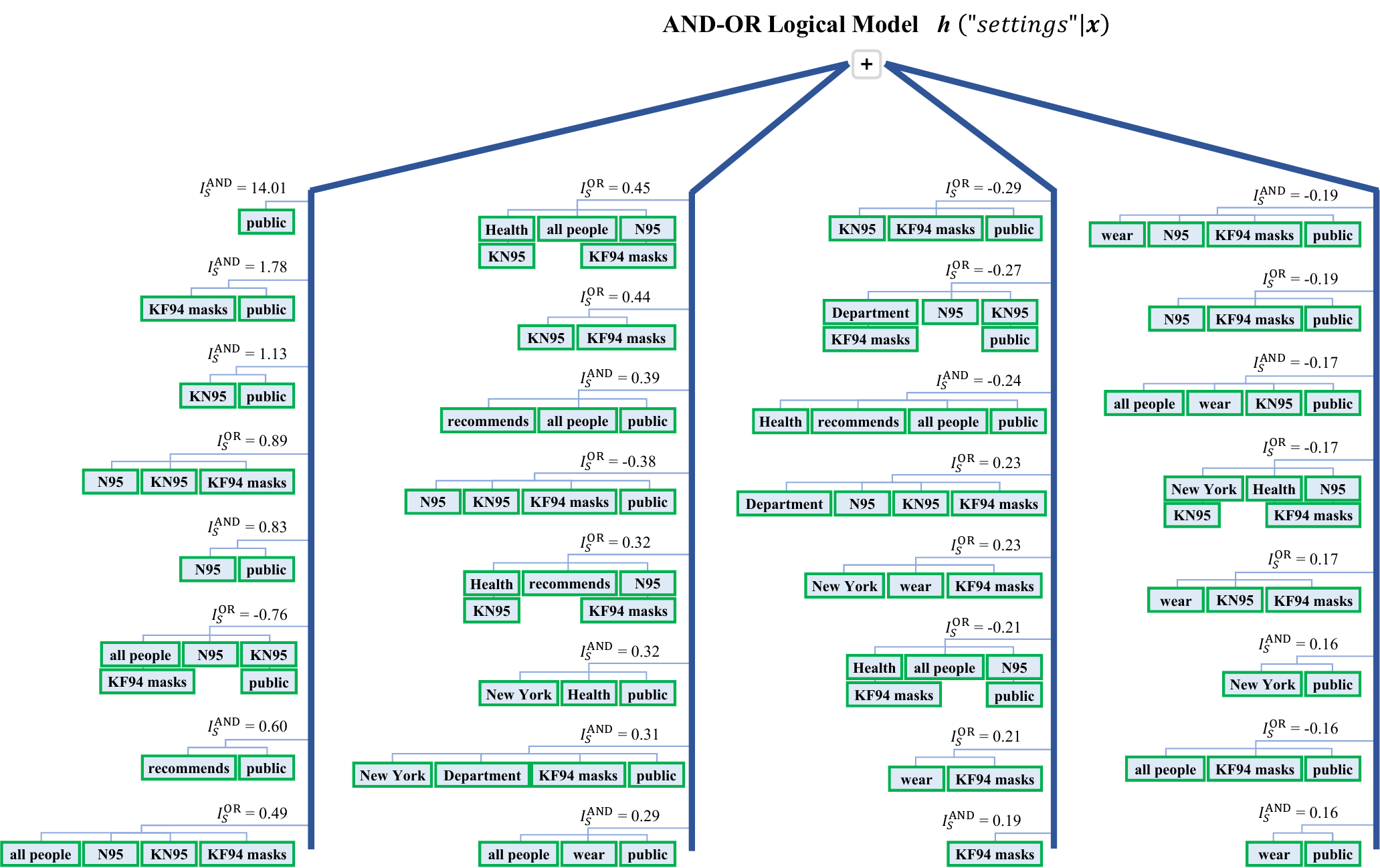}
  	\caption{The logical model representing the inference logic of the Qwen-2.5-7B model on the input prompt ``New York Department of Health recommends that all people should wear N95, KN95, or KF94 masks in all public.'' The predicted next word is ``settings.''}
  	\label{fig:aog_1}
\end{figure}

\begin{figure}[H]
  	\centering
  	\includegraphics[width=0.86\textwidth]{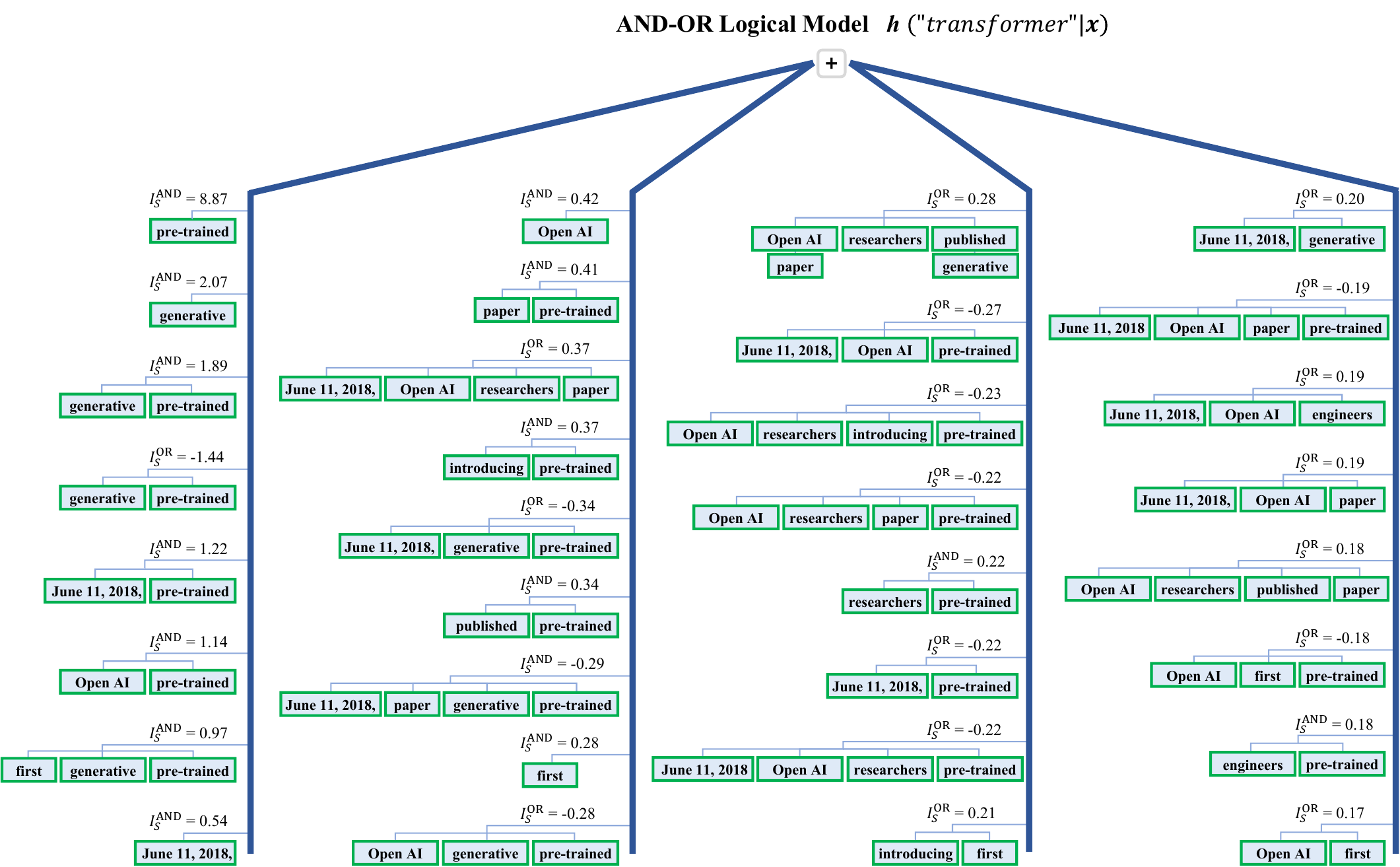}
  	\caption{The logical model representing the inference logic of the DeepSeek-R1-Distill-LLaMA-8B model on the input prompt ``On June 11, 2018, OpenAI researchers and engineers published a paper introducing the first generative pre-trained.'' The predicted next word is ``transformer.''}
  	\label{fig:aog_2}
\end{figure}

\begin{figure}[H]
  	\centering
  	\includegraphics[width=0.86\textwidth]{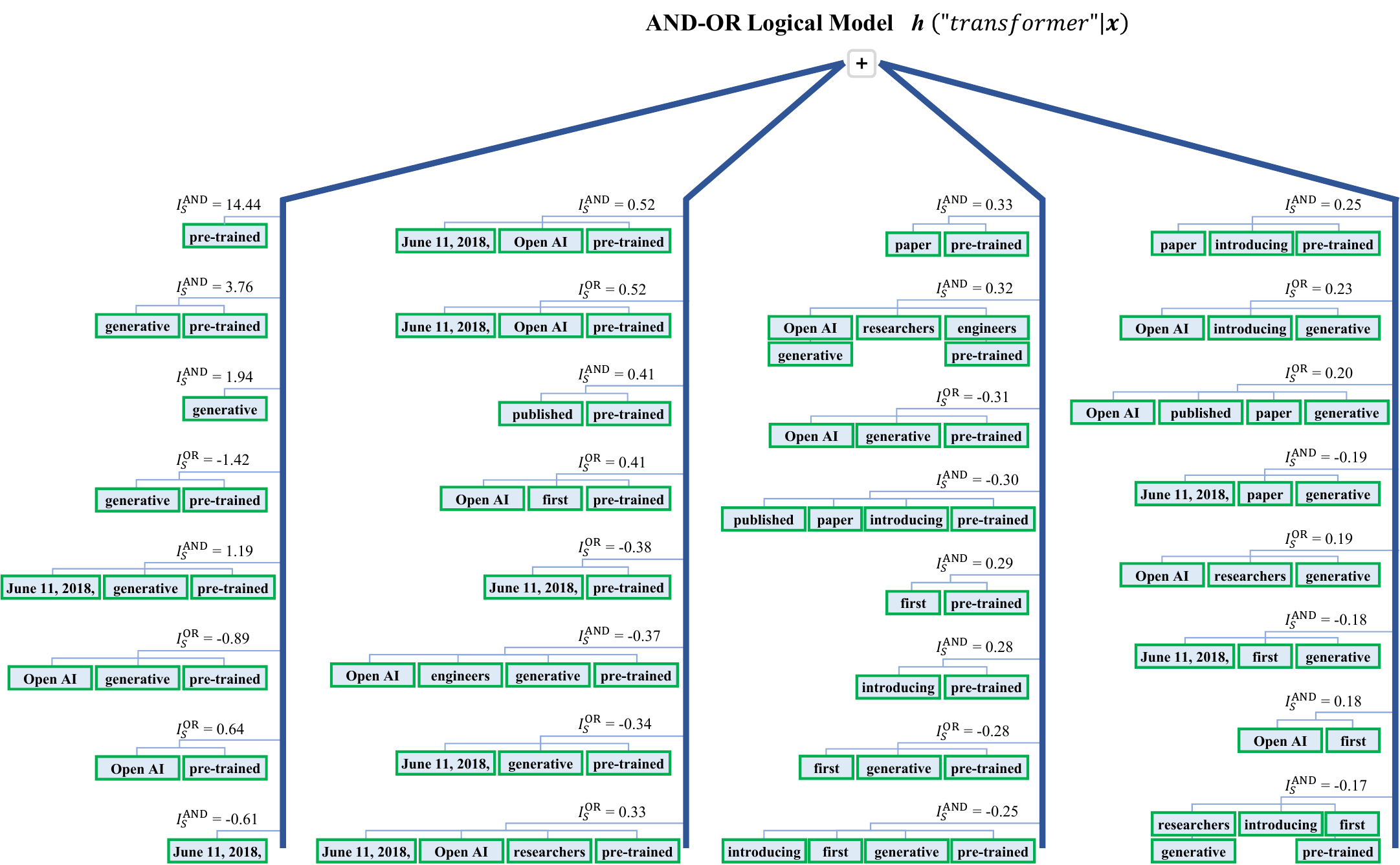}
  	\caption{The logical model representing the inference logic of the Qwen-2.5-7B model on the input prompt ``On June 11, 2018, OpenAI researchers and engineers published a paper introducing the first generative pre-trained.'' The predicted next word is ``transformer.''}
  	\label{fig:aog_3}
\end{figure}

The following four figures show the logical models that mathematically represent the inference logic of DeepSeek-R1-Distill-LLaMA-8B model (before and after pruning operation) on two input prompts, respectively.

\begin{figure}[H]
  	\centering
  	\includegraphics[width=0.86\textwidth]{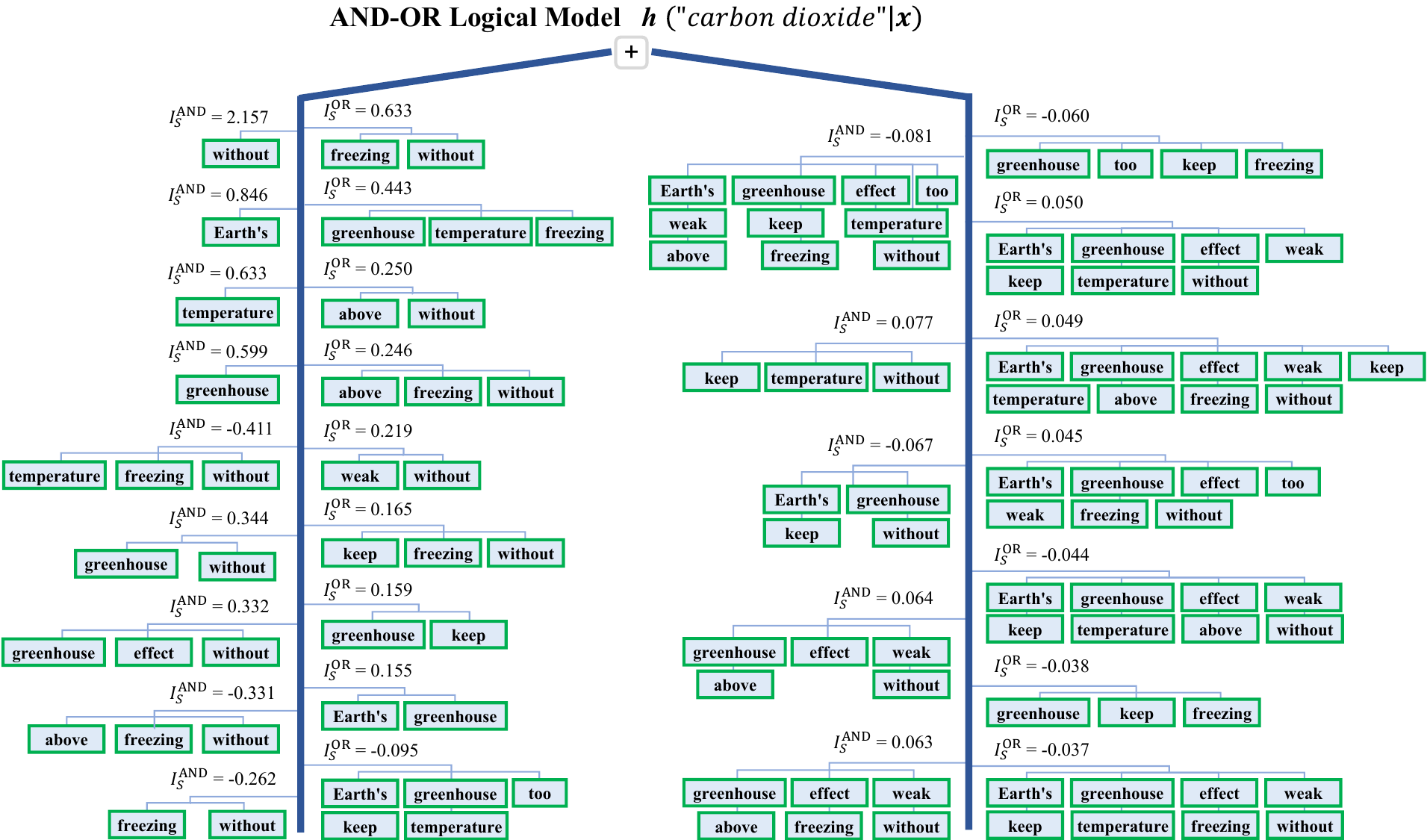}
  	\caption{The logical model representing the inference logic of the original DeepSeek-R1-Distill-LLaMA-8B model on the input prompt ``Earth's greenhouse effect would be too weak to keep temperature above freezing without.'' The predicted next word is ``carbon dioxide.''}
  	\label{fig:aog_4}
\end{figure}

\begin{figure}[H]
  	\centering
  	\includegraphics[width=0.86\textwidth]{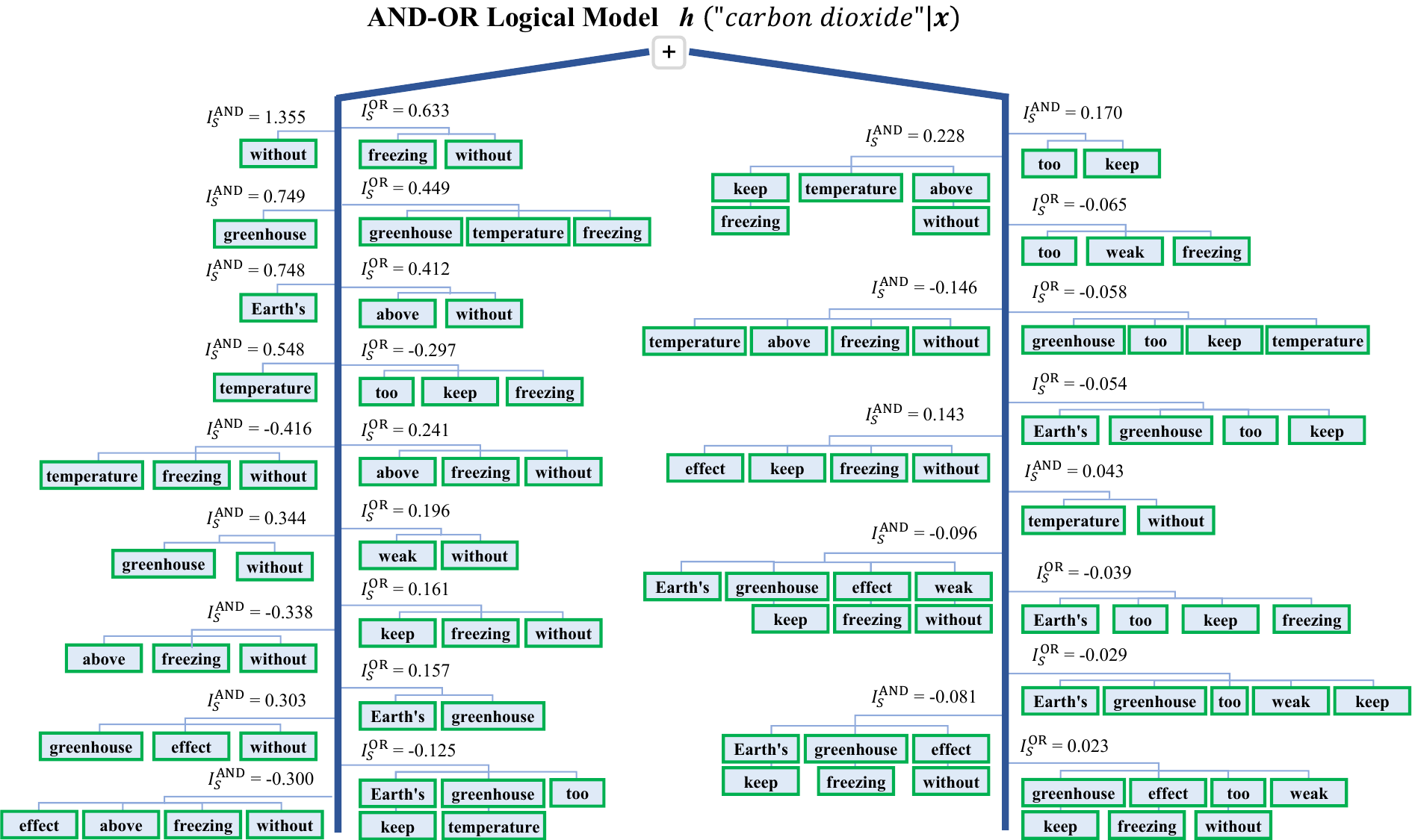}
  	\caption{The logical model representing the inference logic of the pruned DeepSeek-R1-Distill-LLaMA-8B model (under a pruning ratio of 41.01) on the input prompt ``Earth's greenhouse effect would be too weak to keep temperature above freezing without.'' The predicted next word is ``carbon dioxide.''}
  	\label{fig:aog_5}
\end{figure}

\begin{figure}[H]
  	\centering
  	\includegraphics[width=0.86\textwidth]{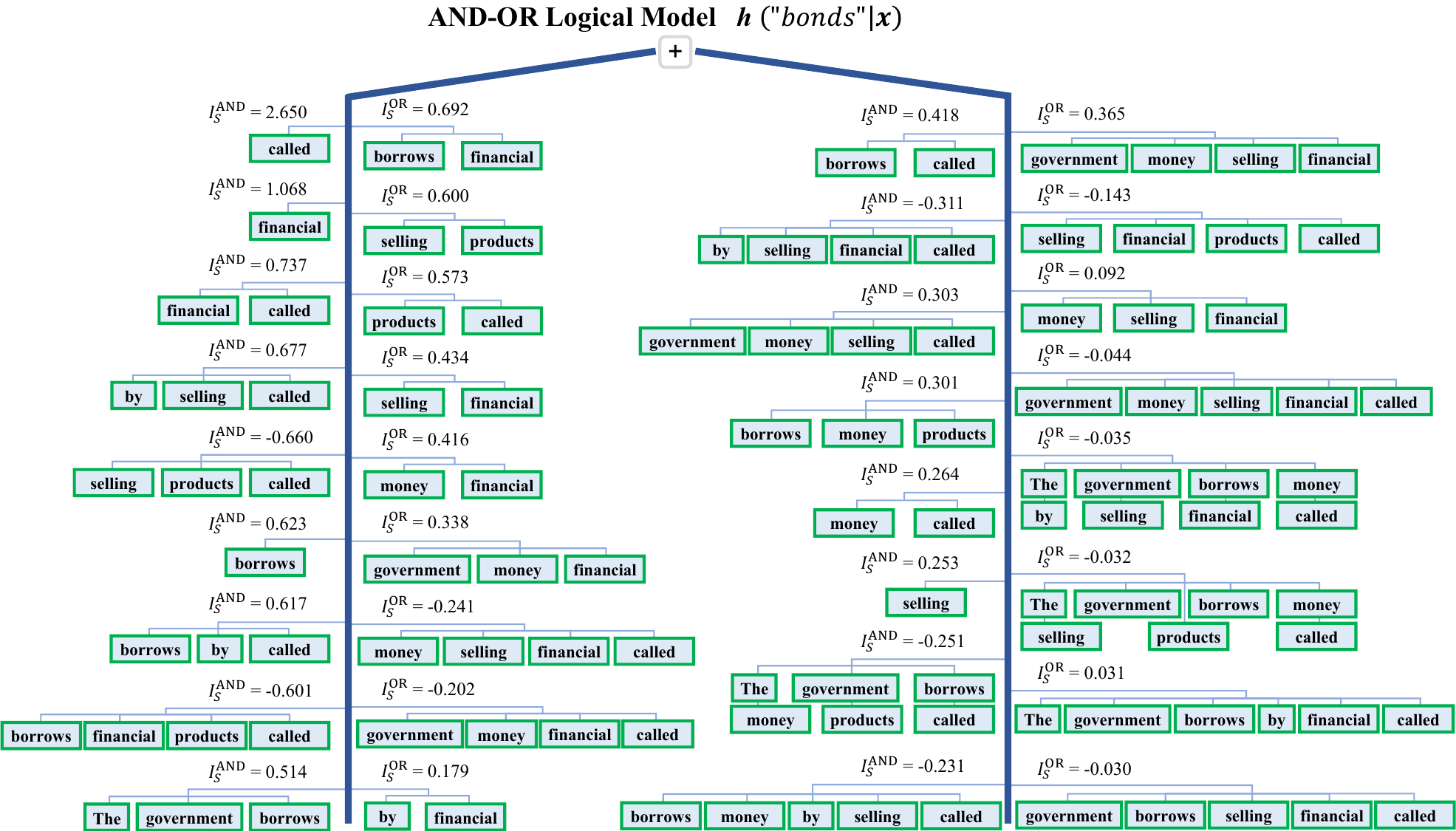}
  	\caption{The logical model representing the inference logic of the original DeepSeek-R1-Distill-LLaMA-8B model on the input prompt ``The government borrows money by selling financial products called.'' The predicted next word is ``bonds.''}
  	\label{fig:aog_6}
\end{figure}

\begin{figure}[H]
  	\centering
  	\includegraphics[width=0.86\textwidth]{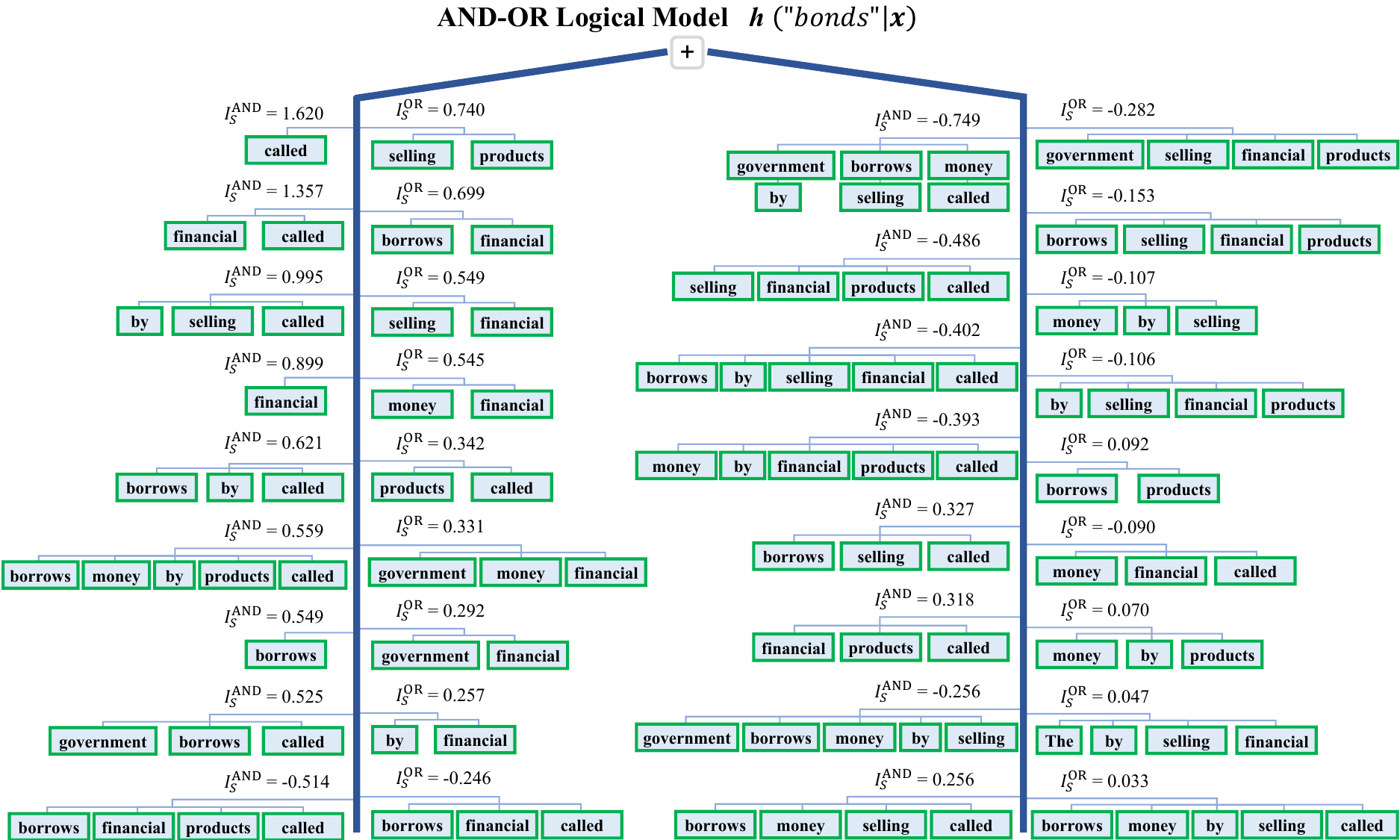}
  	\caption{The logical model representing the inference logic of the pruned DeepSeek-R1-Distill-LLaMA-8B model (under a pruning ratio of 41.01) on the input prompt ``The government borrows money by selling financial products called.'' The predicted next word is ``bonds.''}
  	\label{fig:aog_7}
\end{figure}
\newpage


\section{Declaration of Large Language Model usage}

In this work, large language models (LLMs) were used solely as a general-purpose writing assistant to polish the grammar and improve the clarity of the text. No part of the research ideation, experiment design, data analysis, or substantive content generation relied on LLMs. The authors take full responsibility for the content of the paper.
\newpage

\end{document}